\documentclass{article}

     \usepackage[nonatbib, final]{neurips_2021}

\usepackage[utf8]{inputenc} 
\usepackage[T1]{fontenc}    
\usepackage{hyperref}       
\usepackage{url}            
\usepackage{booktabs}       
\usepackage{amsfonts}       
\usepackage{nicefrac}       
\usepackage{microtype}      
\usepackage{graphicx}
\usepackage{caption}
\usepackage{subcaption}
\usepackage{amsmath}
\usepackage{xcolor}         
\usepackage{dsfont}
\usepackage{lipsum}
\usepackage{tikz}
\usetikzlibrary{positioning}
\usepackage{graphicx}
\usepackage{wrapfig}
\usepackage{cleveref}

\title{ 
       Topographic VAEs learn Equivariant Capsules}

\author{%
  T. Anderson Keller \\
  UvA-Bosch Delta Lab\\
  University of Amsterdam\\
  \texttt{t.anderson.keller@gmail.com}
  \And
  Max Welling\\
  UvA-Bosch Delta Lab\\
  University of Amsterdam\\
  \texttt{m.welling@uva.nl}
}

\begin{document}

\maketitle

\begin{abstract}
In this work we seek to bridge the concepts of topographic organization and equivariance in neural networks. To accomplish this, we introduce the Topographic VAE: a novel method for efficiently training deep generative models with topographically organized latent variables. We show that such a model indeed learns to organize its activations according to salient characteristics such as digit class, width, and style on MNIST. Furthermore, through topographic organization over time (i.e. temporal coherence), we demonstrate how predefined latent space transformation operators can be encouraged for observed transformed input sequences -- a primitive form of unsupervised learned equivariance. We demonstrate that this model successfully learns sets of approximately equivariant features (i.e. "capsules") directly from sequences and achieves higher likelihood on correspondingly transforming test sequences. Equivariance is verified quantitatively by measuring the approximate commutativity of the inference network and the sequence transformations. Finally, we demonstrate approximate equivariance to complex transformations, expanding upon the capabilities of existing group equivariant neural networks.

\end{abstract}

\section{Introduction}
Many parts of the brain are organized topographically. Famous examples are the ocular dominance maps and the orientation maps in V1. What is the advantage of such organization and what can we learn from it to develop better inductive biases for deep neural network architectures?
\begin{figure}[h!]
\centering
\includegraphics[width=1.0\linewidth]{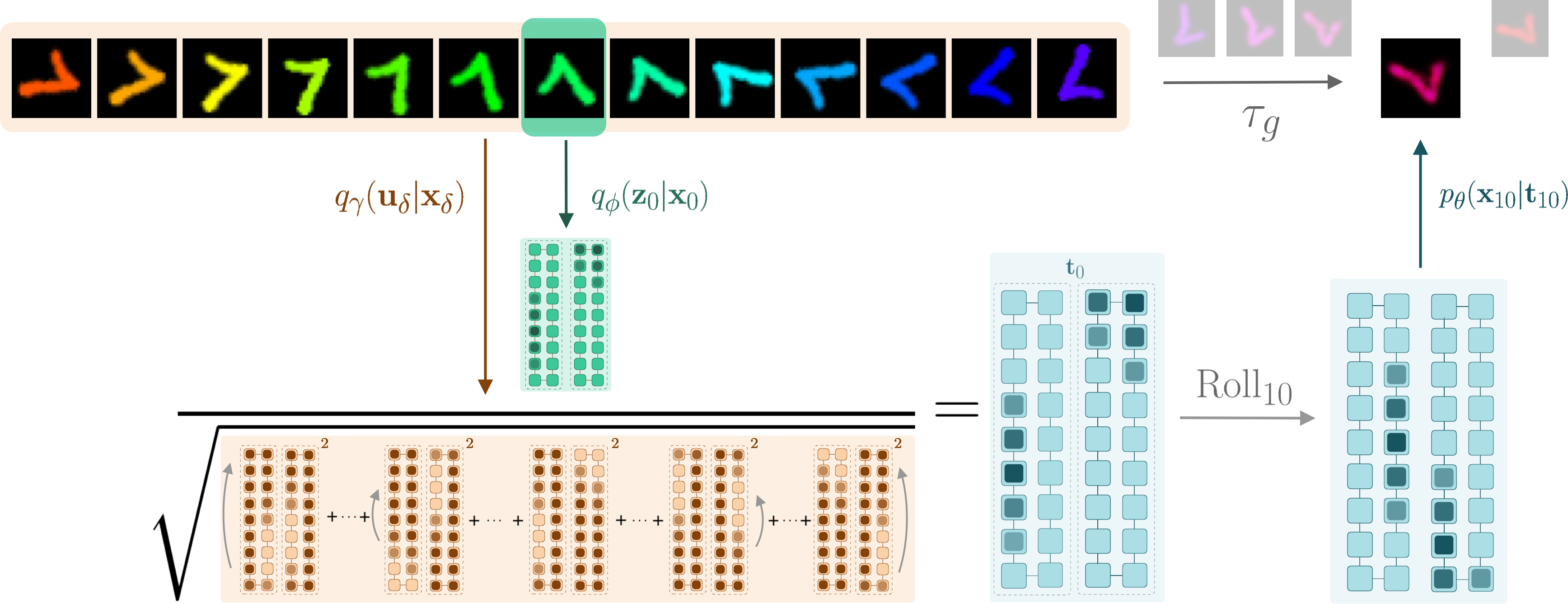}
\caption{Overview of the Topographic VAE with shifting temporal coherence. The combined color/rotation transformation in input space $\tau_g$ becomes encoded as a $\mathrm{Roll}$ within the capsule dimension. The model is thus able decode unseen sequence elements by encoding a partial sequence and $\mathrm{Roll}$ing activations within the capsules. We see this resembles a commutative diagram.}
\label{fig:traversal}
\vspace{-4mm}
\end{figure}

One potential explanation for the emergence of topographic organization is provided by the principle of redundancy reduction \cite{barlow1961possible}. In the language of Information Theory, redundancy wastes channel capacity, and thus to represent information as efficiently as possible, the brain may strive to transform the input to a neural code where the activations are statistically maximally independent.
In the machine learning literature, this idea resulted in Independent Component Analysis (ICA) which linearly transforms the input to a new basis where the activities are independent and sparse \cite{SejnowskiICA, comon1994independent, hyvarinen2000independent, olshausen1997sparse}. It was soon realized that there are remaining higher order dependencies (such as correlation between absolute values) that can not be transformed away by a linear transformation. For example, along edges of an image, linear-ICA components (e.g. gabor filters) still activate in clusters even though the sign of their activity is unpredictable \cite{Portilla03, Wainwright99b}. This led to new algorithms that explicitly model these remaining dependencies through a topographic organization of feature activations \cite{hyvarinen2001topographic, Osindero2006, osindero2004contrastive, welling2003learning}. 
Such topographic features were reminiscent of pinwheel structures observed in V1, encouraging multiple comparisons with topographic organization in the biological visual system \cite{hyvarinen2009natural, HYVARINEN20012413, ma2008overcomplete}. 

A second, almost independent body of literature developed the idea of ``equivariance'' of neural network feature maps under symmetry transformations. The idea of equivariance is that symmetry transformations define equivalence classes as the orbits of their transformations, and we wish to maintain this structure in the deeper layers of a neural network. For instance, for images, asserting a rotated image contains the same object for all rotations, the transformation of rotation then  defines an orbit where the elements of that orbit can be interpreted as pose or angular orientation. When an image is processed by a neural network, we want features at different orientations to be able to be combined to form new features, but we want to ensure the relative pose information between the features is preserved for all orientations. This has the advantage that the equivalence class of rotations for the complex composite features is guaranteed to be maintained, allowing for the extraction of invariant features, a unified pose, and increased data efficiency. Such ideas are reminiscent of the capsule networks of Hinton et al. \cite{capsules2011, e2018matrix, sabour2017dynamic}, and indeed formal connections to equivariance have been made \cite{lenssen2018group}. Interestingly, by explicitly building neural networks to be equivariant, we additionally see geometric organization of activations into these equivalence classes, and further, the elements within an equivalence class are seen to exhibit higher-order non-Gaussian dependencies \cite{Lyu08b, Lyu08, Wainwright99b, Wainwright00}. The insight of this connection between topographic organization and equivariance hints at a possibility to encourage approximate equivariance from an induced topology in feature space.

To build a model, we need to ask what mechanisms could induce topographic organization of \emph{observed transformations} specifically? We have argued that removing dependencies between latent variables is a possible mechanism; however, to obtain the more structured organisation of equivariant capsule representations, the usual approach is to hard-code this structure into the network, or to encourage it through regularization terms \cite{learninginvariance, Learningtoconvole}. To achieve this same structure \emph{unsupervised}, we propose to incorporate another key inductive bias: ``temporal coherence'' \cite{foldiak, hurri2003TC, stone1996, wiskott2002slow}. The principle of temporal coherence, or ``slowness'', asserts than when processing correlated sequences, we wish for our representations to change smoothly and slowly over space and time. Thinking of time sequences as symmetry transformations on the input, we desire features undergoing such transformations to be grouped into equivariant capsules. We therefore suggest that encouraging slow feature transformations to take place \emph{within a capsule} could induce such grouping from sequences alone.

In the following sections we will explain the details of our Topographic Variational Autoencoder  which lies at the intersection of topographic organization, equivariance, and temporal coherence, thereby learning approximately equivariant capsules from sequence data completely unsupervised.

\section{Related Work}
The history of statistical models upon which this work builds is vast,  including sparse coding \cite{olshausen1997sparse}, Independant Component Analysis (ICA) \cite{SejnowskiICA, comon1994independent, hyvarinen2000independent}, Slow Feature Analysis (SFA) \cite{probSFA, wiskott2002slow}, and Gaussian scale mixtures \cite{Lyu08, Portilla03, Wainwright99b, Wainwright00}. Most related to this work are topographic generative models including Generative Topographic Maps \cite{GTM}, Bubbles \cite{bubbles}, Topographic ICA \cite{hyvarinen2001topographic}, and the Topographic Product of Student's-t \cite{osindero2004contrastive, welling2003learning}. Prior work on learning equivariant and invariant representations is similarly vast and also has a deep relationship with these generative models. Specifically, Independant Subspace Analysis \cite{hyvarinen2000emergence, stuhmer2019independent}, models involving temporal coherence \cite{foldiak, hurri2003TC, stone1996, wiskott2002slow}, and Adaptive Subspace Self Organizing Maps \cite{kohonen1996emergence} have all demonstrated the ability to learn invariant feature subspaces and even `disentangle' space and time \cite{Grathwohl16, stuhmer2019independent}. Our work assumes a similar generative model to these works while additionally allowing for efficient estimation of the model through variational inference \cite{kingma2013auto, rezende2014stochastic}. Although our work is not the first to combine Student's-t distributions and variational inference \cite{boenninghoff2020variational}, it is the first to provide an efficient method to do so for Topographic Student's-t distributions. 

\looseness=-1
Another line of work has focused on constructing neural networks with equivariant representations separate from the framework of generative modeling. Analytically equivariant networks such as Group Equivariant Neural Networks \cite{cohen2016group}, and other extensions \cite{cohen2016steerable, finzi2020generalizing, finzi2021emlp, elise, ravanbakhsh2017equivariance, weiler20183d, scalespaces, worrall2017harmonic} propose to explicitly enforce symmetry to group transformations in neural networks through structured weight sharing. Alternatively, others propose supervised and self-supervised methods for \emph{learning} equivariance or invariance directly from the data itself \cite{learninginvariance, connor2021variational, Learningtoconvole}. One related example in this category uses a group sparsity regularization term to similarly learn topographic features for the purpose of modeling invariance \cite{kavukcuoglu2009learning}. We believe the Topographic Variational Autoencoder presented in this paper is another promising step in the direction of learning approximate equivariance, and may even hint at how such structure could be learned in biological neural networks.

\looseness=-1
Furthermore, the idea of disentangled representations \cite{bengio2013representation} has also been been connected to equivariance and representation theory in multiple recent papers \cite{bouchacourt2021addressing, cohen2015transformation, cohen2014learning,  higgins2018definition}. Our work shares a fundamental connection to this distributed operator definition of disentanglement, where the slow roll of capsule activations can be seen as the latent operator. Recently, the authors of \cite{klindt2021nonlinear} demonstrated that incorporating the principle of `slowness' in a variational autoencoder (VAE) yields the ability to learn disentangled representations from natural sequences. While similar in motivation, the generative model proposed in \cite{klindt2021nonlinear} is unrelated to topographic organization and equivariance, and is more aligned with traditional notions of disentanglement. 

\looseness=-1
Finally, and importantly, in the neuroscience literature, another popular explanation for topographic organization arises as the solution to the `wiring length' minimization problem \cite{koulakov2001orientation}. Recently, models which attempt to incorporate wiring length constraints have been shown to yield topographic organization of higher level features, ultimately resembling the `face patches' found in primates \cite{keller2021modeling, TDANN}. Interestingly, the model presented in this paper organizes activity based on the same statistical property (local correlation) as the wiring length proxies developed in \cite{TDANN}, but from a generative modeling perspective, demonstrating a computationally principled explanation for the same phenomenon.

\section{Background}
The model in this paper is a first attempt at bridging two yet disjoint classes of models: Topographic Generative Models, and Equivariant Neural Networks. In this section, we will provide a brief background on these two frameworks.
\subsection{Topographic Generative models}
\looseness=-1
Inspired by Topographic ICA, the class of Topographic Generative models can be understood as generative models where the joint distribution over latent variables does not factorize into entirely independent  factors, as is commonly done in ICA or VAEs, but instead has a more complex `local' correlation structure. The locality is defined by arranging the latent variables into an n-dimensional lattice or grid, and organizing variables such that those which are closer together on this grid have greater correlation of activities than those which are further apart. In the related literature, activations which are nearby in this grid are defined to have higher-order correlation, e.g. correlations of squared activations (aka `energy'), asserting that all first order correlations are removed by the initial ICA de-mixing matrix. 

Such generative models can be seen as hierarchical generative models where there exist higher level independent `variance generating' variables $\mathbf{V}$ which are combined locally to generate the variances $\boldsymbol{\sigma} = \phi(\mathbf{W}\mathbf{V})$ of the lower level topographic variables $\mathbf{T} \sim \mathcal{N}(\mathbf{0}, \boldsymbol{\sigma}^2 \mathbf{I})$, for an appropriate non-linearity $\phi$. The variables $\mathbf{T}$ are thus independent  conditioned on $\boldsymbol{\sigma}$. Other related models which can be described under this umbrella include \emph{Independent Subspace Analysis} (ISA) \cite{hyvarinen2000emergence} where all variables within a predefined subspace (or `capsule') share a common variance, and `\emph{temporally coherent}' models \cite{hurri2003TC} where the energy of a given variable between time steps is correlated by extending the topographic neighborhoods over the time dimension \cite{bubbles}. The topographic latent variable $\mathbf{T}$ can additionally be described as an instance of a Gaussian scale mixture (GSM). GSMs have previously been used to model the observed non-Gaussian dependencies between coefficients of steerable wavelet pyramids (interestingly also equivariant to translation \& rotation) \cite{Portilla03, Wainwright99b, Wainwright00}.

\subsection{Group Equivariant Neural Networks}
\looseness=-1
Equivariance is the mathematical notion of symmetry for functions. A function is said to be an equivariant map if the the result of transforming the input and then computing the function is the same as first computing the function and then transforming the output. In other words, the function and the transformation commute. Formally, $f(\tau_\rho [\mathbf{x}]) = \Gamma_\rho [f(\mathbf{x})]$, where $\tau$ and $\Gamma$ denote the (potentially different) operators on the domain and co-domain respectively, but are indexed by the same element $\rho$. 

It is well known that convolutional maps in neural networks are translation equivariant, i.e., given a translation $\Gamma_{\rho}$
(applied to each feature map separately) and a convolutional map $f(\cdot)$, we have $f(\Gamma_{\rho}[\mathbf{x}]) = \Gamma_{\rho}[f(\mathbf{x})]$. This can be extended to other transformations (e.g. rotation or mirroring) using Group convolutions ($G$-convolutions)~\cite{cohen2016group}. As a result of the design of $G$-convolutions, feature maps that are related to each other by a rotation of the filter/input are grouped together. Moreover, a rotation of the input results in a transformation (i.e. a permutation and rotation) on the activations of each of these groups in the output. Hence, we can think of these equivalence class groups as capsules where transformations of the input only cause structured transformations $\emph{within}$ a capsule.
As we will demonstrate later, this is indeed analogous to the structure of the representation learned by the Topographic VAE with temporal coherence -- a transformation of the input yields a cyclic permutation of activations \emph{within} each capsule. However, due to the approximate \emph{learned} nature of the equivariant representation, the Topographic VAE does not require the transformations $\tau_{\rho}$ to constitute a group.

\section{The Generative Model}
The generative model proposed in this paper is based on the Topographic Product of Student's-t (TPoT) model as developed in \cite{osindero2004contrastive, welling2003learning}. In the following, we will show how a TPoT random variable can be constructed from a set of independent univariate standard normal random variables, enabling efficient training through variational inference. Subsequently, we will construct a new model where topographic neighborhoods are extended over time, introducing temporal coherence and encouraging the unsupervised learning of approximately equivariant subspaces we call `capsules'. 

\subsection{The Product of Student's-t Model}
\looseness=-1
We assume that that our observed data is generated by a latent variable model where the joint distribution over observed and latent variables $\mathbf{x}$ and $\mathbf{t}$ factorizes into the product of the conditional and the prior. The prior distribution $p_{\mathbf{T}}(\mathbf{t})$ is assumed to be a Topographic Product of Student's-t (TPoT) distribution, and we parameterize the conditional distribution with a flexible function approximator:
\begin{equation}
\label{eqn:generative_model}
    p_{\mathbf{X}, \mathbf{T}}(\mathbf{x}, \mathbf{t}) = p_{\mathbf{X}| \mathbf{T}}(\mathbf{x}|\mathbf{t})p_{\mathbf{T}}(\mathbf{t}) \ \ \ \ \ \ \ \  p_{\mathbf{X}|\mathbf{T}}(\mathbf{x}|\mathbf{t}) = p_{\theta}(\mathbf{x} | g_{\theta}(\mathbf{t}))  \ \ \ \ \ \ \ \ p_{\mathbf{T}}(\mathbf{t}) = \mathrm{TPoT}(\mathbf{t}; \nu)
\end{equation}
The goal of training is thus to learn the parameters $\theta$ such that the marginal distribution of the model $p_{\theta}(\mathbf{x})$ matches that of the observed data. Unfortunately, the marginal likelihood is generally intractable except for all but the simplest choices of $g_{\theta}$ and $p_{\mathbf{T}}$ \cite{Osindero2006}. Prior work has therefore resorted to techniques such as contrastive divergence with Gibbs sampling \cite{welling2003learning} to train TPoT models as energy based models. In the following section, we instead demonstrate how TPoT variables can be constructed as a deterministic function of Gaussian random variables, enabling the use of variational inference and efficient maximization of the likelihood through the evidence lower bound (ELBO).

\subsection{Constructing the Product of Student's-t Distribution}
\looseness=-1
First, note a univariate Student's-t random variable $T$ with $\nu$ degrees of freedom can be defined as:
\begin{equation}
    T = \frac{Z}{\sqrt{\frac{1}{\nu}\sum^{\nu}_i U_i^2}} \ \ \ \ \ \mathrm{with}\ \ \  Z, U_i \sim \mathcal{N}(0, 1)\ \  \forall i\ \ 
\end{equation}
Where $Z$ and $\{U_i\}_{i=1}^\nu$ are independent standard normal random variables. If $\mathbf{T}$ is a multidimensional Student's-t random variable, composed of independent  $Z_i$ and $U_i$, then $\mathbf{T} \sim \mathrm{PoT(\nu)}$, i.e.:
\begin{equation}
    \mathbf{T} = \left[\frac{Z_1}{\sqrt{\frac{1}{\nu}\sum^{\nu}_{i=1} U_i^2}},\ \  \frac{Z_2}{\sqrt{\frac{1}{\nu}\sum^{2\cdot\nu}_{i=\nu+1} U_i^2}},\ \  \ldots\ \  \frac{Z_n}{\sqrt{\frac{1}{\nu}\sum^{n\cdot\nu}_{i=(n-1)\cdot\nu+1} U_i^2}}\right] \sim \mathrm{PoT(\nu)}
\end{equation}
Note that the Student's-t variable $T$ is large when most of the $\{U_i\}_i$ in its set are small. We can therefore think of the $\{U_i\}_i$ as constraint violations rather then pattern matches: if the input matches all constraints $U_i\approx 0$, the corresponding $T$ variables will activate (see \cite{hinton2013discovering} for further discussion). 

\subsection{Introducing Topography}
\looseness=-1
To make the PoT distribution topographic, we strive to correlate the scales of $T_j$ which are `nearby' in our topographic layout. One way to accomplish this is by \emph{sharing} some $U_i$-variables between neighboring $T_j$'s. Formally, we define overlapping neighborhoods $\mathsf{N}(j)$ for each variable $T_j$ and write:
\begin{equation}
    \mathbf{T} = \left[\frac{Z_1}{\sqrt{\frac{1}{\nu}\sum_{i \in\mathsf{N}(1)} U_i^2}},\ \  \frac{Z_2}{\sqrt{\frac{1}{\nu}\sum_{i\in\mathsf{N}(2)} U_i^2}},\ \  \ldots\ \  \frac{Z_n}{\sqrt{\frac{1}{\nu}\sum_{i\in\mathsf{N}(n)} U_i^2}}\right] \sim \mathrm{TPoT(\nu)}
\end{equation}
With some abuse of notation, if we define $\mathbf{W}$ to be the adjacency matrix which defines our neighborhood structure, $\mathbf{U}$ and $\mathbf{Z}$ to be the vectors of random variables $U_i$ and $Z_j$, we can write the above succinctly as:
\begin{align}
\label{eqn:full_T}
    \mathbf{T} = \left[\frac{Z_1}{\sqrt{\frac{1}{\nu}W_1\mathbf{U}^2}},\ \  \frac{Z_2}{\sqrt{\frac{1}{\nu}W_2\mathbf{U}^2}},\ \  \ldots\ \  \frac{Z_n}{\sqrt{\frac{1}{\nu}W_n\mathbf{U}^2}}\right]  = \frac{\mathbf{Z}}{\sqrt{\frac{1}{\nu} \mathbf{W} \mathbf{U}^2}} \sim \mathrm{TPoT(\nu)}
\end{align}
Due to non-linearities such as ReLUs which may alter input distributions, it is beneficial to allow the $Z$ variables to model the mean and scale. We found this can be achieved with the following parameterization: \scalebox{0.7}{$\mathbf{T} = \frac{\mathbf{Z} - \mu}{\sigma\sqrt{1/\nu \mathbf{W} \mathbf{U}^2}}$}. In practice, we found that $\sigma = \sqrt{\nu}$ often works well, finally yielding:
\begin{align}
\label{eqn:final_T}
  \mathbf{T} = \frac{\mathbf{Z} - \mu}{\sqrt{\mathbf{W} \mathbf{U}^2}} 
\end{align}
Given this construction, we observe that the TPoT generative model can instead be viewed as a latent variable model where all random variables are Gaussian and the construction of $\mathbf{T}$ in Equation \ref{eqn:final_T} is the first layer of the generative `decoder': $g_{\theta}(\mathbf{t}) = g_{\theta}(\mathbf{u}, \mathbf{z})$. In Section \ref{sec:TVAE} we then leverage this interpretation to show how an approximate posterior for the latent variables $\mathbf{Z}$ and $\mathbf{U}$ can be trained through variational inference.

\subsection{Capsules as Disjoint Topologies}
\begin{wrapfigure}{r}{0.4\linewidth}
\vspace{-10mm}
\centering
  \includegraphics[width=0.4\textwidth]{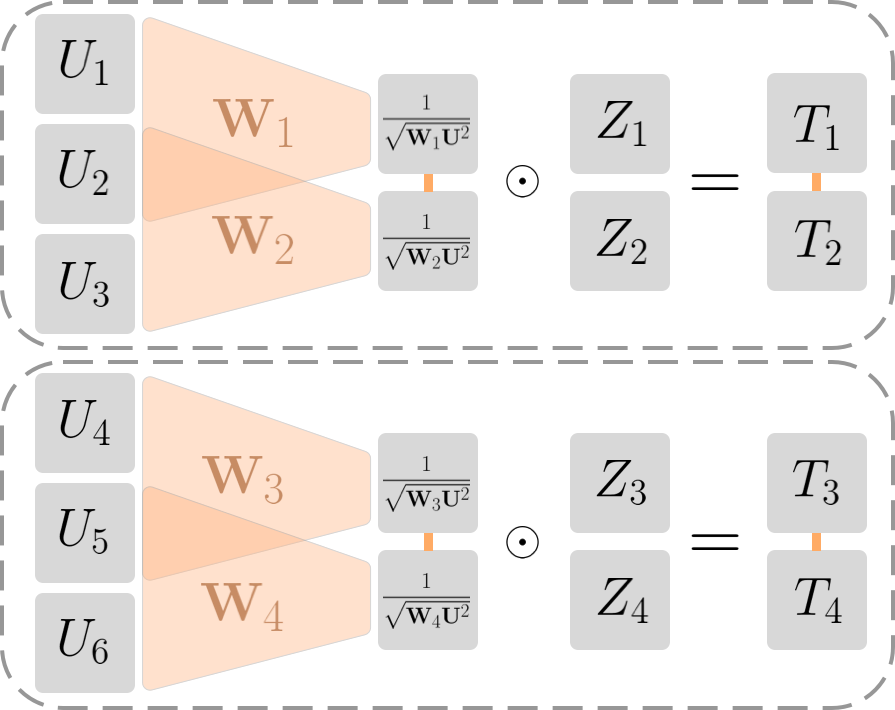}
  \caption{An example of a neighborhood structure which induces disjoint topologies (aka capsules). Lines between variables $T_i$ indicate that sharing of $U_i$, and thus correlation.}
  \label{fig:depindep}
\end{wrapfigure}
One setting of neighborhood structure $\mathbf{W}$ which is of particular interest is when there exist multiple sets of disjoint neighborhoods. Statistically, the variables of two disjoint topologies are completely independent.
An example of a capsule neighborhood structure is shown in Figure \ref{fig:depindep}. The idea of independant subspaces has previously been shown to learn invariant feature subspaces in the linear setting and is present in early work on Independent Subspace Analysis \cite{hyvarinen2000emergence} and Adaptive Subspace Self Organizing Maps (ASSOM) \cite{kohonen1996emergence}. It is also very reminiscent of the transformed sets of features present in a group equivariant convolutional neural network. In the next section, we will show how temporal coherence can be leveraged to induce the encoding of observed transformations into the internal dimensions of such capsules thereby yielding unsupervised approximately equivariant capsules.

\subsection{Temporal Coherence and Learned Equivariance}
\label{sec:temporal_coherence}
We now describe how the induced topographic organization can be leveraged to learn a basis of approximately equivariant capsules for observed transformation sequences. The resulting representation is composed of a large set of `capsules' where the dimensions inside the capsule are topographically structured, but between the capsules there is independence. To benefit from sequences of input, we encourage topographic structure over time between sequentially permuted activations within a capsule, a property we refer to as \emph{shifting temporal coherence}. 

\subsubsection{Temporal Coherence}
Temporal Coherence can be measured as the correlation of squared activation between time steps. One way we can achieve this in our model is by having $T_j$ share $U_i$ between time steps. Formally, the generative model is identical to Equation \ref{eqn:generative_model}, factorizing over timesteps denoted by subscript $l$, i.e. $p_{\mathbf{X}_l, \mathbf{T}_l}(\mathbf{x}_l, \mathbf{t}_l) = p_{\mathbf{X}_l| \mathbf{T}_l}(\mathbf{x}_l|\mathbf{t}_l)p_{\mathbf{T}_l}(\mathbf{t}_l)$. However, $\mathbf{T}_l$ is now a function of a sequence $\{\mathbf{U}_{l+\delta}\}_{\delta=-L}^{L}$:
\begin{align}
\label{eqn:T_seq}
    \mathbf{T}_l = \frac{\mathbf{Z}_l - \mu}{\sqrt{\mathbf{W} 
              \left[
              \mathbf{U}_{l+L}^2; 
              \cdots; \mathbf{U}_{l-L}^2
              \right]
}}
\end{align}
Where $\left[\mathbf{U}_{l+L}^2; \cdots; \mathbf{U}_{l-L}^2\right]$ denotes vertical concatenation of the column vectors $\mathbf{U}_l$, and $2L$ can be seen as the window size. We see that the choice of $\mathbf{W}$ now defines correlation structure over time. In prior work on temporal coherence (denoted `Bubbles' \cite{bubbles}), the grouping over time is such that a given variable $T_{l,i}$ has correlated energy with \emph{the same spatial location} $(i)$ at a previous time step $(l-1)$ \big(i.e. $\mathrm{cov}(T_{l,i}^2, T_{l-1,i}^2) > 0$\big). This can be implemented as: 
\begin{equation}
\label{eqn:temporal_coherence}
\mathbf{W} \left[\mathbf{U}_{l+L}^2; \cdots; \mathbf{U}_{l-L}^2\right] = \sum_{\delta=-L}^{L} \mathbf{W}_\delta\mathbf{U}_{l+\delta}^2
\end{equation}
Where $\mathbf{W}_\delta$ defines the topography for a single timestep, and is typically the same for all timesteps. 

\subsubsection{Learned Equivariance with Shifting Temporal Coherence}
In our model, instead of requiring a single location to have correlated energies over a sequence,
we would like variables at sequentially permuted locations \emph{within a capsule} to have correlated energy between timesteps \big($\mathrm{cov}(T_{l,i}^2, T_{l-1,i-1}^2) > 0$\big). Similarly, this can be implemented as: 
\begin{equation}
\label{eqn:eq_roll}
    \mathbf{W} \left[\mathbf{U}_{l+L}^2; \cdots; \mathbf{U}_{l-L}^2\right] = \sum_{\delta=-L}^{L} \mathbf{W}_\delta \mathrm{Roll}_{\delta}(\mathbf{U}_{l+\delta}^2)
\end{equation}
Where $\mathrm{Roll}_{\delta}(\mathbf{U}_{l+\delta}^2)$ denotes a cyclic permutation of $\delta$ steps along the capsule dimension. 
The exact implementation
of $\mathrm{Roll}$ can be found in Section \ref{sec:roll_def}. As we will show in Section \ref{sec:equivariant_experiments}, TVAE models with such a topographic structure learn to encode observed sequence transformations as $\mathrm{Roll}$s within the capsule dimension, analogous to a group equivariant neural network where 
$\tau_\rho$ and $\mathrm{Roll}_1$ can be seen as the action of the transformation $\rho$ on the input and output spaces respectively.

\section{Topographic VAE}
\label{sec:TVAE}
To train the parameters of the generative model $\theta$, we use the above formulation to parameterize an approximate posterior for $\mathbf{t}$ in terms of a deterministic transformation of approximate posteriors over simpler Gaussian latent variables $\mathbf{u}$ and $\mathbf{z}$. Explicitly:
\begin{gather}
\label{eqn:tvae1}
        q_{\phi}(\mathbf{z}_l|\mathbf{x}_l) = \mathcal{N}\big(\mathbf{z}_l; \mu_{\phi}(\mathbf{x}_l), \sigma_{\phi}(\mathbf{x}_l) \mathbf{I}\big) \hspace{8mm} 
        p_{\theta}(\mathbf{x}_l | g_{\theta}(\mathbf{t}_l)) =  p_{\theta}(\mathbf{x}_l | g_{\theta}(\mathbf{z}_l, \{\mathbf{u}_l\}))
\\
\label{eqn:tvae2}
    q_{\gamma}(\mathbf{u}_l|\mathbf{x}_l) = \mathcal{N}\big(\mathbf{u}_l ; \mu_{\gamma}(\mathbf{x}_l), \sigma_{\gamma}(\mathbf{x}_l) \mathbf{I}\big) \hspace{16mm} 
    \mathbf{t}_l = \frac{\mathbf{z}_l - \mu}{\sqrt{\mathbf{W} \left[ \mathbf{u}_{l+L}^2; \cdots; \mathbf{u}_{l-L}^2\right]}}
\end{gather}
We denote this model the Topographic VAE (TVAE) and optimize the parameters $\theta, \phi, \gamma$ (and $\mu$) through the ELBO, summed over the sequence length $S$:
\fontsize{9.5}{10}
\begin{equation}
    \label{eqn:elbo}
    \sum_{l=1}^S \mathbb{E}_{Q_{\phi,\gamma}(\mathbf{z}_l,\mathbf{u}_l|\{\mathbf{x}_l\})}
    \left([\log p_{\theta}(\mathbf{x}_l|g_{\theta}(\mathbf{t}_l))] - D_{KL}[q_{\phi}(\mathbf{z}_l|\mathbf{x}_l) || p_{\mathbf{Z}}(\mathbf{z}_l)] - D_{KL}[q_{\gamma}(\mathbf{u}_l|\mathbf{x}_l) || p_{\mathbf{U}}(\mathbf{u}_l)]\right)
\end{equation}
\normalsize
where $Q_{\phi,\gamma}(\mathbf{z}_l,\mathbf{u}_l|\{\mathbf{x}_l\})=  q_{\phi}(\mathbf{z}_l|\mathbf{x}_l)\prod_{\delta=-L}^L q_{\gamma}(\mathbf{u}_{l+\delta}|\mathbf{x}_{l+\delta})$, and $\{\cdot\}$ denotes a set over time.

\section{Experiments}
\label{sec:experiments}
In the following experiments, we demonstrate the viability of the Topographic VAE as a novel method for training deep topographic generative models. Additionally, we quantitatively verify that shifting temporal coherence yields approximately equivariant capsules by computing an `equivariance loss' and a correlation metric inspired by the disentanglement literature. We show that equivariant capsule models yield higher likelihood than baselines on test sequences, and qualitatively support these results with visualizations of sequences reconstructed purely from $\mathrm{Roll}$ed capsule activations.  

\subsection{Evaluation Methods}
As depicted in Figure \ref{fig:traversal}, we make use of \emph{capsule traversals} to qualitatively visualize the transformations learned by our network. Simply, these are constructed by encoding a partial sequence into a $\mathbf{t}_0$ variable, and decoding sequentially $\mathrm{Roll}$ed copies of this variable. Explicitly, in the top row we show the data sequence $\{\mathbf{x}_l\}_l$, and in the bottom row we show the decoded sequence: $\{g_{\theta}(\mathrm{Roll}_{l}(\mathbf{t_0}))\}_l$. 

To measure equivariance quantitatively, we measure an \emph{equivariance error} similar to \cite{Learningtoconvole}. The equivariance error can be seen as the difference between traversing the two distinct paths of the commutative diagram, and provides some measure of how precisely the function and the transform commute. Formally, for a sequence of length $S$, and $\mathbf{\hat{t}}=\mathbf{t} / ||\mathbf{t}||_2$, the error is defined as: 
\begin{equation}
\label{eqn:eq_err}
\mathcal{E}_{eq}(\{\mathbf{t}_l\}_{l=1}^{S}) = \sum_{l=1}^{S-1} \sum_{\delta = 1}^{S - l} \left|\left|  \mathrm{Roll}_{\delta}(\mathbf{\hat{t}}_{l})  - \mathbf{\hat{t}}_{l+\delta}\right|\right|_1
\end{equation}
Additionally, inspired by existing disentanglement metrics, we measure the degree to which observed transformations in capsule space are correlated with input transformations by introducing a new metric we call $\mathrm{CapCorr}_y$. Simply, this metric computes the correlation between the amount of observed $\mathrm{Roll}$ of a capsule's activation at two timesteps $l$ and $l+\delta$, and the shift of the ground truth generative factors $y_l$ in that same time. Formally, for a correlation coefficient $\mathrm{Corr}$:
\begin{equation}
    \mathrm{CapCorr}(\mathbf{t}_{l}, \mathbf{t}_{l+\delta}, y_{l},  y_{l+\delta}) = \mathrm{Corr} \left(\mathrm{argmax}\left[\mathbf{t}_{l} \star \mathbf{t}_{l+\delta}\right], |y_l - y_{l+\delta}|\right)
    \label{eqn:capcorr}
\end{equation}
Where $\star$ is discrete periodic cross-correlation across the capsule dimension, and the correlation coefficient is computed across the entire dataset. We see the $\mathrm{argmax}$ of the cross-correlation is an estimate of the degree to which a capsule activation has shifted from time $l$ to $l + \delta$. To extend this to multiple capsules, we can replace the $\mathrm{argmax}$ function with the mode of the $\mathrm{argmax}$ computed for all capsules. We provide additional details and extensions of this metric in Section \ref{sec:appendix_capcorr}. For measuring capsule-metrics on baseline models which do not naturally have capsules, we simply arbitrarily divide the latent space into a fixed set of corresponding capsules and capsule dimensions, and provide such results as equivalent to `random baselines' for these metrics.

\subsection{Topographic VAE without Temporal Coherence}
\label{sec:2d_TVAE}
\begin{wrapfigure}{r}{0.4\linewidth}
\vspace{-10mm}
\includegraphics[width=\linewidth]{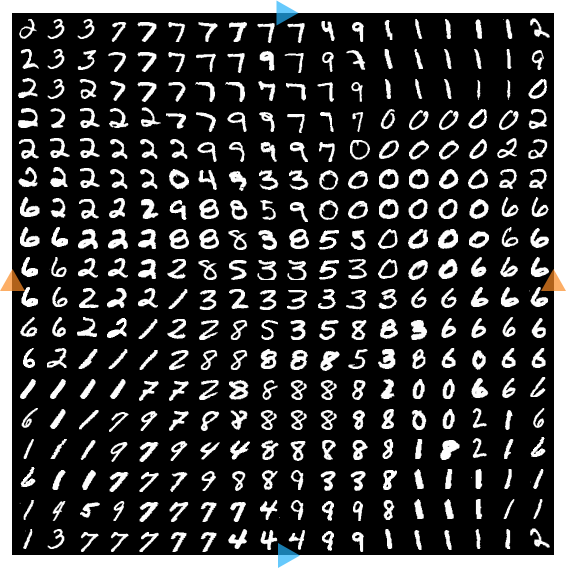}
\caption{Maximum activating images for a Topographic VAE trained with a 2D torus topography on MNIST.}
\vspace{-5mm}
\label{fig:TVAE}
\end{wrapfigure}
To validate the TVAE is capable of learning topographically organized representations with deep neural networks, we first perform experiments on a Topographic VAE without Temporal Coherence. The model is constructed as in Equations \ref{eqn:tvae1} and \ref{eqn:tvae2} with $L=0$, and is trained to maximize Equation \ref{eqn:elbo}. We fix $\mathbf{W}$ such that globally the latent variables are arranged in a grid on a 2-dimensional torus (a single capsule), and locally $\mathbf{W}$ sums over 5x5  2D groups of variables. In this setting, $\mathbf{W}$ can be easily implemented as 2D convolution with a 5x5 kernel of $1$'s, stride 1, and cyclic padding. We see that training the model with 3-layer MLP's for the encoders and decoder indeed yields a 2D topographic organization of higher level features. In Figure \ref{fig:TVAE}, we show the maximum activating image for each final layer neuron of the capsule, plotted as a flattened torus. We see that the neurons  become arranged according to class, orientation, width, and other learned features.

\subsection{Learning Equivariant Capsules}
\label{sec:equivariant_experiments}
In the remaining experiments, we provide evidence that the Topographic VAE can be leveraged to learn equivariant capsules by incorporating shifting temporal coherence into a 1D baseline topographic model. We compare against two baselines: standard normal VAEs and models that have non-shifting `stationary' temporal coherence as defined in Equation \ref{eqn:temporal_coherence} (denoted `BubbleVAE' \cite{bubbles}).

In all experiments we use a 3-layer MLP with ReLU activations for both encoders and the decoder. We arrange the latent space into 15 circular capsules each of 15-dimensions for dSprites \cite{dSprites17}, and 18 circular capsules each of 18-dimensions for MNIST \cite{lecun2010mnist}. Example sequences $\{\mathbf{x}_l\}_{l=1}^S$ are formed by taking a random initial example, and sequentially transforming it according to one of the available transformations: (X-Pos, Y-Pos, Orientation, Scale) for dSprites, and (Color, Scale, Orientation) for MNIST. All transformation sequences are cyclic such that when the maximum transformation parameter is reached, the subsequent value returns to the minimum. We denote the length of a full transformation sequence by $S$, and the time-extent of the induced temporal coherence (i.e. the length of the input sequence) by $2L$. For simplicity, both datasets are constructed such that the sequence length $S$ equals the capsule dimension (for dSprites this involves taking a subset of the full dataset and looping the scale 3-times for a scale-sequence). Exact details are in Sections \ref{sec:mnist} \& \ref{sec:dsprites}.

\begin{figure}[h!]
\centering
\includegraphics[width=1.0\linewidth]{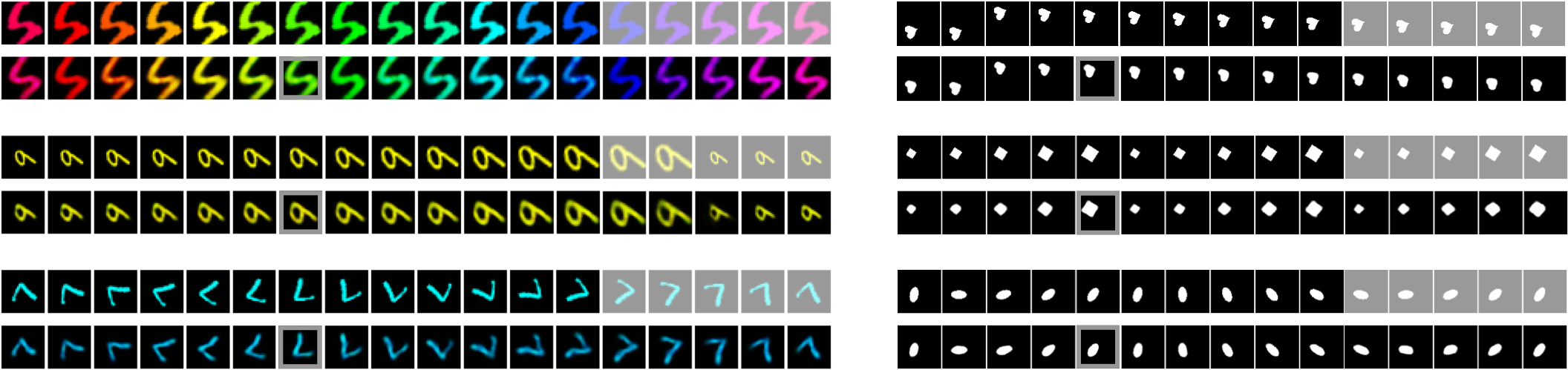}
\caption{Capsule Traversals for TVAE models on dSprites and MNIST. The top rows show the encoded sequences (with greyed-out images held-out), and the bottom rows show the images generated by decoding sequentially $\mathrm{Roll}$ed copies of the initial activation $\mathbf{t}_0$ (indicated by a grey border).} 
\label{fig:all_traversals}
\end{figure}

In Figure \ref{fig:all_traversals}, we show the capsule traversals for TVAE models with {\small$L \approx \frac{1}{3}S$}. We see that despite the $\mathbf{t}_0$ variable encoding only $\frac{2}{3}$ of the sequence, the remainder of the transformation sequence can be decoded nearly perfectly by permuting the activation through the full capsule -- implying the model has learned to be approximately equivariant to full sequences while only observing partial sequences per training point. Furthermore, we see that the model is able to successfully learn all transformations simultaneously for the respective datasets. 
 \\
Capsule traversals for the non-equivariant baselines, as well as TVAEs with smaller values of $L$ (which only learn approximate equivariance to partial sequences) are shown in Section \ref{sec:capsule_traversals}. We note that the capsule traversal plotted in Figure \ref{fig:traversal} demonstrates a transformation where color and rotation change simultaneously, differing from how the models in this section are trained. However, as we describe in more detail in Section \ref{sec:generalization_experiments}, we observe that TVAEs trained with individual transformations in isolation (as in this section) are able to generalize, generating sequences of combined transformations when presented with such partial input sequences at test time. We believe this generalization capability to be promising for data efficiency, but leave further exploration to future work. Additional capsule traversals with such unseen combined transformations are shown in Section \ref{sec:generalization_experiments} and further complex learned transformations (such as perspective transforms) are shown at the end of Section \ref{sec:capsule_traversals}.

For a more quantitative evaluation, in Table \ref{table:mnist} we measure the equivariance error and log-likelihood (reported in nats) of the test data under our trained MNIST models as estimated by importance sampling with 10 samples. We observe that models which incorporate temporal coherence (BubbleVAE and TVAE with $L > 0$) achieve low equivariance error, while the TVAE models with shifting temporal coherence achieve the highest likelihood and the lowest equivariance error simultaneously.

\begin{table}[h!]
    \centering
    \vspace{-4mm}
    \caption{Log Likelihood and Equivariance Error on MNIST for different settings of temporal coherence length $L$ relative to sequence length $S$. Mean $\pm$ std. over 3 random initalizations.}
    \vspace{2mm}
    \begin{tabular}{l r r r r r r}
    \toprule
    Model & TVAE  & TVAE  & TVAE & BubbleVAE & VAE \\ 
    $L$ & $L=\frac{1}{2}S$ & $L=\frac{5}{36}S$ & $L=0$ & $L=\frac{5}{36}S$ &  $L=0$ \\ \midrule
          $\log p(\mathbf{x})$ $\uparrow$ & $\mathbf{-186.8}$ $\pm$ 0.1  & $\mathbf{-186.0}$ $\pm$ 0.7  & -218.5$\pm$ 0.9 & -191.4 $\pm$ 0.5 & -189.0 $\pm$ 0.8 \\
          $\mathcal{E}_{eq}$ $\downarrow$ &  $\mathbf{574}$ $\pm$ 2  & 3247 $\pm$ 3  & 3217 $\pm$ 105 & 3370 $\pm$ 12 & 13274 $\pm$ 1 \\
    \bottomrule
    \end{tabular}
     \label{table:mnist}
\end{table}

\begin{table}[h!]
    \centering
    \caption{Equivariance error ($\mathcal{E}_{eq}$ $\downarrow$) and correlation of observed capsule roll with ground truth factor shift ($\mathrm{CapCorr}$ $\uparrow$) for the dSprites dataset. Mean $\pm$ standard deviation over 3 random initalizations.}
    \vspace{2mm}
    \begin{tabular}{l r r r r r r}
    \toprule
          Model & TVAE & TVAE & TVAE & TVAE & BubbleVAE  & VAE \\ 
          $L$ & $L=\frac{1}{2}S$ & $L = \frac{1}{3}S$ & $L = \frac{1}{6}S$ & $L=0$ & $L=\frac{1}{3}S$ & $L=0$\\
          \midrule
          $\mathrm{CapCorr}_X$ $\uparrow$     & $\mathbf{1.0}$ $\pm$ 0    & $\mathbf{1.0}$ $\pm$ 0  & 0.67 $\pm$ 0.02   & 0.17 $\pm$ 0.03 & 0.13 $\pm$ 0.01 & 0.18 $\pm$ 0.01 \\
          $\mathrm{CapCorr}_Y$ $\uparrow$     & $\mathbf{1.0}$ $\pm$ 0    & $\mathbf{1.0}$ $\pm$ 0  & 0.66 $\pm$ 0.02   & 0.21 $\pm$ 0.02 & 0.12 $\pm$ 0.01 & 0.16 $\pm$ 0.01 \\
          $\mathrm{CapCorr}_{O}$ $\uparrow$   & $\mathbf{1.0}$ $\pm$ 0   & $\mathbf{1.0}$ $\pm$ 0   & 0.52 $\pm$ 0.01   & 0.09 $\pm$ 0.01 & 0.10 $\pm$ 0.01 & 0.11 $\pm$ 0.00 \\
          $\mathrm{CapCorr}_{S}$ $\uparrow$   & $\mathbf{1.0}$ $\pm$ 0   & $\mathbf{1.0}$ $\pm$ 0   & 0.42 $\pm$ 0.01   & 0.51 $\pm$ 0.01 & 0.50 $\pm$ 0.00 & 0.52 $\pm$ 0.00 \\
          \midrule
          $\mathcal{E}_{eq}$ $\downarrow$ & $\mathbf{344}$ $\pm$ 5 & 1034 $\pm$ 6 & 2549 $\pm$ 38 & 2971 $\pm$ 9 & 1951 $\pm$ 34 & 6934 $\pm$ 0\\
    \bottomrule
    \end{tabular}
\label{table:capcorr}
\end{table}

To further understand how capsules transform for observed input transformations, in Table \ref{table:capcorr} we measure $\mathcal{E}_{eq}$ and the $\mathrm{CapCorr}$ metric on the dSprites dataset for the four proposed transformations. We see that the TVAE with $L\geq\frac{1}{3}S$ achieves perfect correlation -- implying the learned representation indeed permutes cyclically within capsules for observed transformation sequences. Further, this correlation gradually decreases as $L$ decreases, eventually reaching the same level as the baselines. We also see that, on both datasets, the equivariance losses for the TVAE with $L=0$ and the BubbleVAE are significantly lower than the baseline VAE, while conversely, the CapCorr metric is not significantly better. We believe this to be due to the fundamental difference between the metrics: $\mathcal{E}_{eq}$ measures continuous L1 similarity which is still low when a representation is locally smooth (even if the change of the representation does not follow the observed transformation), whereas $\mathrm{CapCorr}$ more strictly measures the correspondence between the transformation of the input and the transformation of the representation. In other words, $\mathcal{E}_{eq}$ may be misleadingly low for invariant capsule  representations (as with the BubbleVAE), whereas $\mathrm{CapCorr}$ strictly measures equivariance.

\section{Future Work \& Limitations}
\label{sec:limitations}
\looseness=-1
The model presented in this work has a number of limitations in its existing form which we believe to be interesting directions for future research. Foremost, the model is challenging to compare directly with existing disentanglement and equivariance literature since it requires an input sequence which determines the transformations reachable through the capsule roll.
Related to this, we note the temporal coherence proposed in our model is not `causal' (i.e. $\mathbf{t}_{0}$ depends on future $\mathbf{x}_l$). 
We believe these limitations could be at least partially alleviated with minor extensions detailed in Section \ref{sec:extensions}. 

We additionally note that some model developers may find a priori definition of topographic structure burdensome. While true, we know that the construction of appropriate priors is always a challenging task in latent variable models, and we observe that our proposed TVAE achieves strong performance even with improper specification. Furthermore, in future work, we believe adding learned flexibility to the parameters $\mathbf{W}$ may alleviate some of this burden.

Finally, we note that while this work does demonstrate improved log-likelihood and equivariance error, the study is inherently preliminary and does not examine all important benefits of topographic or approximately equivariant representations. Specifically, further study of the TVAE both with and without temporal coherence in terms of the sample complexity, semi-supervised classification accuracy, and invariance through structured topographic pooling would be enlightening.

\section{Conclusion}
In the above work we introduce the Topographic Variational Autoencoder as a method to train deep topographic generative models, and show how topography can be leveraged to learn approximately equivariant sets of features, a.k.a. capsules, directly from sequences of data with no other supervision. Ultimately, we believe these results may shine some light on how biological systems could hard-wire themselves to more effectively learn representations with equivariant capsule structure.
In terms of broader impact, it is foreseeable our model could be used to generate more realistic transformations of `deepfakes', enhancing disinformation. Given that the model learns \emph{approximate} equivariance, we caution against the over-reliance on equivariant properties as these have no known formal guarantees.

\bibliographystyle{plain}
\bibliography{ref}

\begin{thebibliography}{10}

\bibitem{barlow1961possible}
Horace~B Barlow et~al.
\newblock Possible principles underlying the transformation of sensory
  messages.
\newblock {\em Sensory communication}, 1(01), 1961.

\bibitem{SejnowskiICA}
Anthony~J. Bell and Terrence~J. Sejnowski.
\newblock {An Information-Maximization Approach to Blind Separation and Blind
  Deconvolution}.
\newblock {\em Neural Computation}, 7(6):1129--1159, 11 1995.

\bibitem{bengio2013representation}
Yoshua Bengio, Aaron Courville, and Pascal Vincent.
\newblock Representation learning: A review and new perspectives.
\newblock {\em IEEE transactions on pattern analysis and machine intelligence},
  35(8):1798--1828, 2013.

\bibitem{learninginvariance}
Gregory Benton, Marc Finzi, Pavel Izmailov, and {Andrew Gordon} Wilson.
\newblock Learning invariances in neural networks.
\newblock {\em Advances in Neural Information Processing Systems}, December,
  2020.

\bibitem{wandb}
Lukas Biewald.
\newblock Experiment tracking with weights and biases, 2020.
\newblock Software available from wandb.com.

\bibitem{GTM}
Christopher Bishop, Markus Svensen, and Christopher Williams.
\newblock Gtm: The generative topographic mapping.
\newblock {\em Neural Computation}, 10:215--234, 05 1997.

\bibitem{boenninghoff2020variational}
Benedikt Boenninghoff, Steffen Zeiler, Robert~M. Nickel, and Dorothea Kolossa.
\newblock Variational autoencoder with embedded student-$t$ mixture model for
  authorship attribution.
\newblock {\em ArXiv}, abs/2005.13930, 2020.

\bibitem{bouchacourt2021addressing}
Diane Bouchacourt, Mark Ibrahim, and Stéphane Deny.
\newblock Addressing the topological defects of disentanglement via distributed
  operators.
\newblock {\em ArXiv}, abs/2102.05623, 2021.

\bibitem{cohen2016steerable}
Taco Cohen and M.~Welling.
\newblock Steerable cnns.
\newblock {\em ArXiv}, abs/1612.08498, 2017.

\bibitem{cohen2014learning}
Taco Cohen and Max Welling.
\newblock Learning the irreducible representations of commutative lie groups.
\newblock In {\em Proceedings of the 31st International Conference on Machine
  Learning}, volume~32 of {\em Proceedings of Machine Learning Research}, pages
  1755--1763, Bejing, China, 22--24 Jun 2014. PMLR.

\bibitem{cohen2016group}
Taco Cohen and Max Welling.
\newblock Group equivariant convolutional networks.
\newblock In {\em International conference on machine learning}, pages
  2990--2999, 2016.

\bibitem{cohen2015transformation}
Taco~S. Cohen and Max Welling.
\newblock Transformation properties of learned visual representations.
\newblock In {\em 3rd International Conference on Learning Representations,
  {ICLR} 2015, San Diego, CA, USA, May 7-9, 2015, Conference Track
  Proceedings}, 2015.

\bibitem{comon1994independent}
Pierre Comon.
\newblock Independent component analysis, a new concept?
\newblock {\em Signal processing}, 36(3):287--314, 1994.

\bibitem{connor2021variational}
Marissa Connor, Gregory Canal, and Christopher Rozell.
\newblock Variational autoencoder with learned latent structure.
\newblock In {\em Proceedings of The 24th International Conference on
  Artificial Intelligence and Statistics}, volume 130 of {\em Proceedings of
  Machine Learning Research}, pages 2359--2367. PMLR, 13--15 Apr 2021.

\bibitem{Learningtoconvole}
Nichita Diaconu and Daniel Worrall.
\newblock Learning to convolve: A generalized weight-tying approach.
\newblock In Kamalika Chaudhuri and Ruslan Salakhutdinov, editors, {\em
  Proceedings of the 36th International Conference on Machine Learning},
  volume~97 of {\em Proceedings of Machine Learning Research}, pages
  1586--1595. PMLR, 09--15 Jun 2019.

\bibitem{finzi2020generalizing}
Marc Finzi, Samuel Stanton, Pavel Izmailov, and Andrew~Gordon Wilson.
\newblock Generalizing convolutional neural networks for equivariance to lie
  groups on arbitrary continuous data.
\newblock In {\em Proceedings of the 37th International Conference on Machine
  Learning}, volume 119 of {\em Proceedings of Machine Learning Research},
  pages 3165--3176. PMLR, 13--18 Jul 2020.

\bibitem{finzi2021emlp}
Marc Finzi, Max Welling, and Andrew Gordon~Gordon Wilson.
\newblock A practical method for constructing equivariant multilayer
  perceptrons for arbitrary matrix groups.
\newblock In Marina Meila and Tong Zhang, editors, {\em Proceedings of the 38th
  International Conference on Machine Learning}, volume 139 of {\em Proceedings
  of Machine Learning Research}, pages 3318--3328. PMLR, 18--24 Jul 2021.

\bibitem{foldiak}
Peter Földiák.
\newblock Learning invariance from transformation sequences.
\newblock {\em Neural Computation}, 3:194--200, 06 1991.

\bibitem{Grathwohl16}
Will Grathwohl and Aaron Wilson.
\newblock Disentangling space and time in video with hierarchical variational
  auto-encoders.
\newblock {\em CoRR}, abs/1612.04440, 2016.

\bibitem{higgins2018definition}
Irina Higgins, David Amos, David Pfau, Sebastien Racaniere, Loic Matthey,
  Danilo Rezende, and Alexander Lerchner.
\newblock Towards a definition of disentangled representations.
\newblock {\em ArXiv}, abs/1812.02230, 2018.

\bibitem{capsules2011}
Geoffrey~E. Hinton, Alex Krizhevsky, and Sida~D. Wang.
\newblock Transforming auto-encoders.
\newblock In Timo Honkela, W{\l}odzis{\l}aw Duch, Mark Girolami, and Samuel
  Kaski, editors, {\em Artificial Neural Networks and Machine Learning -- ICANN
  2011}, pages 44--51, Berlin, Heidelberg, 2011. Springer Berlin Heidelberg.

\bibitem{e2018matrix}
Geoffrey~E Hinton, Sara Sabour, and Nicholas Frosst.
\newblock Matrix capsules with {EM} routing.
\newblock In {\em International Conference on Learning Representations}, 2018.

\bibitem{hinton2013discovering}
Geoffrey~E. Hinton and Yee-Whye Teh.
\newblock Discovering multiple constraints that are frequently approximately
  satisfied.
\newblock In {\em Proceedings of the Seventeenth Conference on Uncertainty in
  Artificial Intelligence}, UAI'01, page 227–234, 2001.

\bibitem{hurri2003TC}
Jarmo Hurri and Aapo Hyvärinen.
\newblock {Simple-Cell-Like Receptive Fields Maximize Temporal Coherence in
  Natural Video}.
\newblock {\em Neural Computation}, 15(3):663--691, 03 2003.

\bibitem{bubbles}
A.~Hyv{\"a}rinen, J.~Hurri, and Jaakko~J. V{\"a}yrynen.
\newblock A unifying framework for natural image statistics: spatiotemporal
  activity bubbles.
\newblock {\em Neurocomputing}, 58-60:801--806, 2004.

\bibitem{hyvarinen2000emergence}
Aapo Hyv{\"a}rinen and Patrik Hoyer.
\newblock \href{shorturl.at/aboqY}{Emergence of phase-and shift-invariant
  features by decomposition of natural images into independent feature
  subspaces}.
\newblock {\em Neural computation}, 12(7):1705--1720, 2000.

\bibitem{hyvarinen2001topographic}
Aapo Hyv{\"a}rinen, Patrik~O Hoyer, and Mika Inki.
\newblock
  \href{https://www.cs.helsinki.fi/u/ahyvarin/papers/NC01_TICA.pdf}{Topographic
  independent component analysis}.
\newblock {\em Neural computation}, 13(7):1527--1558, 2001.

\bibitem{hyvarinen2009natural}
Aapo Hyv{\"a}rinen, Jarmo Hurri, and Patrick~O Hoyer.
\newblock {\em Natural image statistics: A probabilistic approach to early
  computational vision.}, volume~39.
\newblock Springer Science \& Business Media, 2009.

\bibitem{hyvarinen2000independent}
Aapo Hyv{\"a}rinen and Erkki Oja.
\newblock
  \href{https://www.cs.helsinki.fi/u/ahyvarin/papers/bookfinal_ICA.pdf}{Independent
  component analysis: algorithms and applications}.
\newblock {\em Neural networks}, 13(4-5):411--430, 2000.

\bibitem{HYVARINEN20012413}
Aapo Hyvärinen and Patrik~O. Hoyer.
\newblock A two-layer sparse coding model learns simple and complex cell
  receptive fields and topography from natural images.
\newblock {\em Vision Research}, 41(18):2413--2423, 2001.

\bibitem{kavukcuoglu2009learning}
Koray Kavukcuoglu, Marc'Aurelio Ranzato, Rob Fergus, and Yann LeCun.
\newblock
  \href{https://ieeexplore.ieee.org/abstract/document/5206545?casa_token=-kv1vXY9bvEAAAAA:N0MAAbC8zxqH8ZG4lhjrx8SV0kpw-xsYnzdwcAqF1JXIT2KqDXqXRDF1Y5qBFH1vOKUhTygHmmY}{Learning
  invariant features through topographic filter maps}.
\newblock In {\em 2009 IEEE Conference on Computer Vision and Pattern
  Recognition}, pages 1605--1612. IEEE, 2009.

\bibitem{keller2021modeling}
T.~Anderson Keller, Qinghe Gao, and Max Welling.
\newblock Modeling category-selective cortical regions with topographic
  variational autoencoders.
\newblock In {\em SVRHM 2021 Workshop @ NeurIPS}, 2021.

\bibitem{kingma2013auto}
Diederik~P. Kingma and Max Welling.
\newblock {Auto-Encoding Variational Bayes}.
\newblock In {\em 2nd International Conference on Learning Representations,
  {ICLR} 2014, Banff, AB, Canada, April 14-16, 2014, Conference Track
  Proceedings}, 2014.

\bibitem{klindt2021nonlinear}
David~A. Klindt, Lukas Schott, Yash Sharma, Ivan Ustyuzhaninov, Wieland
  Brendel, Matthias Bethge, and Dylan Paiton.
\newblock Towards nonlinear disentanglement in natural data with temporal
  sparse coding.
\newblock In {\em International Conference on Learning Representations}, 2021.

\bibitem{kohonen1996emergence}
Teuvo Kohonen.
\newblock
  \href{https://link.springer.com/content/pdf/10.1007/s004220050295.pdf}{Emergence
  of invariant-feature detectors in the adaptive-subspace self-organizing map}.
\newblock {\em Biological cybernetics}, 75(4):281--291, 1996.

\bibitem{koulakov2001orientation}
Alexei~A Koulakov and Dmitri~B Chklovskii.
\newblock Orientation preference patterns in mammalian visual cortex: a wire
  length minimization approach.
\newblock {\em Neuron}, 29(2):519--527, 2001.

\bibitem{lecun2010mnist}
Yann LeCun, Corinna Cortes, and CJ~Burges.
\newblock Mnist handwritten digit database.
\newblock {\em ATT Labs [Online]. Available: http://yann.lecun.com/exdb/mnist},
  2, 2010.

\bibitem{TDANN}
Hyodong Lee, Eshed Margalit, Kamila~M. Jozwik, Michael~A. Cohen, Nancy
  Kanwisher, Daniel L.~K. Yamins, and James~J. DiCarlo.
\newblock Topographic deep artificial neural networks reproduce the hallmarks
  of the primate inferior temporal cortex face processing network.
\newblock {\em bioRxiv}, 07/2020 2020.

\bibitem{lenssen2018group}
Jan~Eric Lenssen, Matthias Fey, and Pascal Libuschewski.
\newblock Group equivariant capsule networks.
\newblock In {\em NeurIPS}, pages 8858--8867, 2018.

\bibitem{Lyu08b}
S~Lyu and E~P Simoncelli.
\newblock Nonlinear image representation using divisive normalization.
\newblock In {\em Proc. Computer Vision and Pattern Recognition}, pages 1--8.
  IEEE Computer Society, Jun 23-28 2008.

\bibitem{Lyu08}
S~Lyu and E~P Simoncelli.
\newblock Modeling multiscale subbands of photographic images with fields of
  {Gaussian} scale mixtures.
\newblock {\em IEEE Trans. Patt. Analysis and Machine Intelligence},
  31(4):693--706, Apr 2009.

\bibitem{ma2008overcomplete}
Libo Ma and Liqing Zhang.
\newblock
  \href{https://reader.elsevier.com/reader/sd/pii/S0925231208000957?token=FEFABF17DBC9D13F8F221FA7EC2F1AE1F07AA22B11FFEE0E90383CF39C696F0DADBE29276AC1CD7FF6F2F13E63846C99}{Overcomplete
  topographic independent component analysis}.
\newblock {\em Neurocomputing}, 71(10-12):2217--2223, 2008.

\bibitem{dSprites17}
Loic Matthey, Irina Higgins, Demis Hassabis, and Alexander Lerchner.
\newblock dsprites: Disentanglement testing sprites dataset.
\newblock https://github.com/deepmind/dsprites-dataset/, 2017.

\bibitem{olshausen1997sparse}
Bruno~A Olshausen and David~J Field.
\newblock \href{shorturl.at/byGPV}{Sparse coding with an overcomplete basis
  set: A strategy employed by V1?}
\newblock {\em Vision research}, 37(23):3311--3325, 1997.

\bibitem{Osindero2006}
Simon Osindero, Max Welling, and Geoffrey~E. Hinton.
\newblock {Topographic Product Models Applied to Natural Scene Statistics}.
\newblock {\em Neural Computation}, 18(2):381--414, 02 2006.

\bibitem{osindero2004contrastive}
Simon~Kayode Osindero.
\newblock {\em Contrastive Topographic Models}.
\newblock PhD thesis, University of London, 2004.

\bibitem{pytorch}
Adam Paszke, Sam Gross, Francisco Massa, Adam Lerer, James Bradbury, Gregory
  Chanan, Trevor Killeen, Zeming Lin, Natalia Gimelshein, Luca Antiga, Alban
  Desmaison, Andreas Kopf, Edward Yang, Zachary DeVito, Martin Raison, Alykhan
  Tejani, Sasank Chilamkurthy, Benoit Steiner, Lu~Fang, Junjie Bai, and Soumith
  Chintala.
\newblock Pytorch: An imperative style, high-performance deep learning library.
\newblock In {\em Advances in Neural Information Processing Systems 32}, pages
  8024--8035. 2019.

\bibitem{Portilla03}
J~Portilla, V~Strela, M~J Wainwright, and E~P Simoncelli.
\newblock Image denoising using scale mixtures of {Gaussians} in the wavelet
  domain.
\newblock {\em IEEE Trans Image Processing}, 12(11):1338--1351, Nov 2003.
\newblock Recipient, IEEE Signal Processing Society Best Paper Award, 2008.

\bibitem{ravanbakhsh2017equivariance}
Siamak Ravanbakhsh, Jeff Schneider, and Barnab{\'a}s P{\'o}czos.
\newblock Equivariance through parameter-sharing.
\newblock In Doina Precup and Yee~Whye Teh, editors, {\em Proceedings of the
  34th International Conference on Machine Learning}, volume~70 of {\em
  Proceedings of Machine Learning Research}, pages 2892--2901. PMLR, 06--11 Aug
  2017.

\bibitem{rezende2014stochastic}
Danilo~Jimenez Rezende, Shakir Mohamed, and Daan Wierstra.
\newblock Stochastic backpropagation and approximate inference in deep
  generative models.
\newblock In {\em Proceedings of the 31st International Conference on
  International Conference on Machine Learning - Volume 32}, ICML'14, page
  II–1278–II–1286. JMLR.org, 2014.

\bibitem{sabour2017dynamic}
Sara Sabour, Nicholas Frosst, and Geoffrey~E. Hinton.
\newblock Dynamic routing between capsules.
\newblock In {\em Proceedings of the 31st International Conference on Neural
  Information Processing Systems}, NIPS'17, page 3859–3869, Red Hook, NY,
  USA, 2017. Curran Associates Inc.

\bibitem{stone1996}
James~V. Stone.
\newblock {Learning Perceptually Salient Visual Parameters Using Spatiotemporal
  Smoothness Constraints}.
\newblock {\em Neural Computation}, 8(7):1463--1492, 10 1996.

\bibitem{stuhmer2019independent}
Jan Stühmer, Richard~E. Turner, and Sebastian Nowozin.
\newblock Independent subspace analysis for unsupervised learning of
  disentangled representations, 2019.

\bibitem{probSFA}
Richard Turner and Maneesh Sahani.
\newblock A maximum-likelihood interpretation for slow feature analysis.
\newblock {\em Neural computation}, 19:1022--38, 05 2007.

\bibitem{elise}
Elise van~der Pol, Daniel~E. Worrall, Herke van Hoof, Frans~A. Oliehoek, and
  Max Welling.
\newblock {MDP} homomorphic networks: Group symmetries in reinforcement
  learning.
\newblock {\em CoRR}, abs/2006.16908, 2020.

\bibitem{Wainwright99b}
M~J Wainwright and E~P Simoncelli.
\newblock Scale mixtures of {Gaussians} and the statistics of natural images.
\newblock In S.~A. Solla, T.~K. Leen, and K.-R. {M\"{u}ller}, editors, {\em
  Adv. Neural Information Processing Systems (NIPS*99)}, volume~12, pages
  855--861, Cambridge, MA, May 2000. MIT Press.

\bibitem{Wainwright00}
M~J Wainwright, E~P Simoncelli, and A~S Willsky.
\newblock Random cascades on wavelet trees and their use in analyzing and
  modeling natural images.
\newblock {\em Applied and Computational Harmonic Analysis}, 11(1):89--123, Jul
  2001.

\bibitem{weiler20183d}
Maurice Weiler, Mario Geiger, Max Welling, Wouter Boomsma, and Taco Cohen.
\newblock 3d steerable cnns: Learning rotationally equivariant features in
  volumetric data.
\newblock In {\em Proceedings of the 32nd International Conference on Neural
  Information Processing Systems}, NIPS'18, page 10402–10413, Red Hook, NY,
  USA, 2018. Curran Associates Inc.

\bibitem{welling2003learning}
Max Welling, Simon Osindero, and Geoffrey~E Hinton.
\newblock
  \href{http://papers.nips.cc/paper/2177-learning-sparse-topographic-representations-with-products-of-student-t-distributions.pdf}{Learning
  sparse topographic representations with products of student-t distributions}.
\newblock In {\em Advances in neural information processing systems}, pages
  1383--1390, 2003.

\bibitem{wiskott2002slow}
Laurenz Wiskott and Terrence~J Sejnowski.
\newblock Slow feature analysis: Unsupervised learning of invariances.
\newblock {\em Neural computation}, 14(4):715--770, 2002.

\bibitem{scalespaces}
Daniel Worrall and Max Welling.
\newblock Deep scale-spaces: Equivariance over scale.
\newblock In H.~Wallach, H.~Larochelle, A.~Beygelzimer, F.~d\textquotesingle
  Alch\'{e}-Buc, E.~Fox, and R.~Garnett, editors, {\em Advances in Neural
  Information Processing Systems}, volume~32. Curran Associates, Inc., 2019.

\bibitem{worrall2017harmonic}
Daniel~E. Worrall, Stephan~J. Garbin, Daniyar Turmukhambetov, and Gabriel~J.
  Brostow.
\newblock Harmonic networks: Deep translation and rotation equivariance.
\newblock {\em 2017 IEEE Conference on Computer Vision and Pattern Recognition
  (CVPR)}, pages 7168--7177, 2017.

\end{thebibliography}

\section{Acknowledgements}
We would like to thank Jorn Peters for his invaluable contributions to this work at its earliest stages. We would additionally like to thank  Patrick Forr\'{e}, Emiel Hoogeboom, and Anna Khoreva for their helpful guidance throughout the project. We would like to thank the creators of Weight \& Biases \cite{wandb} and PyTorch \cite{pytorch}. Without these tools our work would not have been possible. Finally, we thank the Bosch Center for Artificial Intelligence for funding, and the reviewers for their helpful comments.

\newpage

\appendix

\section{Experiment Details}
The code for reproducing all experiments in this paper can be found in the following GitHub repository: \url{https://github.com/AKAndyKeller/TopographicVAE}

\subsection{Optimizer Parameters} 
Given the differences between the training procedures of the model presented in Section 6.2, and those in Section 6.3, the optimizer parameters for the two settings differed slightly. The 2D Topographic VAE without Temporal Coherence presented in Figure 3 was trained with stochastic gradient descent on batches of size $128$, using a learning rate of $1 \times 10^{-4}$, and standard momentum of $0.9$ for 250 epochs. All models in Section 6.3 were trained with stochastic gradient descent on batches of size $8$ (due to each batch-example being a length 15 or 18 sequence), using a learning rate of $1 \times 10^{-4}$, and standard momentum of $0.9$ for 100 epochs.

\subsection{Initalization} All weights of the models were initialized with uniformly random samples from $U(-\frac{1}{\sqrt{m}}, \frac{1}{\sqrt{m}})$, where $m$ is the number of input units. For all topographic models including BubbleVAE, $\mu$ was initialized to a large value (30.0) as this was observed to increase the speed of convergence and was sometimes necessary for observed topographic organization in deeper models. For the 2D topographic model in Figure \ref{fig:TVAE}, $\mu$ was initialized to 10.

\subsection{Model Architectures}
All models presented in this paper make use of the same 3-Layer MLP for parameterizing the encoders and decoders. Specifically, the model is constructed as 3 fully connected layers with ReLU activations in-between the layers. For MNIST, the layers of both the $\mathbf{u}$ and $\mathbf{z}$ encoders have (972, 648, 648) output units each for the first, second, and third layers respectively. The 648 units in the third layer are divided into two sets to compute the mean and log standard deviation of the respective $u$'s and $z$'s, yielding 324 $t$ variables. This is then divided into 18 capsules, each of 18 dimensions. The layers of the decoder have (648, 972, 2352) output units respectively. For dSprites, both encoder layers have output sizes (674, 450, 450), where the resulting 225 $t$ variables are divided into 15 capsules, each of 15 dimensions. The decoder layers then have output sizes (450, 675, 4096). We note the non-topographic VAE baselines make use of only a single encoder for the Gaussian variable $\mathbf{z}$ (as $\mathbf{u}$ is not needed), and do not incorporate a $\mu$ parameter.

\subsection{Choice of $\mathbf{W}$}
For all topographic models (TVAE and BubbleVAE) in Section 6.3, the global topographic organization afforded by $\mathbf{W}$ was fixed to a set of 1-D tori (`circular capsules') as depicted in Figure 1. The model presented in Section 6.2 organizes its variables as a single 2-D torus. Practically, multiplication by $\mathbf{W}$ was performed by convolution over the appropriate dimensions (time \& capsule dimension) with a kernel of all $1$'s, taking advantage of circular padding to achieve toroidal structure.

\subsection{Choice of $\mathbf{W}_{\delta}$} The choice of $\mathbf{W}_{\delta}$ determines the local topographic structure within a single timestep. For all TVAE models with $L>0$, we experimented with local neighborhood sizes (denoted $K$) of 3 units (effective kernel size 3 in the capsule dimension), and 1 unit (no neighborhood). For MNIST it was observed that $K=3$ performed best, while $K=1$ worked best for dSprites. This is likely due to the slower, smoother, and more overlapping transformations constructed on MNIST, whereas our subset of dSprites contained non-smooth transformations where the overlap between successive images was smaller (e.g. due to sub-setting, see Section \ref{sec:dsprites}), which made larger neighborhood sizes $K>1$ less fitting. 
For TVAE models with $L=0$,  $\mathbf{W}_{\delta} = \mathbf{W}$ was fixed to sum over neighborhoods of size $K=9$ for MNIST and $K=3$ for dSprites. These values were chosen to be sufficiently large to achieve notably lower equivariance error than the VAE baseline, and thus demonstrate the impact of topographic organization without temporal coherence. 
For BubbleVAE models, the extent of topographic organization in the capsule dimension was set to $K=3$ on MNIST to match the TVAE, and was set to be equal to the organization in time dimension $K=2L$ for dSprites. A further quantitative comparison on the impact of the choice of the $K$ parameter can be found in Section \ref{sec:k_impact}.

\subsection{Choice of $L$} The choice of $L$ determines the extent of temporal coherence where $2L$ equals the input sequence length, and $L=0$ corresponds to single inputs. For Table 1, we experimented with values of $L$ in the set $\{0, \frac{5}{36}S, \frac{1}{4}S, \frac{1}{2}S \}$ for both the TVAE and BubbleVAE. Both the BubbleVAE and TVAE achieved highest likelihoods at $L=\frac{5}{36}S$, and TVAE achieved lowest equivariance error at $L=\frac{1}{2}S$. We additionally included TVAE experiments with $L=\frac{13}{36}S$ for purposes of visualization in Figures 1 and 4 as this yielded the best qualitative generalization. For Table 2, we experimented with values of $L$ in the set $\{0, \frac{1}{6}S, \frac{4}{15}S, \frac{1}{3}S, \frac{2}{5}S, \frac{1}{2}S\}$ for both TVAE and BubbleVAE, and presented a broad selection in the table. The results of all models are shown in Section \ref{sec:extended_results} below.

\subsection{Hyperparameter Selection}
Hyperparameters such as learning rate, batch size, number of capsules, capsule size, and ultimately model architecture were chosen to allow for quick training on limited resources and were not tuned significantly. Since it was conceptually simpler to have an equal number of capsule dimensions and sequence elements, this limited the number of capsules we could then train efficiently. In Section \ref{sec:roll_extension} we explain how a model with fewer capsule dimensions than sequence elements could be constructed with an alternative $\mathrm{Roll}$ operator. Additionally, from preliminary experiments, we observe that models with a number of internal capsule dimensions different from the number of sequence elements achieve similar likelihood values while also learning coherent transformations as decoded through the capsule roll. We believe these findings in combination with the extra studies provided in Section \ref{sec:extended_results} suggest a satisfying degree of robustness to hyperparameter selection.

\subsection{MNIST Transformations}
\label{sec:mnist}
\looseness=-1
The first set of experiments presented in this paper are based on the MNIST dataset \cite{lecun2010mnist} (MIT Licence). For Section 6.2 (Figure 3) an MNIST training set of 48,000 images was used, while the standard test set of 10,000 images was used to compute the maximum activating image. For Section 6.3 (Figure 4 and Table 1), sequences of MNIST images were created by picking a random training image (with a random transformation `pose') and successively transforming it according to one of the 3 available transformations (e.g. only one attribute is changed per sequence). The available transformations consisted of rotation, color (hue rotation), and scale with increments of 20-degrees for rotation and color, and $3.66\%$ increments for scale. Since scale is inherently non-cyclic, the bounds of the transformation were set at $60\%$ and $126\%$, and the transformations were constructed to be periodic such then once scale reached $126\%$, the next element was at $60\%$ scale. The final sequences were thus constructed to be 18 images long, where each element in the batch had an independently randomly chosen transformation. Again, the likelihood $\log p(\mathbf{x})$ and equivariance error $\mathcal{E}_{eq}$ were computed on the held-out 10,000 example test set, where the same random transformation sequences were applied. 

\subsection{dSprites Transformations}
\label{sec:dsprites}
\looseness=-1
The second set of experiments presented in this paper are based on the dSprites dataset \cite{dSprites17} (Apache-2.0 License). To reduce computational complexity of this dataset, we took a subset of the dataset which consisted of all 3 shapes, the largest 5 scales, and every other example from the first 30 orientations, x-positions, and y-positions. The resulting dataset thus had 50,625 total images (3 shapes, 5 scales, 15 orientations, 15 x-positions, 15 y-positions), compared to the original 737,280 images. To construct sequences, we followed the same procedure as for MNIST, whereby first a random example and transformation were chosen, and a sequence of 15 images was constructed where only the chosen transformation was applied successively. We define the transformations available for sequences as scale, orientation, x-position, and y-position, omitting shape since smooth shape transforms are not present in the dSprites dataset. Again, we define all transformations to be cyclic such that once the 15th element is reached, the 1st element follows. For scale transformations, we simply loop over all 5 scales 3 times per sequence. We observe that although these sequences do not match the latent priors exactly, the models still train relatively well, implying some degree of robustness. 

\subsection{Capsule Correlation Metric ($\mathrm{CapCorr}$)} 
\label{sec:appendix_capcorr}
Here we define $\mathrm{CapCorr}$ more precisely as it is implemented in our work. First, we denote the ground truth transformation parameter of the sequence at timestep $l$ as $y_l$ (e.g. the rotation angle at timestep $l$ for a rotation sequence), and the corresponding activation at time $l$ as $\mathbf{t}_l$. Next, to get an arbitrary starting point, we let $l=\Omega$ denote the timestep when $y_l$ is at its canonical position (e.g. rotation angle 0, x-position 0, or scale 1). We see $\Omega$ is not necessarily $0$ since the first timestep of each sequence ($l=0$) is a randomly transformed example. Then, we observe that we can measure the approximate observed roll in the capsule dimension between time $0$ and $\Omega$ as a `phase shift' by computing the index of the maximum value of a discrete (periodic) cross-correlation of $\mathbf{t}_\Omega$ and $\mathbf{t}_0$:
\begin{equation}
    \mathrm{ObservedRoll}(\mathbf{t}_{\Omega}, \mathbf{t}_{0}) = \mathrm{argmax}\left[\mathbf{t}_{\Omega} \star \mathbf{t}_{0}\right]
\end{equation}
Where $\star$ is discrete (periodic) cross-correlation across the (cyclic) capsule dimension and $\mathrm{argmax}$ is also subsequently performed over the capsule dimension. Then, the $\mathrm{CapCorr}$ metric for a single capsule is given as:
\begin{equation}
    \mathrm{CapCorr}(\mathbf{t}_{\Omega}, \mathbf{t}_{0}, y_{\Omega},  y_{0}) = \mathrm{Corr} \left(\mathrm{ObservedRoll}(\mathbf{t}_{\Omega}, \mathbf{t}_{0}), |y_{\Omega} - y_{0}|\right)
    \label{eqn:capcorr_extended}
\end{equation}
Where the correlation coefficient $\mathrm{Corr}$ is then computed across all examples for the entire dataset. In our experiments we use the Pearson correlation coefficient for $\mathrm{Corr}$. We thus see this metric is the correlation of the estimated observed capsule roll with the shift in ground truth generative factors, which is equal to $1$ when the model is perfectly equivariant. To extend this definition to multiple capsules, we estimate $\mathrm{ObservedRoll}$ for each capsule separately, and then correlate the mode of all $\mathrm{ObservedRoll}$ values with the true shift in ground truth generative factors. We see empirically that the $\mathrm{ObservedRoll}$s for all capsules are almost always identical (i.e. all capsules roll simultaneously for each transformation), therefore computing the mode does not destroy significant information. Finally, for transformation sequences which have multiple timesteps where $y_l$ is at the canonical position (e.g. scale transformations on dSprites where scale is looped 3 times), we select $l=\Omega$ to be the one from this possible set which yields the minimal absolute distance between $|y_{\Omega} - y_{0}|$ and $\mathrm{ObservedRoll}(\mathbf{t}_{\Omega}, \mathbf{t}_{0})$.

\subsection{Definition of $\mathrm{Roll}$ for Capsules}
\label{sec:roll_def}
As stated in Section 4.5.2, $\mathrm{Roll}_{\delta}(\mathbf{u})$, is defined as a cyclic permutation of $\delta$ steps along the capsule dimension of $\mathbf{u}$. Explicitly, if $\mathbf{u}$ is divided into $C$ capsules each with $D$ dimensions, the $\mathrm{Roll}_{\delta}$ operation can be written as:
\begin{align}
    \mathrm{Roll}_{\delta}(\mathbf{u}) & = \mathrm{Roll}_{\delta}\left(\left[u_1, u_2, \ldots, u_{C \cdot D} \right]\right) \nonumber \\
    & = \left[u_D, u_1, \ldots, u_{D-1}, u_{2 \cdot D}, u_{D+1}, \ldots, u_{2 \cdot D-1}, u_{3 \cdot D}, \ \ \ldots\ \ ,\  \ldots \ \  u_{C \cdot D-1} \right]
\end{align}

\section{Extended Results}
\label{sec:extended_results}
In this section we provide extended results for all tested hyperparamters (Tables \ref{table:mnist_extended} \& \ref{table:capcorr_extended}), a further analysis of the impact of the coherence window within a capsule $\mathbf{W}_{\delta}$ (Table \ref{table:choice_of_k}), samples from the model in Section \ref{sec:2d_TVAE}, and additional capsule traversal experiments highlighting the generalization capabilities of the TVAE to combinations of transformations unseen during training (Figure \ref{fig:generalization}).

\subsection{Extended Tables 1 \& 2}
In Tables \ref{table:mnist_extended} \& \ref{table:capcorr_extended} below, we present extended versions of Tables 1 \& 2 respectively, showing all tested settings of the TVAE \& BubbleVAE. 
We observe the TVAE achieves perfect correlation ($\mathrm{CapCorr}=1$) for $L \geq \frac{1}{3}$, and steadily decreasing correlation for lower values of $L$. 
\begin{table}[]
    \caption{Log Likelihood and Equivariance Error on MNIST for all models tested. Mean $\pm$ std. over 3 random initalizations.}
    \vspace{2mm}
    \begin{tabular}{l r r r r r r}
    \toprule
    Model & TVAE & TVAE & TVAE & TVAE & TVAE \\ 
    $L$ & $L=\frac{1}{2}S$ & $L=\frac{13}{36}S$ & $L=\frac{1}{4}S$ & $L=\frac{5}{36}S$ & $L=0$  \\ 
    $K$ & $K=3$ & $K=3$ & $K=3$ & $K=3$ & $K=9$ \\ 
    \midrule
          $\log p(\mathbf{x})$ $\uparrow$ & $\mathbf{-186.8}$ $\pm$ 0.1            &  -188.0 $\pm$ 0.5 & -187.0 $\pm$ 0.2   &  $\mathbf{-186.0}$ $\pm$ 0.7 & -218.5 $\pm$ 0.9     \\
          $\mathcal{E}_{eq}$ $\downarrow$ & $\mathbf{573.9}$ $\pm$ 1.5  &   1089.8 $\pm$ 2.4 & 2136.9 $\pm$ 7.8  &  3246.6 $\pm$ 3.3          & 3216.6 $\pm$ 104.9  \\
    \bottomrule\\
    \toprule
    Model & BubbleVAE  & BubbleVAE & BubbleVAE & BubbleVAE & VAE \\ 
    $L$ & $L=\frac{1}{2}S$ & $L=\frac{1}{4}S$ & $L=\frac{5}{36}S$ & $L=\frac{5}{36}S$ & $L=0$\\
    $K$ & $K=2L$ & $K=2L$ & $K=2L$ & $K=3$ & $K=1$\\
    \midrule
          $\log p(\mathbf{x})$ $\uparrow$  & -200.9 $\pm$ 0.7    & -202.3 $\pm$ 1.4   & -190.8 $\pm$ 0.7 & -191.4 $\pm$ 0.5 &  -189.0 $\pm$ 0.8 \\
          $\mathcal{E}_{eq}$ $\downarrow$  & 4206.7 $\pm$ 903.3  & 1141.7 $\pm$ 9.6  & 2605.7 $\pm$ 16.1 & 3369.5 $\pm$ 11.9  &  13273.9 $\pm$ 0.5 \\
    \bottomrule
    \end{tabular}
     \label{table:mnist_extended}
\end{table}
\begin{table}[]
    \centering
    \caption{Equivariance error and $\mathrm{CapCorr}$ for all models tested on the dSprites dataset. Mean $\pm$ standard deviation over 3 random initalizations.}
    \vspace{2mm}
    \begin{tabular}{l r r r r r r}
    \toprule
          Model & TVAE & TVAE & TVAE & TVAE & TVAE & TVAE \\ 
          $L$                        & $L=\frac{1}{2}S$         & $L=\frac{2}{5}S$         & $L = \frac{1}{3}S$       & $L=\frac{4}{15}S$ & $L=\frac{1}{6}S$  & $L=0$\\
          $K$ & $K=1$ & $K=1$ & $K=1$ & $K=1$ & $K=1$ & $K=3$\\
          \midrule
          $\mathrm{CapCorr}_X$ $\uparrow$   & $\mathbf{1.0}$ $\pm$ 0   & $\mathbf{1.0}$ $\pm$ 0   & $\mathbf{1.0}$ $\pm$ 0   & 0.95 $\pm$ 0.00   & 0.67 $\pm$ 0.02   & 0.17 $\pm$ 0.03\\
          $\mathrm{CapCorr}_Y$ $\uparrow$   & $\mathbf{1.0}$ $\pm$ 0   & $\mathbf{1.0}$ $\pm$ 0   & $\mathbf{1.0}$ $\pm$ 0   & 0.96 $\pm$ 0.01   & 0.66 $\pm$ 0.02   & 0.21 $\pm$ 0.02\\
          $\mathrm{CapCorr}_{O}$ $\uparrow$ & $\mathbf{1.0}$ $\pm$ 0   & $\mathbf{1.0}$ $\pm$ 0   & $\mathbf{1.0}$ $\pm$ 0   & 0.88 $\pm$ 0.01   & 0.52 $\pm$ 0.01   & 0.09 $\pm$ 0.01\\
          $\mathrm{CapCorr}_{S}$ $\uparrow$ & $\mathbf{1.0}$ $\pm$ 0   & $\mathbf{1.0}$ $\pm$ 0   & $\mathbf{1.0}$ $\pm$ 0   & 0.96 $\pm$ 0.01   & 0.42 $\pm$ 0.01   & 0.51 $\pm$ 0.01\\
          \midrule
          $\mathcal{E}_{eq}$ $\downarrow$   & $\mathbf{344}$ $\pm$ 5    & 759 $\pm$ 9              & 1034 $\pm$ 6              & 1395 $\pm$ 7     & 2549 $\pm$ 38     & 2971 $\pm$ 9   \\
    \bottomrule
    \\
    \toprule
          Model & BubbleVAE & BubbleVAE & BubbleVAE & BubbleVAE & BubbleVAE & VAE \\ 
          $L$ &                        $L=\frac{1}{2}S$   & $L=\frac{2}{5}S$   & $L = \frac{1}{3}S$  & $L=\frac{4}{15}S$ & $L=\frac{1}{6}S$ & $L=0$\\
          $K$ & $K=2L$ & $K=2L$ & $K=2L$ & $K=2L$ & $K=2L$ & $K=1$\\
          \midrule
          $\mathrm{CapCorr}_X$ $\uparrow$   & 0.16 $\pm$ 0.01   & 0.15 $\pm$ 0.01    & 0.13 $\pm$ 0.01   & 0.12 $\pm$ 0.02   & 0.09 $\pm$ 0.01   & 0.18 $\pm$ 0.01 \\
          $\mathrm{CapCorr}_Y$ $\uparrow$   & 0.15 $\pm$ 0.01   & 0.14 $\pm$ 0.01    & 0.12 $\pm$ 0.01   & 0.12 $\pm$ 0.01   & 0.11 $\pm$ 0.02   & 0.16 $\pm$ 0.01 \\
          $\mathrm{CapCorr}_{O}$ $\uparrow$ & 0.12 $\pm$ 0.00   & 0.13 $\pm$ 0.02    & 0.10 $\pm$ 0.01   & 0.09 $\pm$ 0.00   & 0.06 $\pm$ 0.01   & 0.11 $\pm$ 0.00 \\
          $\mathrm{CapCorr}_{S}$ $\uparrow$ & 0.52 $\pm$ 0.02   & 0.55 $\pm$ 0.00    & 0.52 $\pm$ 0.00   & 0.48 $\pm$ 0.02   & 0.27 $\pm$ 0.01   & 0.52 $\pm$ 0.00 \\
          \midrule
          $\mathcal{E}_{eq}$ $\downarrow$   & 6825 $\pm$ 126    & 6917 $\pm$ 13      & 1951 $\pm$ 34       & 2181 $\pm$ 627       & 1721 $\pm$ 27     & 6934 $\pm$ 0\\
    \bottomrule
    \end{tabular}
\label{table:capcorr_extended}
\end{table}

\subsection{Impact of $\mathbf{W}_{\delta}$}
\label{sec:k_impact}
In Table \ref{table:choice_of_k}, we show a small set of experiments with different settings of $\mathbf{W}_{\delta}$, and specifically changing values of $K$ (the coherence window within a capsule). As can be seen, increasing $K$ generally reduces equivariance error, but decreases the log-likelihood. This can be further understood by examining the capsule traversals of such models in Figures \ref{fig:tvae_L5d36_k3}, \ref{fig:tvae_L5d36_k9}, \ref{fig:tvae_L0_k3}, \ref{fig:tvae_L0_k9}, \& \ref{fig:tvae_L0_k18}. We see that larger values of $K$ appear to induce smoother transformations within the capsule dimensions, eventually resulting in invariant representations when $K$ is equal to the capsule dimensionality. 

\begin{table}[]
    \centering
    \caption{Impact of $\mathbf{W}_{\delta}$ (i.e. $K$) on MNIST performance.}
    \vspace{2mm}
    \begin{tabular}{l r r r r r r}
    \toprule
    Model & TVAE & TVAE & TVAE & TVAE & TVAE\\ 
    $L$ & $L=\frac{5}{36}S$  & $L=\frac{5}{36}S$ & $L=0$ & $L=0$ & $L=0$\\
    $K$ & $K=3$ & $K=9$ & $K=3$ & $K=9$ & $K=18$\\ 
    \midrule
          $\log p(\mathbf{x})$ $\uparrow$  &  $\mathbf{-186.0}$ $\pm$ 0.7 & -190.6 $\pm$ 0.2 & -213.4 $\pm$ 1.2 & -218.5 $\pm$ 0.9  & -224.8 $\pm$ 1.0 \\
          $\mathcal{E}_{eq}$ $\downarrow$  &  3246.6 $\pm$ 3.3            & 2606.3 $\pm$ 17.0 & 12085.7 $\pm$ 68.5 & 3216.6 $\pm$ 104.9 & 1090.3 $\pm$ 19.3 \\
    \bottomrule
    \end{tabular}
     \label{table:choice_of_k}
\end{table}

\subsection{Samples}
\begin{wrapfigure}{r}{0.3\linewidth}
\vspace{-10mm}
\includegraphics[width=1.0\linewidth]{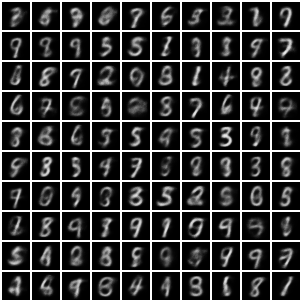}
\caption{Samples from the TVAE in Section \ref{sec:2d_TVAE}.}
\label{fig:samples}
\vspace{-15mm}
\end{wrapfigure}
In Figure \ref{fig:samples}, we provide samples from our model in the $L=0$ setting to validate that the learned latent distribution closely matches the $TPoT$ distribution described in Equation \ref{eqn:final_T}. Explicitly, the samples are generated by sampling standard normal random variables $\mathbf{Z}$ and $\mathbf{U}$, constructing $\mathbf{T}$ as in Equation \ref{eqn:final_T}, and then passing these sampled $\mathbf{T}$ through the decoder. We see that the samples resemble true MNIST digits (accounting for the limited capacity of the model), implying that the distribution after training indeed follows the desired distribution, and the model has learned to become a good generative model of the data. 
\vspace{10mm}

\subsection{Generalization to Combined Transformations at Test Time}
\label{sec:generalization_experiments}
In this section, we test the ability of the model to generate sequences composed of multiple transformations through a capsule roll, despite only being trained on individual transformations in isolation. In other words, we intend to measure the extent to which the transformations learned by a set of capsules can be combined simply by passing input sequences with corresponding combined transformations. Such generalization suggests powerful benefits to data efficiency, effectively factorizing a set of complex transformations.

Explicitly, we train the model identically to that presented in Figure \ref{fig:all_traversals}, (TVAE $L=\frac{13}{36}S$), and examine the sequences generated by a capsule roll when the partial input sequences contain combinations of transformations previously unseen during training. The results of this experiment, tested on combinations of rotation and color transforms on the MNIST test set, are presented in Figure \ref{fig:generalization} below. Although this generalization capability is not known to be guaranteed a priori, we see that the capsule traversals are frequently remarkably coherent with the input transformation, implying that the model may indeed be able to generalize to combinations of transformations. Furthermore, we observe with $L=\frac{1}{2}S$ (results not shown), this generalization capability is nearly perfect.

\begin{figure}[!h]
\centering
\begin{subfigure}{.5\linewidth}
    \centering
    \includegraphics[width=0.95\linewidth]{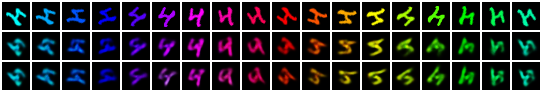}
\end{subfigure}%
\begin{subfigure}{.5\linewidth}
    \centering
    \includegraphics[width=0.95\linewidth]{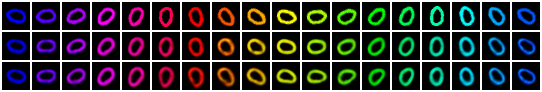}
\end{subfigure}
\vspace{2mm}

\begin{subfigure}{.5\linewidth}
    \centering
    \includegraphics[width=0.95\linewidth]{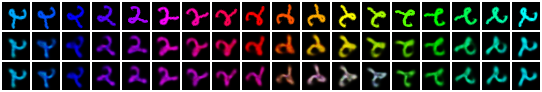}
\end{subfigure}%
\begin{subfigure}{.5\linewidth}
    \centering
    \includegraphics[width=0.95\linewidth]{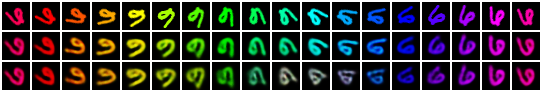}
\end{subfigure}
\vspace{2mm}

\begin{subfigure}{.5\linewidth}
    \centering
    \includegraphics[width=0.95\linewidth]{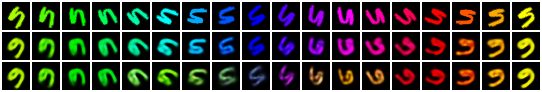}
\end{subfigure}%
\begin{subfigure}{.5\linewidth}
    \centering
    \includegraphics[width=0.95\linewidth]{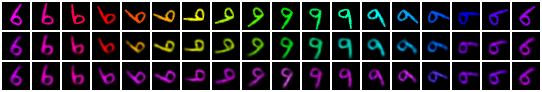}
\end{subfigure}
\vspace{2mm}

\begin{subfigure}{.5\linewidth}
    \centering
    \includegraphics[width=0.95\linewidth]{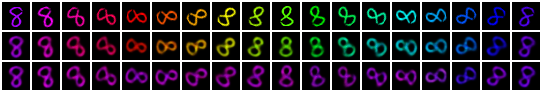}
\end{subfigure}%
\begin{subfigure}{.5\linewidth}
    \centering
    \includegraphics[width=0.95\linewidth]{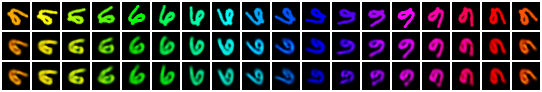}
\end{subfigure}
\vspace{2mm}

\begin{subfigure}{.5\linewidth}
    \centering
    \includegraphics[width=0.95\linewidth]{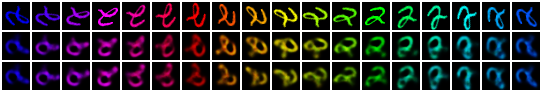}
\end{subfigure}%
\begin{subfigure}{.5\linewidth}
    \centering
    \includegraphics[width=0.95\linewidth]{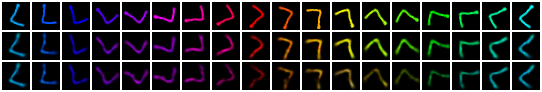}
\end{subfigure}
\vspace{2mm}

\begin{subfigure}{.5\linewidth}
    \centering
    \includegraphics[width=0.95\linewidth]{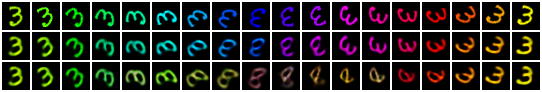}
\end{subfigure}%
\begin{subfigure}{.5\linewidth}
    \centering
    \includegraphics[width=0.95\linewidth]{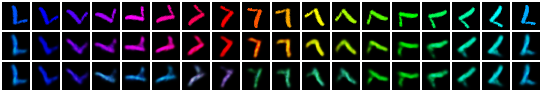}
\end{subfigure}
\caption{Capsule Traversals for MNIST TVAE $L = \frac{13}{36}S$, trained on individual transformations in isolation, and tested on combined color and rotation transformations. Top row shows the input sequence, middle row shows the direct reconstruction $\{g_{\theta}(\mathbf{t}_l)\}_l$, and bottom row shows the capsule traversal $\{g_{\theta}(\mathrm{Roll}_l[\mathbf{t}_0])\}_l$.}
\label{fig:generalization}
\end{figure}

\section{Proposed Model Extensions}
\label{sec:extensions}
\subsection{Extensions to $\mathrm{Roll}$ \& $\mathrm{CapCorr}$}
\label{sec:roll_extension}
The $\mathrm{Roll}$ operation can be seen as defining the speed at which $\mathbf{t}$ transforms corresponding to an observed transformation. For example, with $\mathrm{Roll}$ defined as in Section \ref{sec:roll_def} above, we implicitly assume that for each observed timestep, we would like the representation $\mathbf{t}$ to cyclically permute 1-unit within the capsule. For this to match the observed data, it requires the model to have an equal number of capsule dimensions and sequence elements. If we wish to reduce the size of our representation, we could instead encourage a `partial permutation' for each observed transformation. For a single capsule with $D$ elements, an example of a simple linear version of such a partial permutation (for $0 < \alpha \leq 1$) can be implemented as:
\begin{equation}
    \mathrm{Roll}_{\alpha}(\mathbf{u}) = \big[\alpha u_D + (1-\alpha)u_1, \ \ \alpha u_1 + (1-\alpha) u_2, \ \ \ldots,\ \  \alpha u_{D-1} + (1-\alpha) u_D \big]
\end{equation}
A slightly more principled partial roll for periodic signals could also be achieved by performing a phase shift of the signal in Fourier space, and performing the inverse Fourier transform to obtain the resulting rolled signal. To extend the $\mathrm{CapCorr}$ metric to similarly allow for partial $\mathrm{Rolls}$, we see that we can simply redefine the $\mathrm{ObservedRoll}$ (originally given by discrete cross-correlation) to be given by the argmax of the inner product of a sequentially partially rolled activation with the initial activation $\mathbf{t}_{\Omega}$. Formally:
\begin{equation}
\mathrm{ObservedRoll}(\mathbf{t}_{\Omega}, \mathbf{t}_{0}) = \mathrm{argmax}\left[\mathbf{t}_{\Omega} \cdot \mathrm{Roll}_0(\mathbf{t}_{0}), \ \mathbf{t}_{\Omega} \cdot \mathrm{Roll}_\alpha(\mathbf{t}_{0}),\  \ldots \ , \mathbf{t}_{\Omega} \cdot \mathrm{Roll}_{D-\alpha}(\mathbf{t}_{0})\right]
\end{equation}

\subsection{Non-Cyclic Capsules}
\label{sec:noncyclic_extension}
We can also see that there is nothing beyond convenience which inherently requires the capsules to be circular (i.e. have periodic boundary conditions). To implement linear capsules, we propose one solution is to add  $L$ additional $U_i$ variables to both the left and right boundaries of each capsule. In this way, the vector $\mathbf{U}$ is larger than the vector $\mathbf{Z}$ and can be seen as a `padded' version, where the padding is composed of independant random variables. Additionally, the transformation sequences can then be padded on both sides by replicating the first and final elements $L$ times. The construction of $\mathbf{T}$ variables is then performed identically as in Equations \ref{eqn:T_seq} and \ref{eqn:eq_roll}. The $\mathrm{Roll}$ operation can then be similarly defined as filling the boundaries with $0$ since these values will not be used as part of the computation.  

\subsection{Multi-dimensional Temporally Coherent Capsules}
In consideration of transformations which may naturally live in multiple dimensions, we wish to extend the original model to support multi-dimensional capsules. Such multi-dimensional capsules could additionally support more well-defined `disentanglement' of transformations by encouraging each transformation to be axis-aligned with one dimension of each capsule. We see that in the non-temporally coherent case ($L=0$), the model can easily be extended to capsules of multiple dimensions through multi-dimensional neighborhoods. An example of a model with 2-dimensional neighborhoods is presented in Figure 3. However, when considering shifting temporal coherence as we defined in Section 6.3, it is not clear how the shift operator or the neighborhoods should be defined for higher dimensional capsules. In this section we propose to modify the definitions of $\mathbf{T}$ in Equations \ref{eqn:T_seq} and \ref{eqn:eq_roll} with an extension resembling `group sparsity' in the denominator. 

First, we again assume that each input sequence is an observation of a single transformation at a time. Formally, the multi-dimensional capsules are then constructed by arranging $\mathbf{U}$ into a $D$ dimensional lattice. In such a model, we desire to roll and sum only along a single axis of the lattice for a given sequence. Incorporating this into the construction of $\mathbf{T}$ yields the following:
\begin{align}
\label{eqn:T_seq_md}
    \mathbf{T}_l = \frac{\mathbf{Z}_l - \mu}{\sum_{d=1}^D\sqrt{\mathbf{W}^d 
              \left[
              \mathbf{U}_{l+L}^2; 
              \cdots; \mathbf{U}_{l-L}^2
              \right]}} =
               \frac{\mathbf{Z}_l - \mu}{\sum_{d=1}^D\sqrt{\sum_{\delta=-L}^{L} \mathbf{W}^d_{\delta} \mathrm{Roll}^d_{\delta}(\mathbf{U}_{l+\delta}^2)}}
\end{align}
Where $\mathbf{W}^d_{\delta}$ refers to a matrix which sums locally along the $d^{th}$ dimension of each capsule, and not at all along the others, and similarly $\mathrm{Roll}^d_{\delta}$ rolls only along the $d^{th}$ dimension. 
In practice we observe such models can indeed disentangle up to 2 distinct transformations, but become more challenging to optimize for higher dimensions. We believe this is potentially due to the exponential growth in capsule size with increasing dimension, but leave further exploration to future work.

\subsection{Causal Temporal Coherence}
As noted in the limitations, the sequence model in this paper is not `causal', meaning that each variable $\mathbf{T}_l$ requires variables from future timesteps in the sequence ($\mathbf{U}_{l+\delta}$ for $\delta > 0$). Although for the purpose of learning equivariance in practice this may not be an issue, it may be relevant for some online learning applications. We can modify Equations \ref{eqn:T_seq} and \ref{eqn:eq_roll} by changing the matrix $\mathbf{W}$ (implemented as convolution) to a causal convolution (i.e. masking out $\mathbf{W}_{\delta}$ for $\delta>0$). Formally:
\begin{align}
\label{eqn:T_seq_causal}
    \mathbf{T}_l = \frac{\mathbf{Z}_l - \mu}{\sqrt{\mathbf{W}
              \left[
              \mathbf{U}_{l}^2; 
              \cdots; \mathbf{U}_{l-L}^2
              \right]}} =
               \frac{\mathbf{Z}_l - \mu}{\sqrt{\sum_{\delta=-L}^{0} \mathbf{W}_{\delta} \mathrm{Roll}_{\delta}(\mathbf{U}_{l+\delta}^2)}}
\end{align}
In a causal setting, it is also likely the transformations are no longer assumed to be circular. We thus refer the reader to Section \ref{sec:noncyclic_extension} above on non-circular capsules, which can be combined with Equation \ref{eqn:T_seq_causal}, to achieve such a model.

\section{Capsule Traversals}
\label{sec:capsule_traversals}
In this section we provide a set of 12 capsule traversals for each of the models presented in main text. The traversals are randomly selected such that all transformations (and dSprites shapes) are shown evenly. Unlike the main section, we additionally include a middle row which shows the direct reconstruction of the input without any rolling (i.e. $\{g_{\theta}(\mathbf{t}_l)\}_l$). We find the direct reconstructions valuable to determine if poor traversals are due to bad reconstructions (low $\log p_{\theta}(\mathbf{x}|\mathbf{t})$) or a lack of equivariance (high $\mathcal{E}_{eq}$). For example, with the baseline VAE models, we see that the reconstructions in the middle row are accurate for the full sequence, while the capsule traversals obtained by sequentially rolling the initial activation (shown in the bottom row) are nothing like the input transformation (top row). In all traversals, the left-most image corresponds to $\mathbf{t}_0$, and thus input sequences of length $2L$ cover both the left and right edges when $L>0$. 

Finally, in Figures \ref{fig:color_rot} \& \ref{fig:perspective} at the end of the section, we include capsule traversals for models trained on MNIST with more complex transformations such as combined color \& rotation, and combined color \& perspective transforms. These models were trained in an identical manner to the other MNIST models, with the same architecture, only changing the transformation sequences of the training dataset.

\begin{figure}[!h]
\centering
\begin{subfigure}{.5\linewidth}
    \centering
    \includegraphics[width=0.95\linewidth]{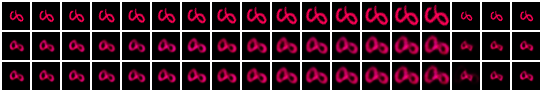}
\end{subfigure}%
\begin{subfigure}{.5\linewidth}
    \centering
    \includegraphics[width=0.95\linewidth]{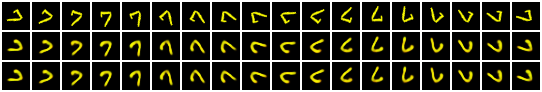}
\end{subfigure}
\vspace{2mm}

\begin{subfigure}{.5\linewidth}
    \centering
    \includegraphics[width=0.95\linewidth]{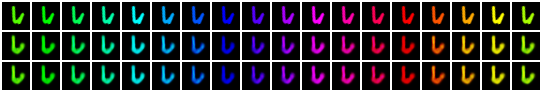}
\end{subfigure}%
\begin{subfigure}{.5\linewidth}
    \centering
    \includegraphics[width=0.95\linewidth]{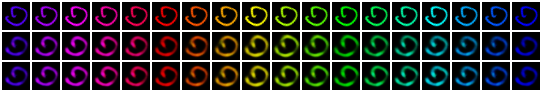}
\end{subfigure}
\vspace{2mm}

\begin{subfigure}{.5\linewidth}
    \centering
    \includegraphics[width=0.95\linewidth]{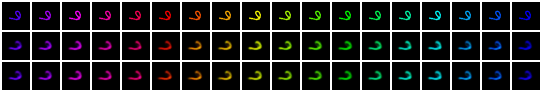}
\end{subfigure}%
\begin{subfigure}{.5\linewidth}
    \centering
    \includegraphics[width=0.95\linewidth]{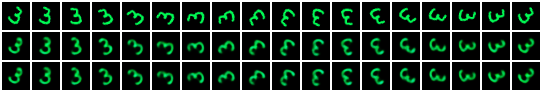}
\end{subfigure}
\vspace{2mm}

\begin{subfigure}{.5\linewidth}
    \centering
    \includegraphics[width=0.95\linewidth]{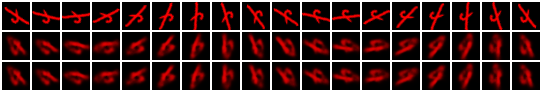}
\end{subfigure}%
\begin{subfigure}{.5\linewidth}
    \centering
    \includegraphics[width=0.95\linewidth]{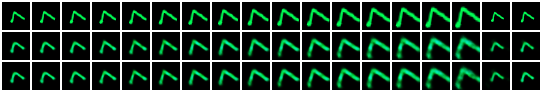}
\end{subfigure}
\vspace{2mm}

\begin{subfigure}{.5\linewidth}
    \centering
    \includegraphics[width=0.95\linewidth]{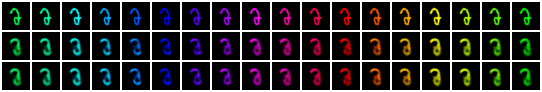}
\end{subfigure}%
\begin{subfigure}{.5\linewidth}
    \centering
    \includegraphics[width=0.95\linewidth]{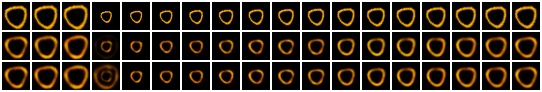}
\end{subfigure}
\vspace{2mm}

\begin{subfigure}{.5\linewidth}
    \centering
    \includegraphics[width=0.95\linewidth]{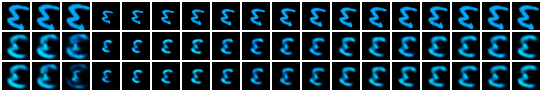}
\end{subfigure}%
\begin{subfigure}{.5\linewidth}
    \centering
    \includegraphics[width=0.95\linewidth]{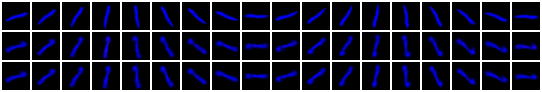}
\end{subfigure}
\caption{MNIST TVAE $L = \frac{1}{2}S$, $K=3$}
\end{figure}

\begin{figure}[!]
\centering
\begin{subfigure}{.5\linewidth}
    \centering
    \includegraphics[width=0.95\linewidth]{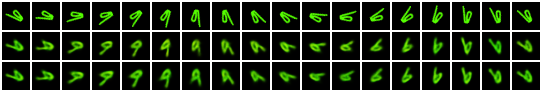}
\end{subfigure}%
\begin{subfigure}{.5\linewidth}
    \centering
    \includegraphics[width=0.95\linewidth]{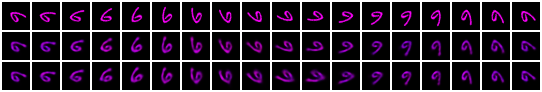}
\end{subfigure}
\vspace{2mm}

\begin{subfigure}{.5\linewidth}
    \centering
    \includegraphics[width=0.95\linewidth]{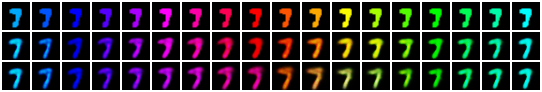}
\end{subfigure}%
\begin{subfigure}{.5\linewidth}
    \centering
    \includegraphics[width=0.95\linewidth]{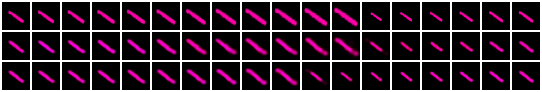}
\end{subfigure}
\vspace{2mm}

\begin{subfigure}{.5\linewidth}
    \centering
    \includegraphics[width=0.95\linewidth]{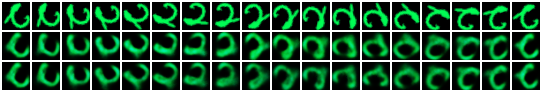}
\end{subfigure}%
\begin{subfigure}{.5\linewidth}
    \centering
    \includegraphics[width=0.95\linewidth]{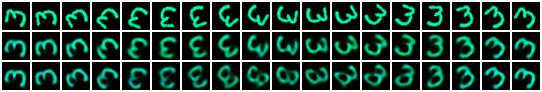}
\end{subfigure}
\vspace{2mm}

\begin{subfigure}{.5\linewidth}
    \centering
    \includegraphics[width=0.95\linewidth]{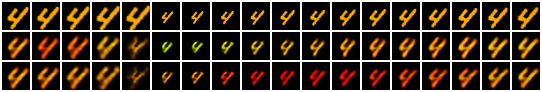}
\end{subfigure}%
\begin{subfigure}{.5\linewidth}
    \centering
    \includegraphics[width=0.95\linewidth]{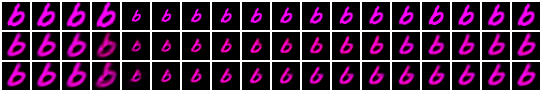}
\end{subfigure}
\vspace{2mm}

\begin{subfigure}{.5\linewidth}
    \centering
    \includegraphics[width=0.95\linewidth]{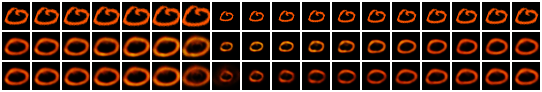}
\end{subfigure}%
\begin{subfigure}{.5\linewidth}
    \centering
    \includegraphics[width=0.95\linewidth]{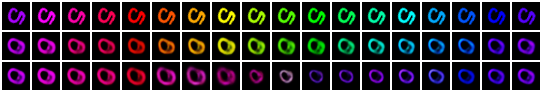}
\end{subfigure}
\vspace{2mm}

\begin{subfigure}{.5\linewidth}
    \centering
    \includegraphics[width=0.95\linewidth]{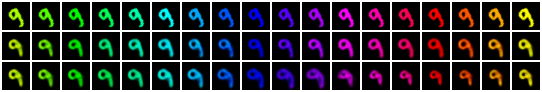}
\end{subfigure}%
\begin{subfigure}{.5\linewidth}
    \centering
    \includegraphics[width=0.95\linewidth]{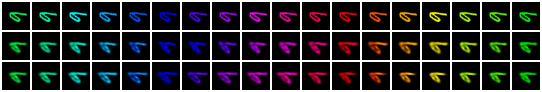}
\end{subfigure}
\caption{MNIST TVAE $L = \frac{13}{36}S$, $K=3$}
\end{figure}

\begin{figure}[!]
\centering
\begin{subfigure}{.5\linewidth}
    \centering
    \includegraphics[width=0.95\linewidth]{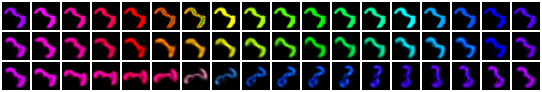}
\end{subfigure}%
\begin{subfigure}{.5\linewidth}
    \centering
    \includegraphics[width=0.95\linewidth]{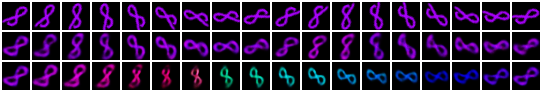}
\end{subfigure}
\vspace{2mm}

\begin{subfigure}{.5\linewidth}
    \centering
    \includegraphics[width=0.95\linewidth]{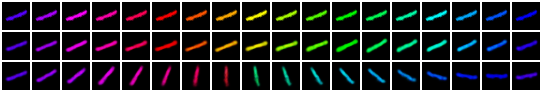}
\end{subfigure}%
\begin{subfigure}{.5\linewidth}
    \centering
    \includegraphics[width=0.95\linewidth]{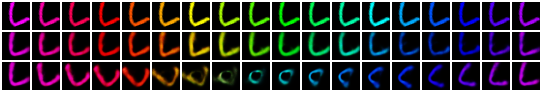}
\end{subfigure}
\vspace{2mm}

\begin{subfigure}{.5\linewidth}
    \centering
    \includegraphics[width=0.95\linewidth]{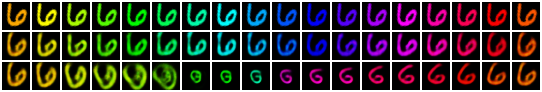}
\end{subfigure}%
\begin{subfigure}{.5\linewidth}
    \centering
    \includegraphics[width=0.95\linewidth]{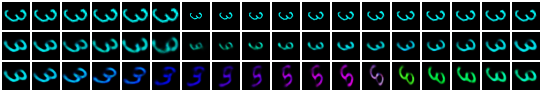}
\end{subfigure}
\vspace{2mm}

\begin{subfigure}{.5\linewidth}
    \centering
    \includegraphics[width=0.95\linewidth]{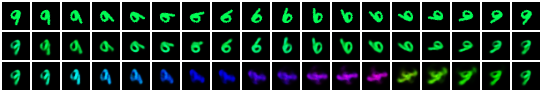}
\end{subfigure}%
\begin{subfigure}{.5\linewidth}
    \centering
    \includegraphics[width=0.95\linewidth]{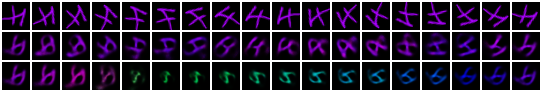}
\end{subfigure}
\vspace{2mm}

\begin{subfigure}{.5\linewidth}
    \centering
    \includegraphics[width=0.95\linewidth]{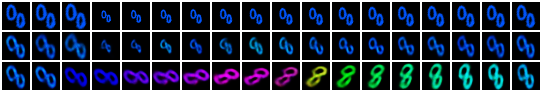}
\end{subfigure}%
\begin{subfigure}{.5\linewidth}
    \centering
    \includegraphics[width=0.95\linewidth]{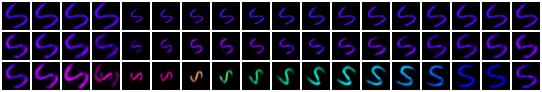}
\end{subfigure}
\vspace{2mm}

\begin{subfigure}{.5\linewidth}
    \centering
    \includegraphics[width=0.95\linewidth]{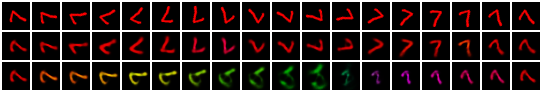}
\end{subfigure}%
\begin{subfigure}{.5\linewidth}
    \centering
    \includegraphics[width=0.95\linewidth]{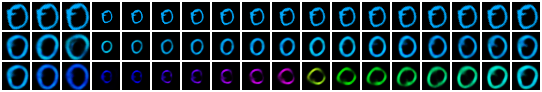}
\end{subfigure}
\caption{MNIST TVAE $L = \frac{5}{36}S$, $K=3$. We see with values of $L<\frac{1}{3}S$ the transformations decoded through the capsule roll are only partially coherent with the input sequence.}
\label{fig:tvae_L5d36_k3}
\end{figure}

\begin{figure}[!]
\centering
\begin{subfigure}{.5\linewidth}
    \centering
    \includegraphics[width=0.95\linewidth]{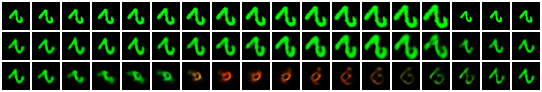}
\end{subfigure}%
\begin{subfigure}{.5\linewidth}
    \centering
    \includegraphics[width=0.95\linewidth]{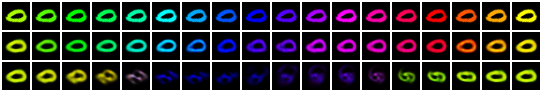}
\end{subfigure}
\vspace{2mm}

\begin{subfigure}{.5\linewidth}
    \centering
    \includegraphics[width=0.95\linewidth]{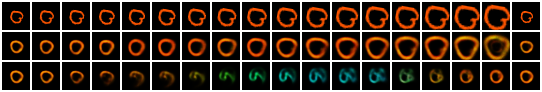}
\end{subfigure}%
\begin{subfigure}{.5\linewidth}
    \centering
    \includegraphics[width=0.95\linewidth]{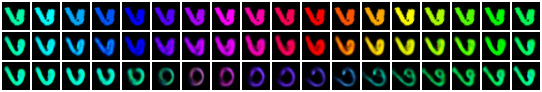}
\end{subfigure}
\vspace{2mm}

\begin{subfigure}{.5\linewidth}
    \centering
    \includegraphics[width=0.95\linewidth]{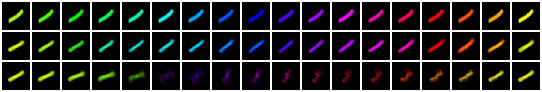}
\end{subfigure}%
\begin{subfigure}{.5\linewidth}
    \centering
    \includegraphics[width=0.95\linewidth]{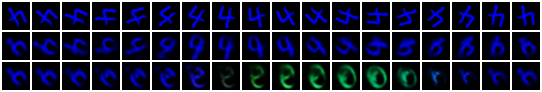}
\end{subfigure}
\vspace{2mm}

\begin{subfigure}{.5\linewidth}
    \centering
    \includegraphics[width=0.95\linewidth]{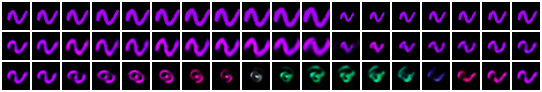}
\end{subfigure}%
\begin{subfigure}{.5\linewidth}
    \centering
    \includegraphics[width=0.95\linewidth]{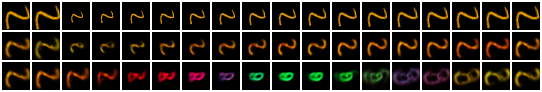}
\end{subfigure}
\vspace{2mm}

\begin{subfigure}{.5\linewidth}
    \centering
    \includegraphics[width=0.95\linewidth]{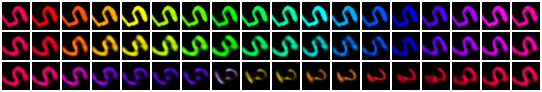}
\end{subfigure}%
\begin{subfigure}{.5\linewidth}
    \centering
    \includegraphics[width=0.95\linewidth]{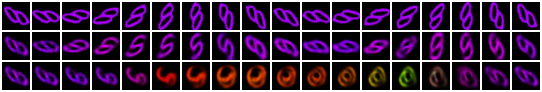}
\end{subfigure}
\vspace{2mm}

\begin{subfigure}{.5\linewidth}
    \centering
    \includegraphics[width=0.95\linewidth]{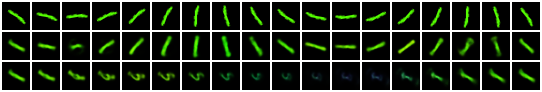}
\end{subfigure}%
\begin{subfigure}{.5\linewidth}
    \centering
    \includegraphics[width=0.95\linewidth]{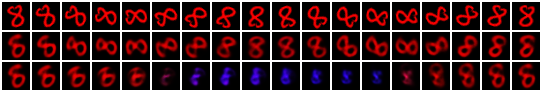}
\end{subfigure}
\caption{MNIST TVAE $L = \frac{5}{36}S$, $K=9$}
\label{fig:tvae_L5d36_k9}
\end{figure}

\begin{figure}[!]
\centering
\begin{subfigure}{.5\linewidth}
    \centering
    \includegraphics[width=0.95\linewidth]{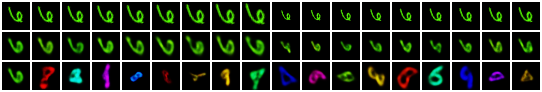}
\end{subfigure}%
\begin{subfigure}{.5\linewidth}
    \centering
    \includegraphics[width=0.95\linewidth]{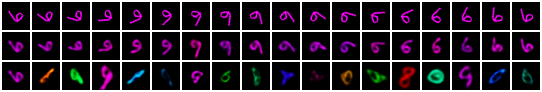}
\end{subfigure}
\vspace{2mm}

\begin{subfigure}{.5\linewidth}
    \centering
    \includegraphics[width=0.95\linewidth]{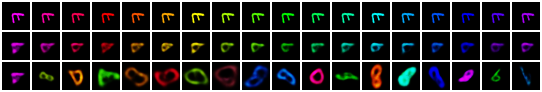}
\end{subfigure}%
\begin{subfigure}{.5\linewidth}
    \centering
    \includegraphics[width=0.95\linewidth]{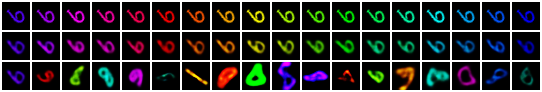}
\end{subfigure}
\vspace{2mm}

\begin{subfigure}{.5\linewidth}
    \centering
    \includegraphics[width=0.95\linewidth]{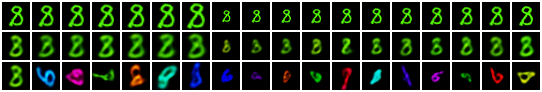}
\end{subfigure}%
\begin{subfigure}{.5\linewidth}
    \centering
    \includegraphics[width=0.95\linewidth]{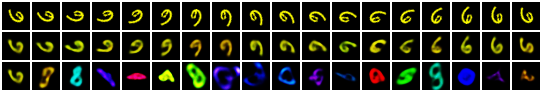}
\end{subfigure}
\vspace{2mm}

\begin{subfigure}{.5\linewidth}
    \centering
    \includegraphics[width=0.95\linewidth]{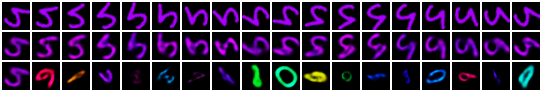}
\end{subfigure}%
\begin{subfigure}{.5\linewidth}
    \centering
    \includegraphics[width=0.95\linewidth]{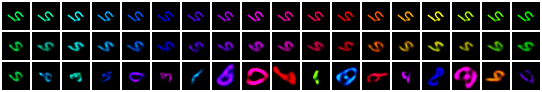}
\end{subfigure}
\vspace{2mm}

\begin{subfigure}{.5\linewidth}
    \centering
    \includegraphics[width=0.95\linewidth]{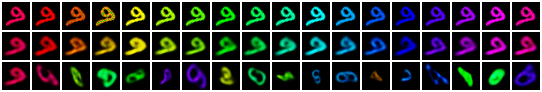}
\end{subfigure}%
\begin{subfigure}{.5\linewidth}
    \centering
    \includegraphics[width=0.95\linewidth]{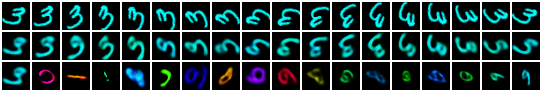}
\end{subfigure}
\vspace{2mm}

\begin{subfigure}{.5\linewidth}
    \centering
    \includegraphics[width=0.95\linewidth]{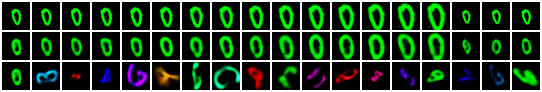}
\end{subfigure}%
\begin{subfigure}{.5\linewidth}
    \centering
    \includegraphics[width=0.95\linewidth]{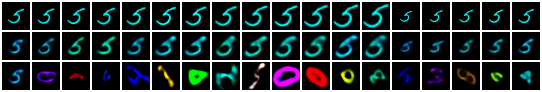}
\end{subfigure}
\caption{MNIST TVAE $L = 0$, $K=3$. We see for sufficiently small values of $K$, the TVAE can reach a degenerate solution where topographic organization is almost entirely lost.}
\label{fig:tvae_L0_k3}
\end{figure}

\begin{figure}[!]
\centering
\begin{subfigure}{.5\linewidth}
    \centering
    \includegraphics[width=0.95\linewidth]{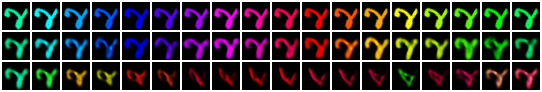}
\end{subfigure}%
\begin{subfigure}{.5\linewidth}
    \centering
    \includegraphics[width=0.95\linewidth]{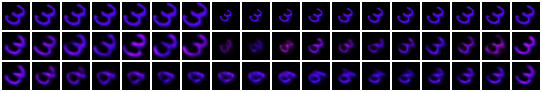}
\end{subfigure}
\vspace{2mm}

\begin{subfigure}{.5\linewidth}
    \centering
    \includegraphics[width=0.95\linewidth]{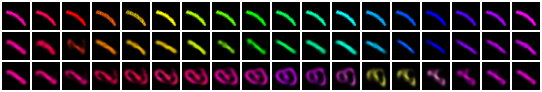}
\end{subfigure}%
\begin{subfigure}{.5\linewidth}
    \centering
    \includegraphics[width=0.95\linewidth]{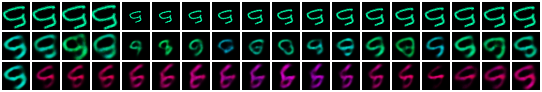}
\end{subfigure}
\vspace{2mm}

\begin{subfigure}{.5\linewidth}
    \centering
    \includegraphics[width=0.95\linewidth]{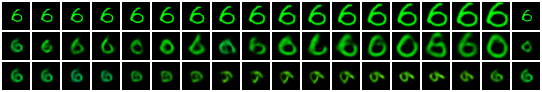}
\end{subfigure}%
\begin{subfigure}{.5\linewidth}
    \centering
    \includegraphics[width=0.95\linewidth]{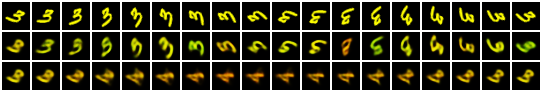}
\end{subfigure}
\vspace{2mm}

\begin{subfigure}{.5\linewidth}
    \centering
    \includegraphics[width=0.95\linewidth]{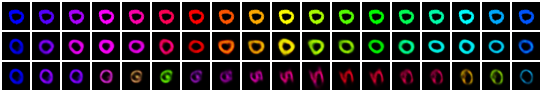}
\end{subfigure}%
\begin{subfigure}{.5\linewidth}
    \centering
    \includegraphics[width=0.95\linewidth]{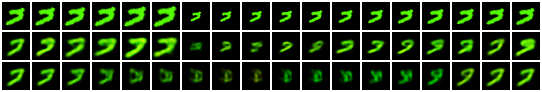}
\end{subfigure}
\vspace{2mm}

\begin{subfigure}{.5\linewidth}
    \centering
    \includegraphics[width=0.95\linewidth]{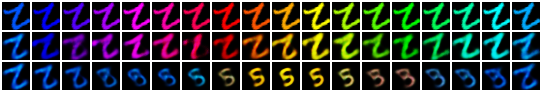}
\end{subfigure}%
\begin{subfigure}{.5\linewidth}
    \centering
    \includegraphics[width=0.95\linewidth]{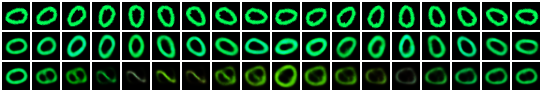}
\end{subfigure}
\vspace{2mm}

\begin{subfigure}{.5\linewidth}
    \centering
    \includegraphics[width=0.95\linewidth]{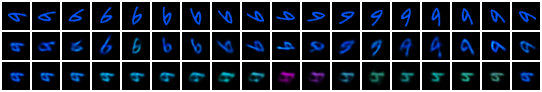}
\end{subfigure}%
\begin{subfigure}{.5\linewidth}
    \centering
    \includegraphics[width=0.95\linewidth]{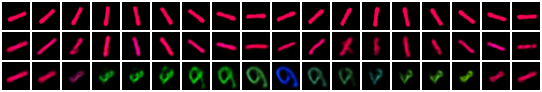}
\end{subfigure}
\caption{MNIST TVAE $L = 0$, $K=9$}
\label{fig:tvae_L0_k9}
\end{figure}

\begin{figure}[!]
\centering
\begin{subfigure}{.5\linewidth}
    \centering
    \includegraphics[width=0.95\linewidth]{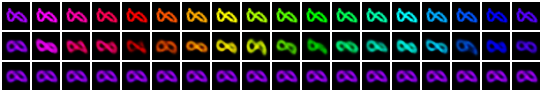}
\end{subfigure}%
\begin{subfigure}{.5\linewidth}
    \centering
    \includegraphics[width=0.95\linewidth]{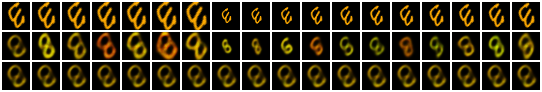}
\end{subfigure}
\vspace{2mm}

\begin{subfigure}{.5\linewidth}
    \centering
    \includegraphics[width=0.95\linewidth]{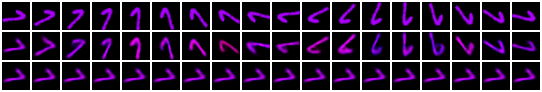}
\end{subfigure}%
\begin{subfigure}{.5\linewidth}
    \centering
    \includegraphics[width=0.95\linewidth]{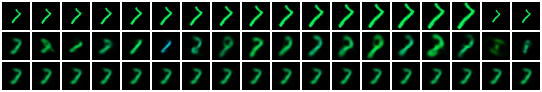}
\end{subfigure}
\vspace{2mm}

\begin{subfigure}{.5\linewidth}
    \centering
    \includegraphics[width=0.95\linewidth]{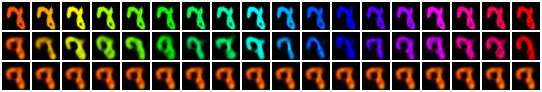}
\end{subfigure}%
\begin{subfigure}{.5\linewidth}
    \centering
    \includegraphics[width=0.95\linewidth]{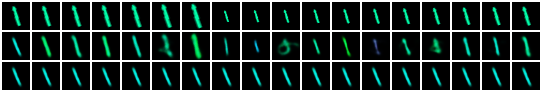}
\end{subfigure}
\vspace{2mm}

\begin{subfigure}{.5\linewidth}
    \centering
    \includegraphics[width=0.95\linewidth]{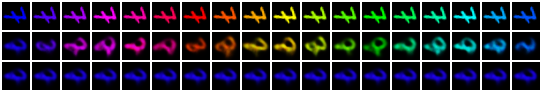}
\end{subfigure}%
\begin{subfigure}{.5\linewidth}
    \centering
    \includegraphics[width=0.95\linewidth]{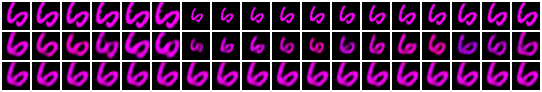}
\end{subfigure}
\vspace{2mm}

\begin{subfigure}{.5\linewidth}
    \centering
    \includegraphics[width=0.95\linewidth]{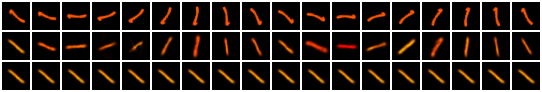}
\end{subfigure}%
\begin{subfigure}{.5\linewidth}
    \centering
    \includegraphics[width=0.95\linewidth]{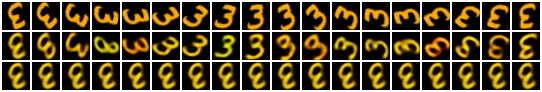}
\end{subfigure}
\vspace{2mm}

\begin{subfigure}{.5\linewidth}
    \centering
    \includegraphics[width=0.95\linewidth]{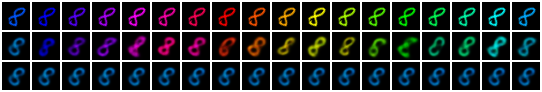}
\end{subfigure}%
\begin{subfigure}{.5\linewidth}
    \centering
    \includegraphics[width=0.95\linewidth]{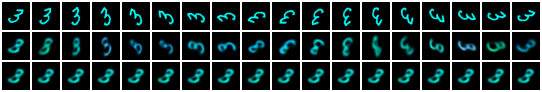}
\end{subfigure}
\caption{MNIST TVAE $L = 0$, $K=18$. We see when $K$ is equal to the capsule size (making the model analogous to ISA), the model learns an invariant capsule representation -- meaning $\mathrm{Roll}$ing a capsule activation produces no significant transformation in the observation space.}
\label{fig:tvae_L0_k18}
\end{figure}

\begin{figure}[!]
\centering
\begin{subfigure}{.5\linewidth}
    \centering
    \includegraphics[width=0.95\linewidth]{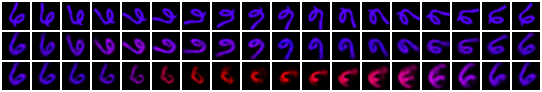}
\end{subfigure}%
\begin{subfigure}{.5\linewidth}
    \centering
    \includegraphics[width=0.95\linewidth]{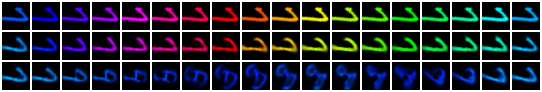}
\end{subfigure}
\vspace{2mm}

\begin{subfigure}{.5\linewidth}
    \centering
    \includegraphics[width=0.95\linewidth]{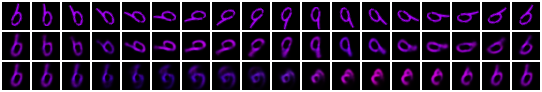}
\end{subfigure}%
\begin{subfigure}{.5\linewidth}
    \centering
    \includegraphics[width=0.95\linewidth]{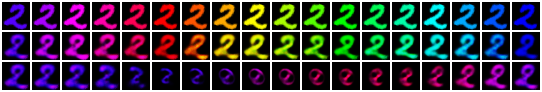}
\end{subfigure}
\vspace{2mm}

\begin{subfigure}{.5\linewidth}
    \centering
    \includegraphics[width=0.95\linewidth]{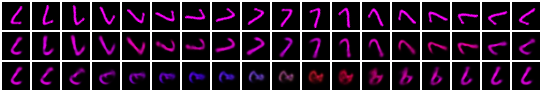}
\end{subfigure}%
\begin{subfigure}{.5\linewidth}
    \centering
    \includegraphics[width=0.95\linewidth]{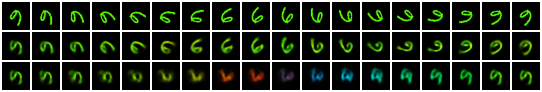}
\end{subfigure}
\vspace{2mm}

\begin{subfigure}{.5\linewidth}
    \centering
    \includegraphics[width=0.95\linewidth]{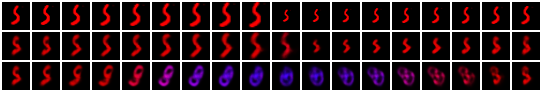}
\end{subfigure}%
\begin{subfigure}{.5\linewidth}
    \centering
    \includegraphics[width=0.95\linewidth]{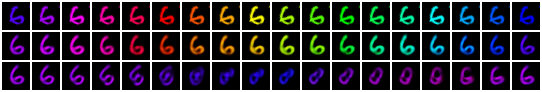}
\end{subfigure}
\vspace{2mm}

\begin{subfigure}{.5\linewidth}
    \centering
    \includegraphics[width=0.95\linewidth]{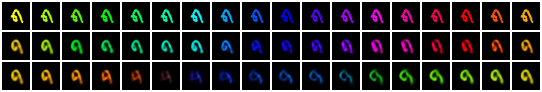}
\end{subfigure}%
\begin{subfigure}{.5\linewidth}
    \centering
    \includegraphics[width=0.95\linewidth]{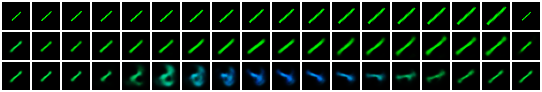}
\end{subfigure}
\vspace{2mm}

\begin{subfigure}{.5\linewidth}
    \centering
    \includegraphics[width=0.95\linewidth]{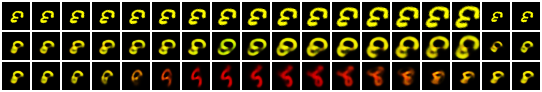}
\end{subfigure}%
\begin{subfigure}{.5\linewidth}
    \centering
    \includegraphics[width=0.95\linewidth]{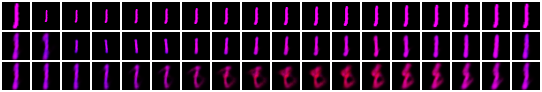}
\end{subfigure}
\caption{MNIST BubbleVAE $L = \frac{5}{36}S$, $K=2L$}
\end{figure}

\begin{figure}[!]
\centering
\begin{subfigure}{.5\linewidth}
    \centering
    \includegraphics[width=0.95\linewidth]{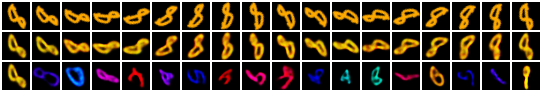}
\end{subfigure}%
\begin{subfigure}{.5\linewidth}
    \centering
    \includegraphics[width=0.95\linewidth]{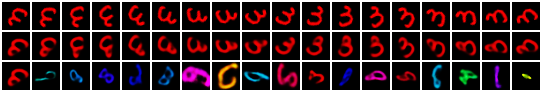}
\end{subfigure}
\vspace{2mm}

\begin{subfigure}{.5\linewidth}
    \centering
    \includegraphics[width=0.95\linewidth]{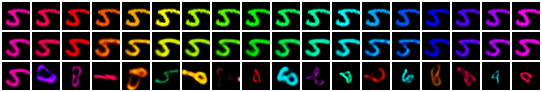}
\end{subfigure}%
\begin{subfigure}{.5\linewidth}
    \centering
    \includegraphics[width=0.95\linewidth]{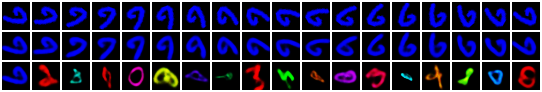}
\end{subfigure}
\vspace{2mm}

\begin{subfigure}{.5\linewidth}
    \centering
    \includegraphics[width=0.95\linewidth]{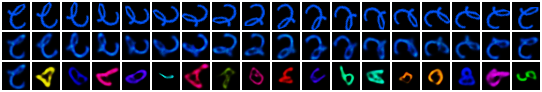}
\end{subfigure}%
\begin{subfigure}{.5\linewidth}
    \centering
    \includegraphics[width=0.95\linewidth]{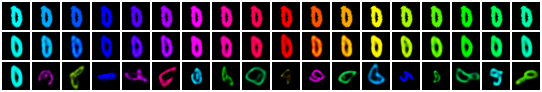}
\end{subfigure}
\vspace{2mm}

\begin{subfigure}{.5\linewidth}
    \centering
    \includegraphics[width=0.95\linewidth]{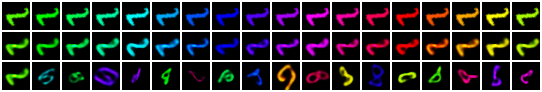}
\end{subfigure}%
\begin{subfigure}{.5\linewidth}
    \centering
    \includegraphics[width=0.95\linewidth]{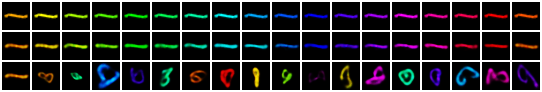}
\end{subfigure}
\vspace{2mm}

\begin{subfigure}{.5\linewidth}
    \centering
    \includegraphics[width=0.95\linewidth]{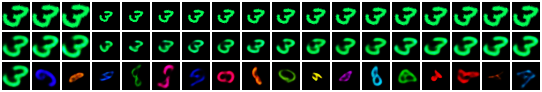}
\end{subfigure}%
\begin{subfigure}{.5\linewidth}
    \centering
    \includegraphics[width=0.95\linewidth]{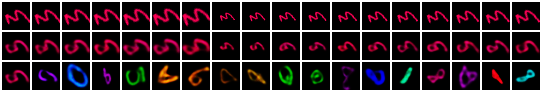}
\end{subfigure}
\vspace{2mm}

\begin{subfigure}{.5\linewidth}
    \centering
    \includegraphics[width=0.95\linewidth]{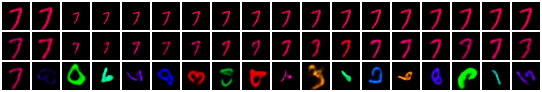}
\end{subfigure}%
\begin{subfigure}{.5\linewidth}
    \centering
    \includegraphics[width=0.95\linewidth]{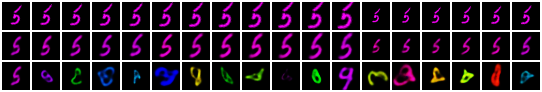}
\end{subfigure}
\caption{MNIST VAE $L = 0$, $K=1$. We see images generated through capsule traversal with the baseline VAE appear entirely random, as expected due to the non-topographic nature of the VAE's latent space.}
\end{figure}

\begin{figure}[!]
\centering
\begin{subfigure}{.5\linewidth}
    \centering
    \includegraphics[width=0.95\linewidth]{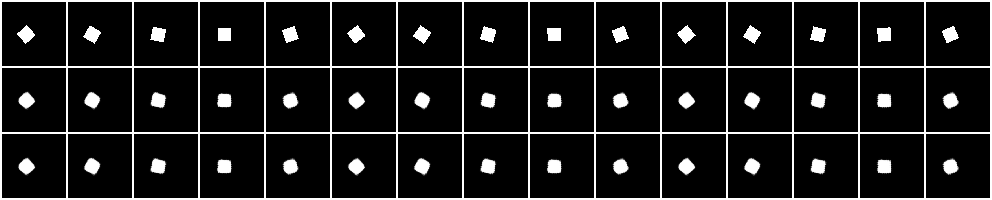}
\end{subfigure}%
\begin{subfigure}{.5\linewidth}
    \centering
    \includegraphics[width=0.95\linewidth]{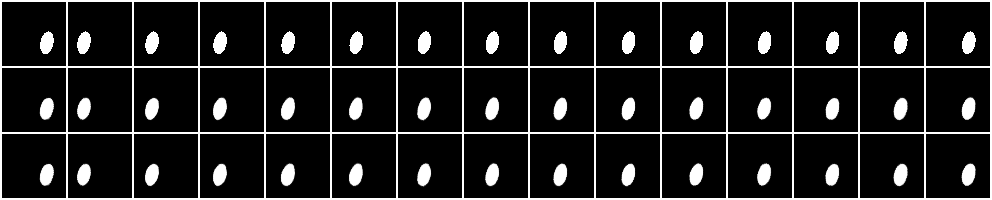}
\end{subfigure}
\vspace{2mm}

\begin{subfigure}{.5\linewidth}
    \centering
    \includegraphics[width=0.95\linewidth]{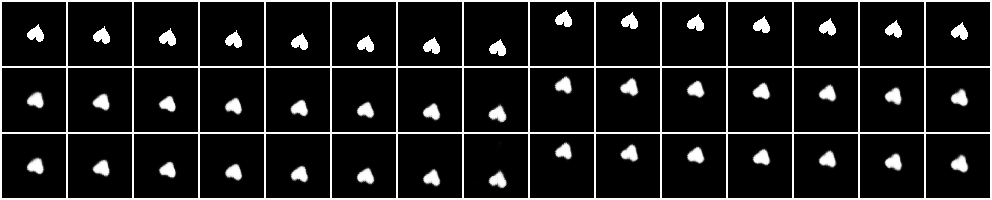}
\end{subfigure}%
\begin{subfigure}{.5\linewidth}
    \centering
    \includegraphics[width=0.95\linewidth]{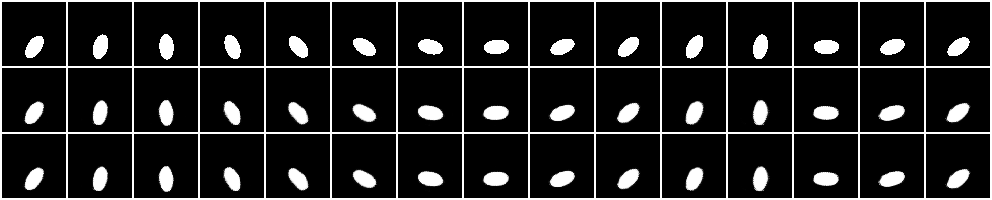}
\end{subfigure}
\vspace{2mm}

\begin{subfigure}{.5\linewidth}
    \centering
    \includegraphics[width=0.95\linewidth]{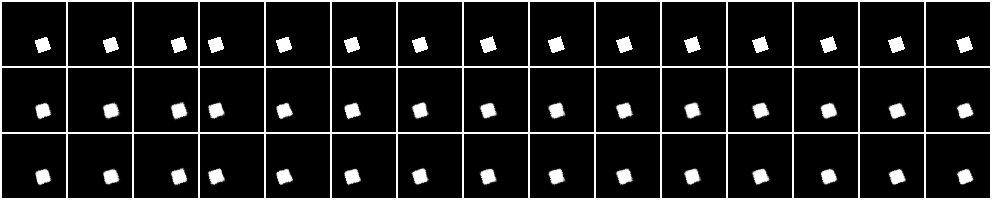}
\end{subfigure}%
\begin{subfigure}{.5\linewidth}
    \centering
    \includegraphics[width=0.95\linewidth]{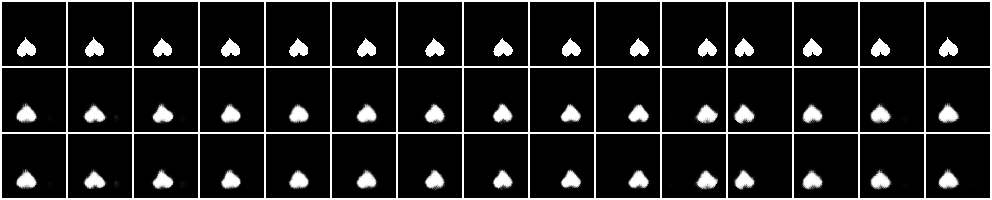}
\end{subfigure}
\vspace{2mm}

\begin{subfigure}{.5\linewidth}
    \centering
    \includegraphics[width=0.95\linewidth]{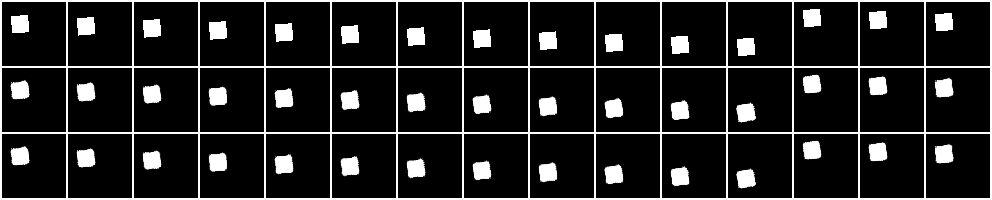}
\end{subfigure}%
\begin{subfigure}{.5\linewidth}
    \centering
    \includegraphics[width=0.95\linewidth]{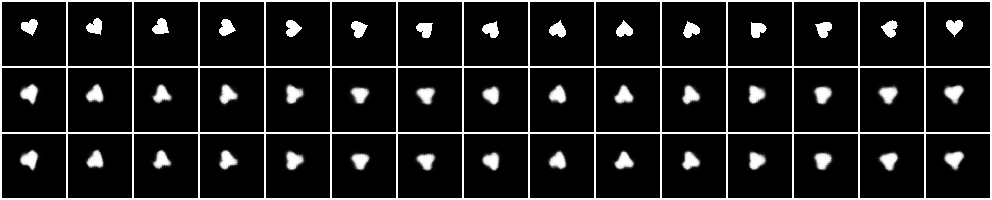}
\end{subfigure}
\vspace{2mm}

\begin{subfigure}{.5\linewidth}
    \centering
    \includegraphics[width=0.95\linewidth]{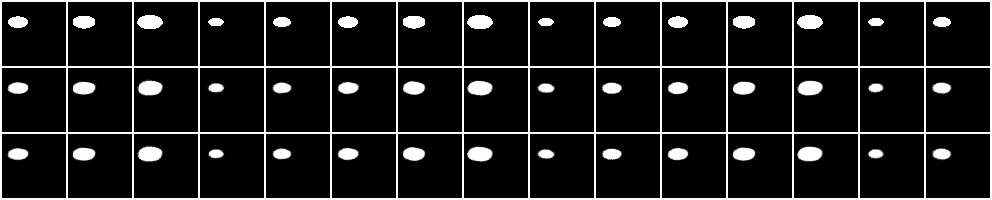}
\end{subfigure}%
\begin{subfigure}{.5\linewidth}
    \centering
    \includegraphics[width=0.95\linewidth]{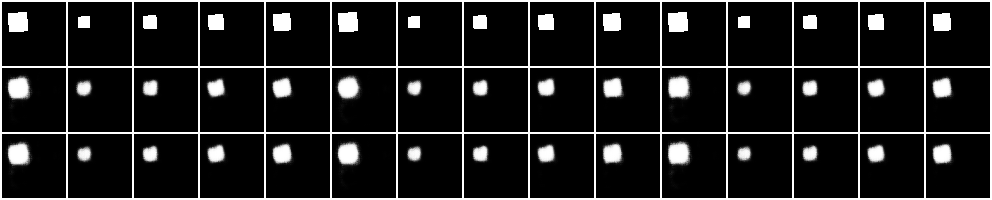}
\end{subfigure}
\vspace{2mm}

\begin{subfigure}{.5\linewidth}
    \centering
    \includegraphics[width=0.95\linewidth]{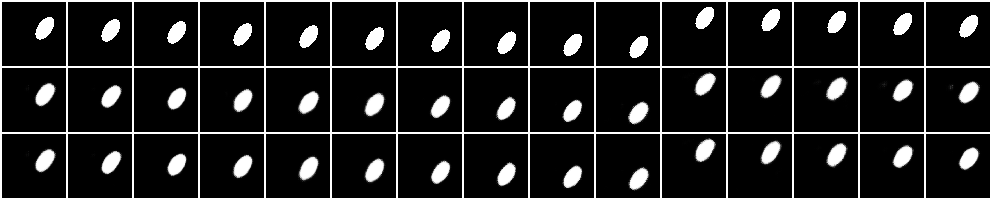}
\end{subfigure}%
\begin{subfigure}{.5\linewidth}
    \centering
    \includegraphics[width=0.95\linewidth]{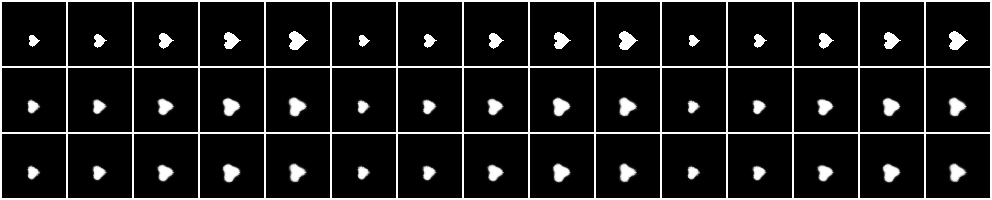}
\end{subfigure}
\caption{dSprites TVAE $L = \frac{1}{2}S$, $K=1$}
\end{figure}

\begin{figure}[!]
\centering
\begin{subfigure}{.5\linewidth}
    \centering
    \includegraphics[width=0.95\linewidth]{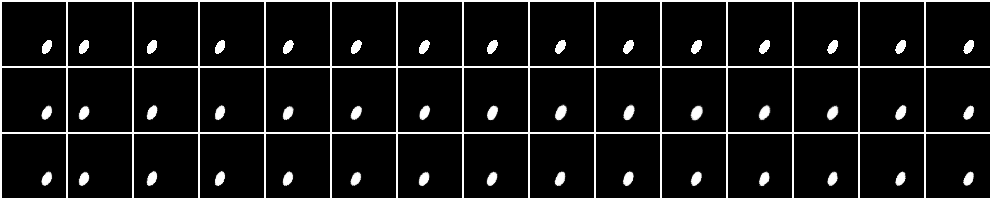}
\end{subfigure}%
\begin{subfigure}{.5\linewidth}
    \centering
    \includegraphics[width=0.95\linewidth]{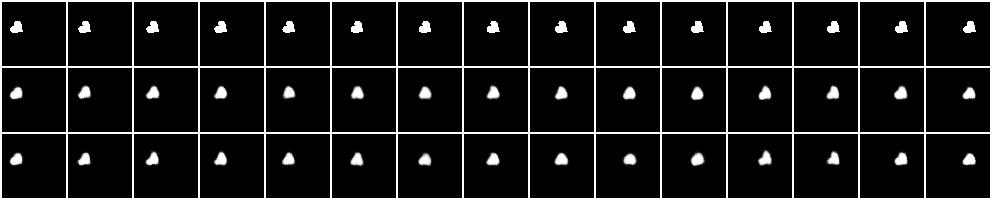}
\end{subfigure}
\vspace{2mm}

\begin{subfigure}{.5\linewidth}
    \centering
    \includegraphics[width=0.95\linewidth]{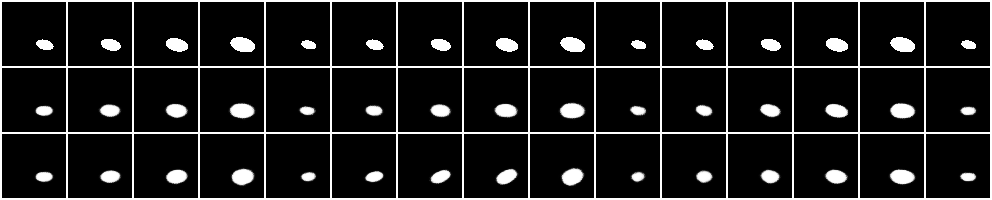}
\end{subfigure}%
\begin{subfigure}{.5\linewidth}
    \centering
    \includegraphics[width=0.95\linewidth]{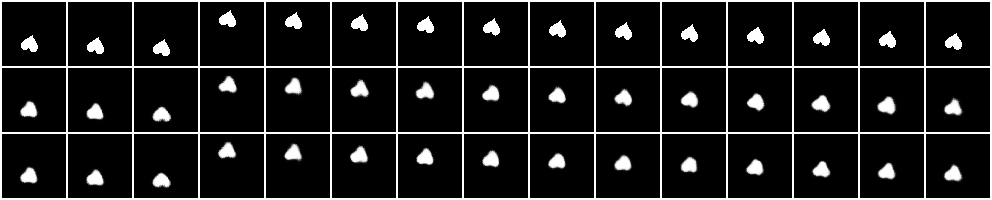}
\end{subfigure}
\vspace{2mm}

\begin{subfigure}{.5\linewidth}
    \centering
    \includegraphics[width=0.95\linewidth]{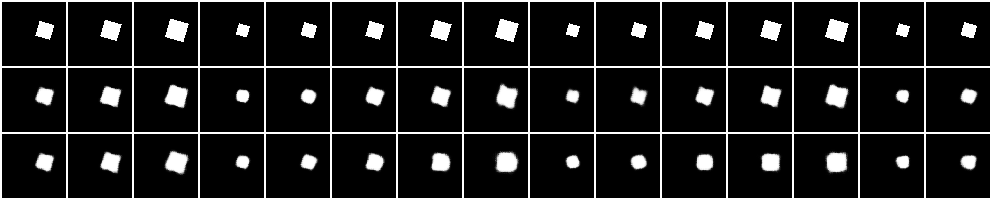}
\end{subfigure}%
\begin{subfigure}{.5\linewidth}
    \centering
    \includegraphics[width=0.95\linewidth]{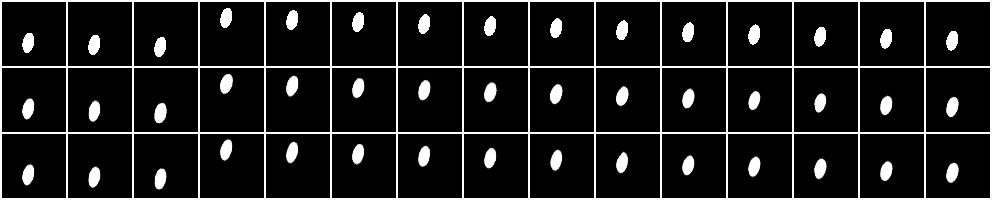}
\end{subfigure}
\vspace{2mm}

\begin{subfigure}{.5\linewidth}
    \centering
    \includegraphics[width=0.95\linewidth]{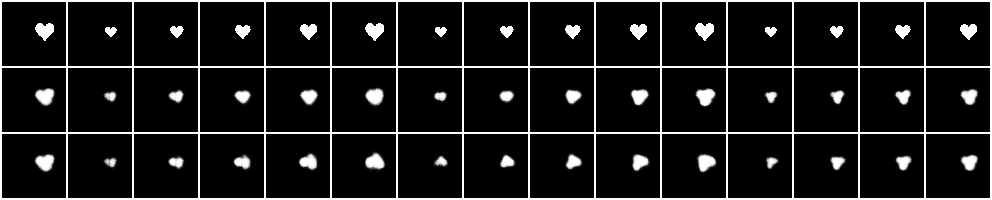}
\end{subfigure}%
\begin{subfigure}{.5\linewidth}
    \centering
    \includegraphics[width=0.95\linewidth]{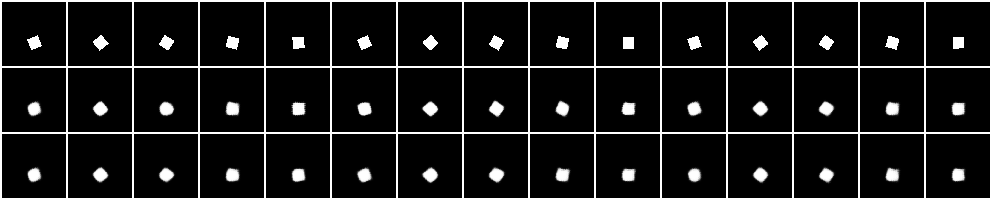}
\end{subfigure}
\vspace{2mm}

\begin{subfigure}{.5\linewidth}
    \centering
    \includegraphics[width=0.95\linewidth]{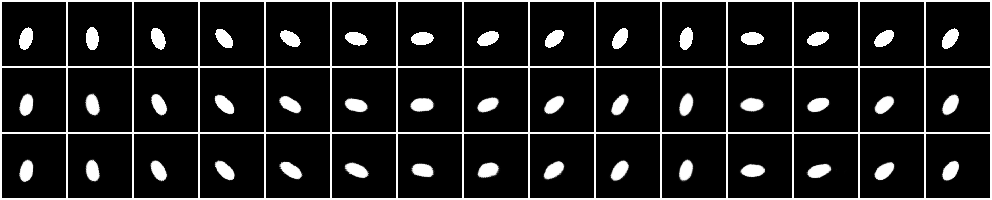}
\end{subfigure}%
\begin{subfigure}{.5\linewidth}
    \centering
    \includegraphics[width=0.95\linewidth]{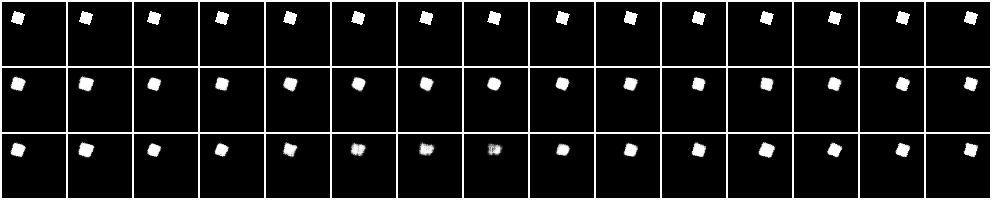}
\end{subfigure}
\vspace{2mm}

\begin{subfigure}{.5\linewidth}
    \centering
    \includegraphics[width=0.95\linewidth]{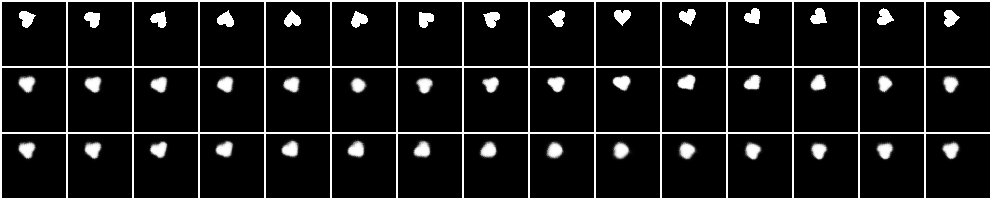}
\end{subfigure}%
\begin{subfigure}{.5\linewidth}
    \centering
    \includegraphics[width=0.95\linewidth]{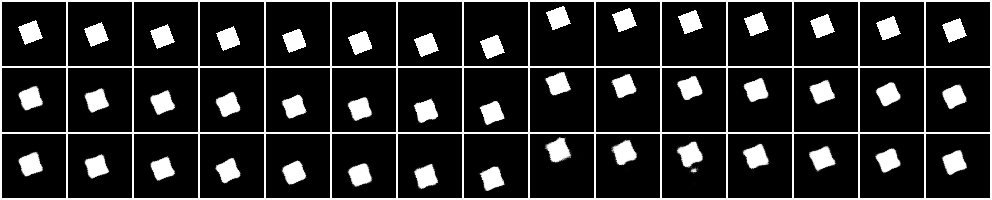}
\end{subfigure}
\caption{dSprites TVAE $L = \frac{1}{3}S$, $K=1$}
\end{figure}

\begin{figure}[!]
\centering
\begin{subfigure}{.5\linewidth}
    \centering
    \includegraphics[width=0.95\linewidth]{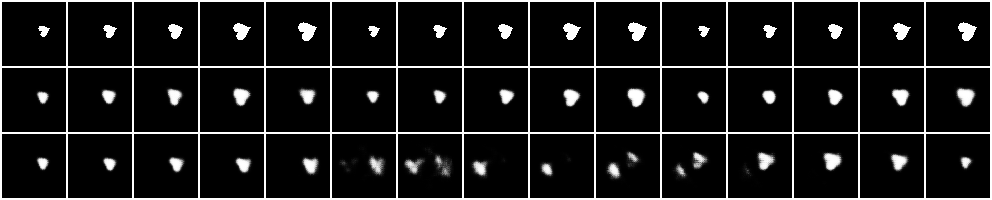}
\end{subfigure}%
\begin{subfigure}{.5\linewidth}
    \centering
    \includegraphics[width=0.95\linewidth]{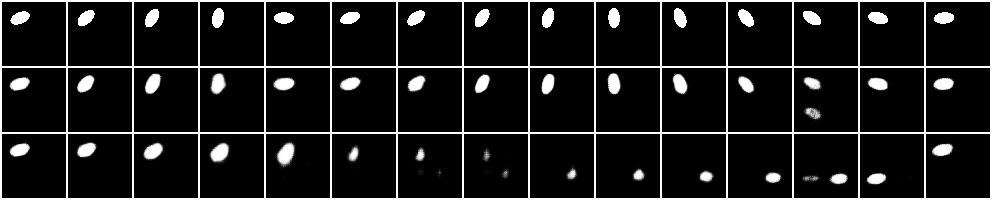}
\end{subfigure}
\vspace{2mm}

\begin{subfigure}{.5\linewidth}
    \centering
    \includegraphics[width=0.95\linewidth]{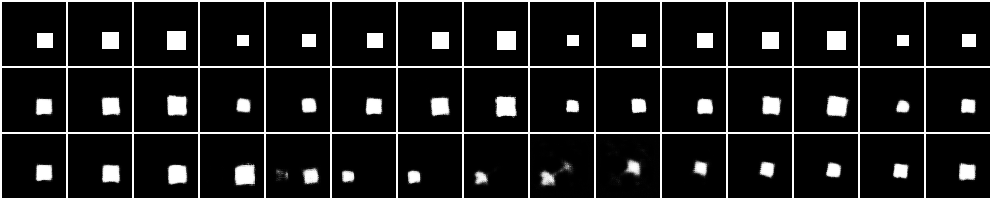}
\end{subfigure}%
\begin{subfigure}{.5\linewidth}
    \centering
    \includegraphics[width=0.95\linewidth]{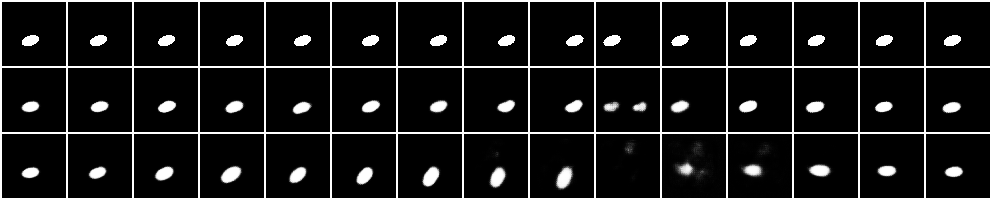}
\end{subfigure}
\vspace{2mm}

\begin{subfigure}{.5\linewidth}
    \centering
    \includegraphics[width=0.95\linewidth]{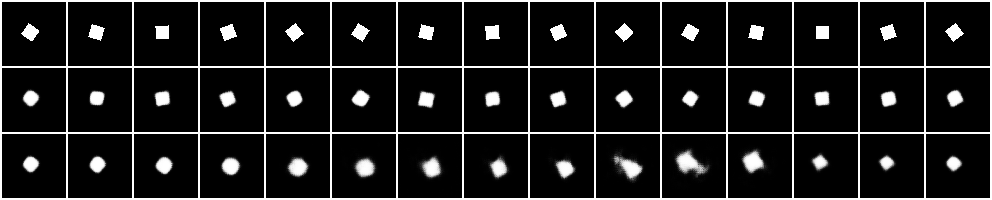}
\end{subfigure}%
\begin{subfigure}{.5\linewidth}
    \centering
    \includegraphics[width=0.95\linewidth]{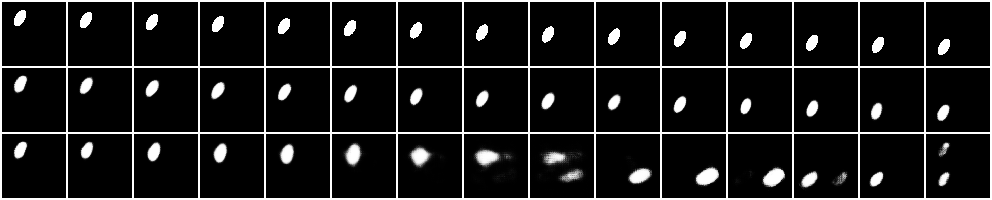}
\end{subfigure}
\vspace{2mm}

\begin{subfigure}{.5\linewidth}
    \centering
    \includegraphics[width=0.95\linewidth]{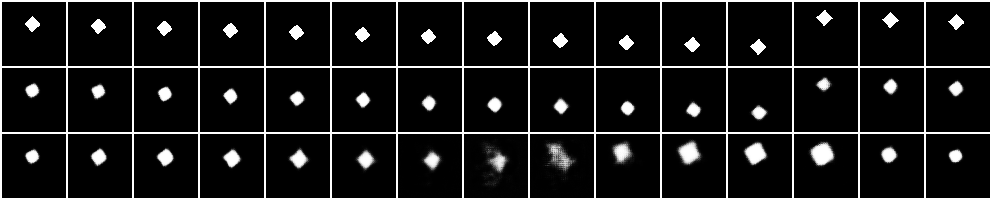}
\end{subfigure}%
\begin{subfigure}{.5\linewidth}
    \centering
    \includegraphics[width=0.95\linewidth]{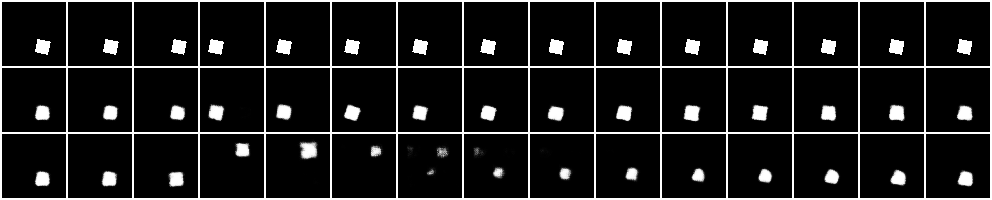}
\end{subfigure}
\vspace{2mm}

\begin{subfigure}{.5\linewidth}
    \centering
    \includegraphics[width=0.95\linewidth]{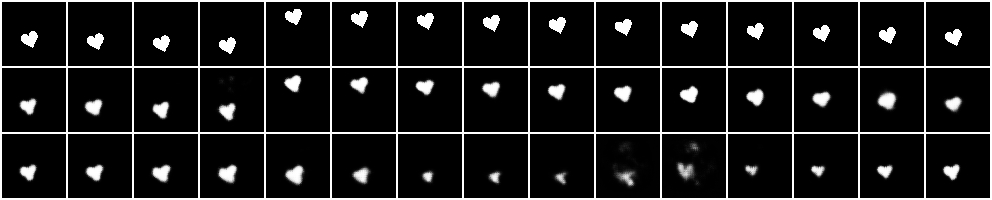}
\end{subfigure}%
\begin{subfigure}{.5\linewidth}
    \centering
    \includegraphics[width=0.95\linewidth]{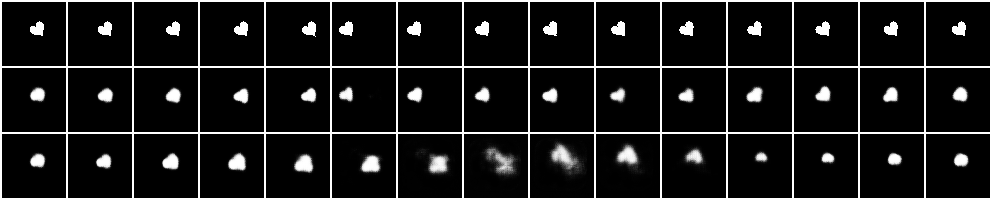}
\end{subfigure}
\vspace{2mm}

\begin{subfigure}{.5\linewidth}
    \centering
    \includegraphics[width=0.95\linewidth]{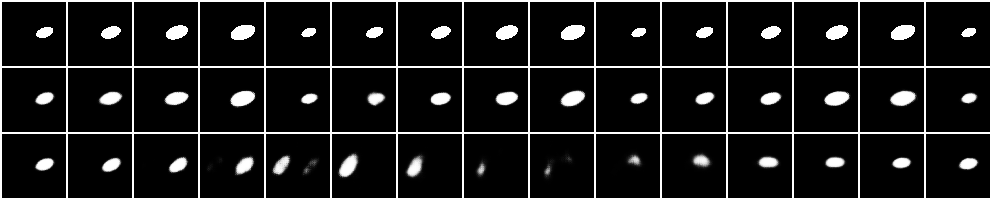}
\end{subfigure}%
\begin{subfigure}{.5\linewidth}
    \centering
    \includegraphics[width=0.95\linewidth]{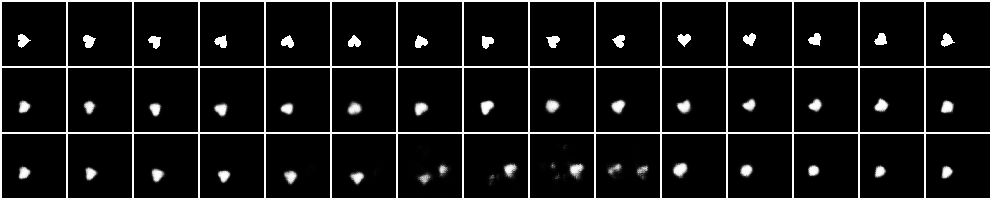}
\end{subfigure}
\caption{dSprites TVAE $L = \frac{1}{6}S$, $K=1$}
\end{figure}

\begin{figure}[!]
\centering
\begin{subfigure}{.5\linewidth}
    \centering
    \includegraphics[width=0.95\linewidth]{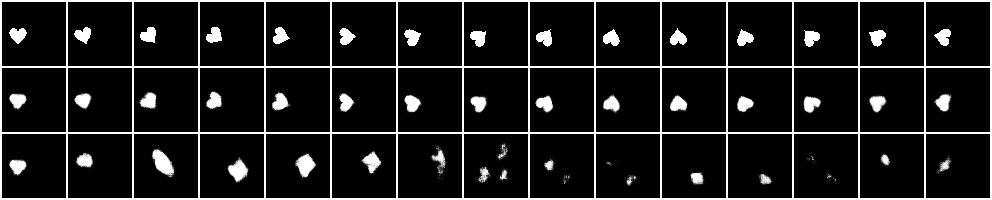}
\end{subfigure}%
\begin{subfigure}{.5\linewidth}
    \centering
    \includegraphics[width=0.95\linewidth]{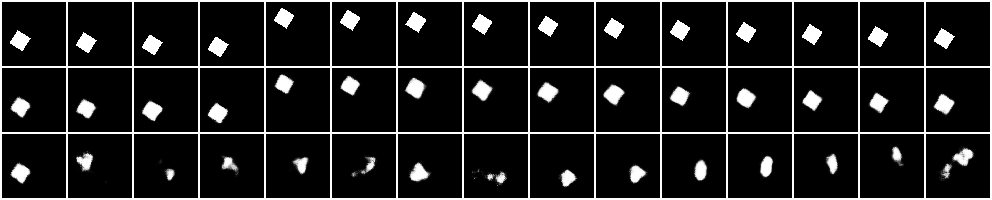}
\end{subfigure}
\vspace{2mm}

\begin{subfigure}{.5\linewidth}
    \centering
    \includegraphics[width=0.95\linewidth]{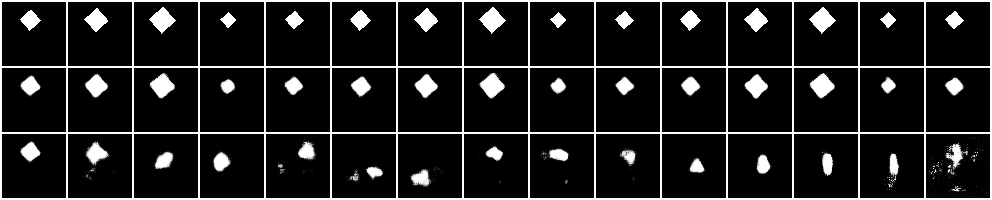}
\end{subfigure}%
\begin{subfigure}{.5\linewidth}
    \centering
    \includegraphics[width=0.95\linewidth]{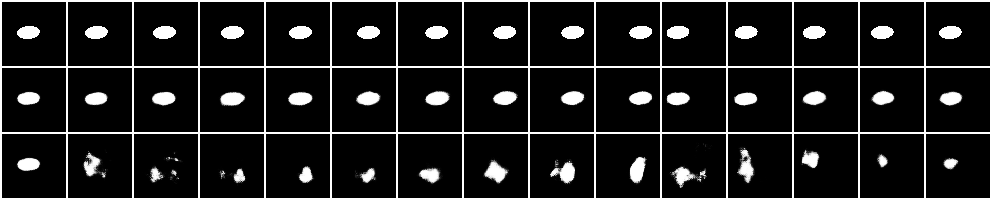}
\end{subfigure}
\vspace{2mm}

\begin{subfigure}{.5\linewidth}
    \centering
    \includegraphics[width=0.95\linewidth]{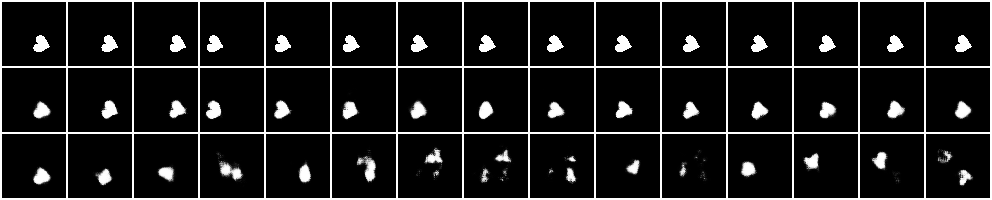}
\end{subfigure}%
\begin{subfigure}{.5\linewidth}
    \centering
    \includegraphics[width=0.95\linewidth]{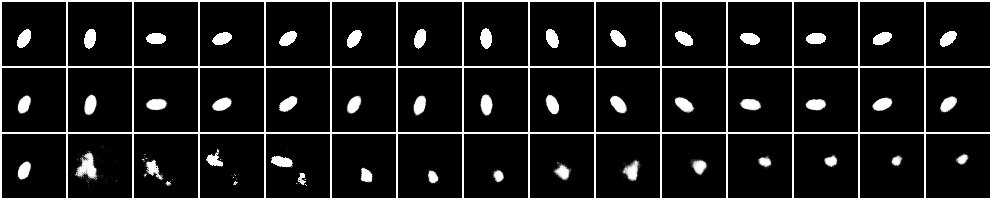}
\end{subfigure}
\vspace{2mm}

\begin{subfigure}{.5\linewidth}
    \centering
    \includegraphics[width=0.95\linewidth]{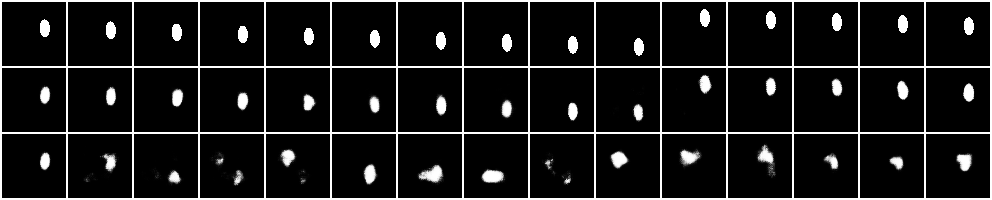}
\end{subfigure}%
\begin{subfigure}{.5\linewidth}
    \centering
    \includegraphics[width=0.95\linewidth]{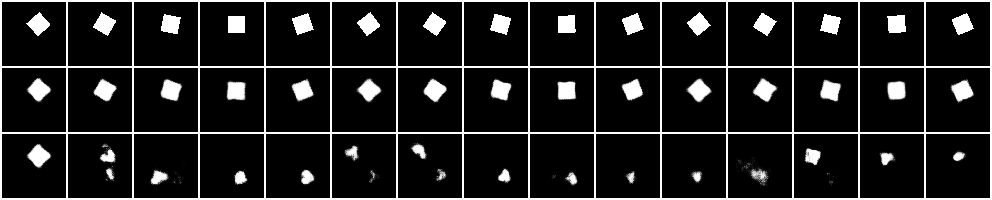}
\end{subfigure}
\vspace{2mm}

\begin{subfigure}{.5\linewidth}
    \centering
    \includegraphics[width=0.95\linewidth]{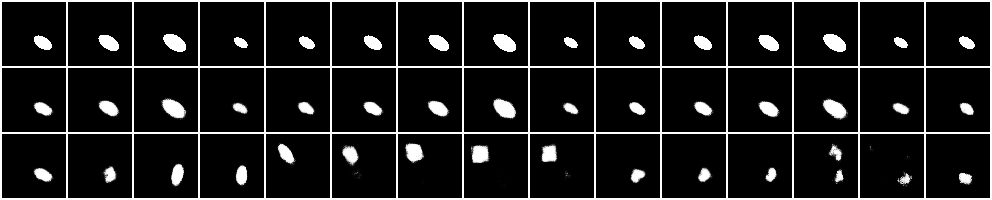}
\end{subfigure}%
\begin{subfigure}{.5\linewidth}
    \centering
    \includegraphics[width=0.95\linewidth]{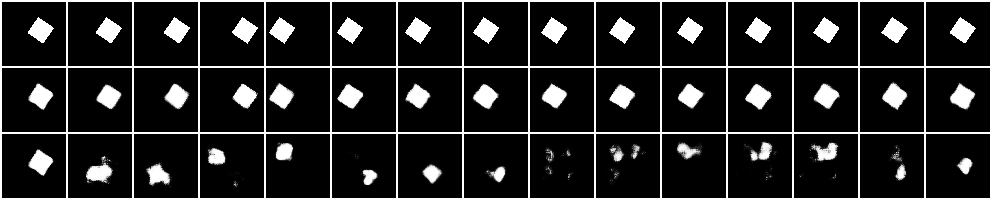}
\end{subfigure}
\vspace{2mm}

\begin{subfigure}{.5\linewidth}
    \centering
    \includegraphics[width=0.95\linewidth]{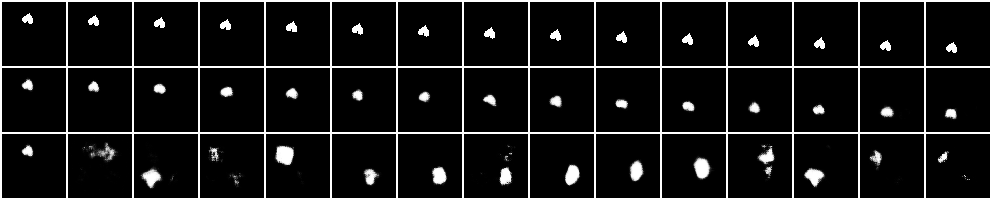}
\end{subfigure}%
\begin{subfigure}{.5\linewidth}
    \centering
    \includegraphics[width=0.95\linewidth]{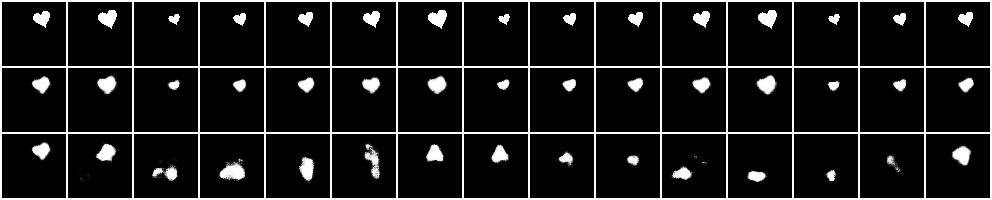}
\end{subfigure}
\caption{dSprites TVAE $L = 0$, $K=3$}
\end{figure}

\begin{figure}[!]
\centering
\begin{subfigure}{.5\linewidth}
    \centering
    \includegraphics[width=0.95\linewidth]{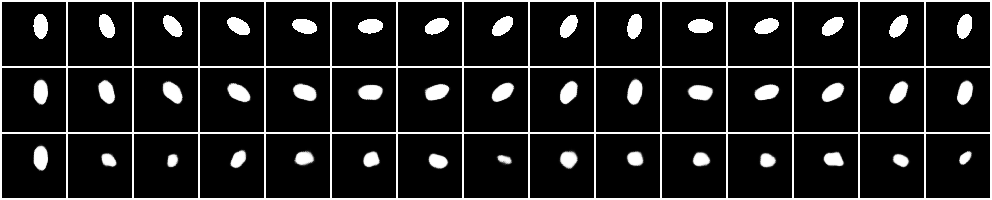}
\end{subfigure}%
\begin{subfigure}{.5\linewidth}
    \centering
    \includegraphics[width=0.95\linewidth]{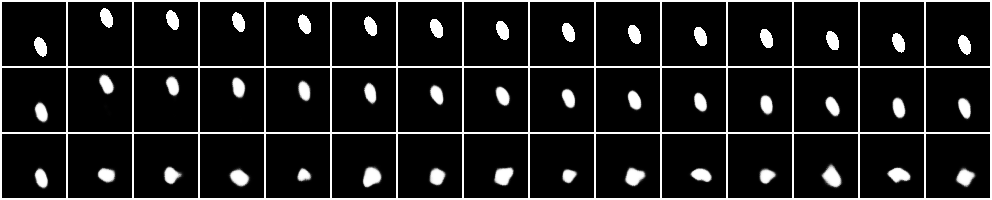}
\end{subfigure}
\vspace{2mm}

\begin{subfigure}{.5\linewidth}
    \centering
    \includegraphics[width=0.95\linewidth]{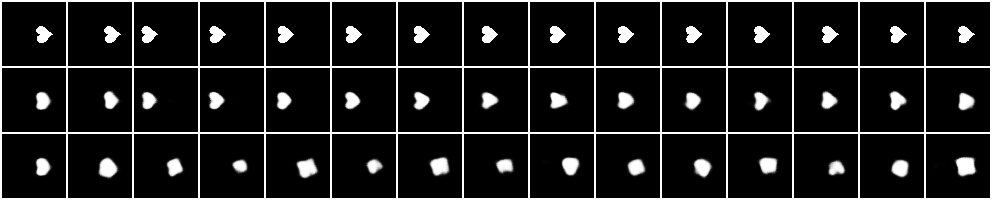}
\end{subfigure}%
\begin{subfigure}{.5\linewidth}
    \centering
    \includegraphics[width=0.95\linewidth]{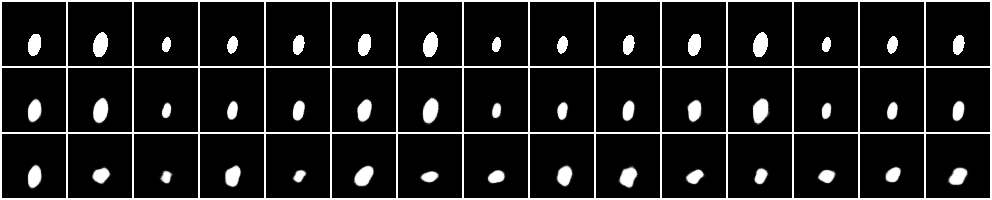}
\end{subfigure}
\vspace{2mm}

\begin{subfigure}{.5\linewidth}
    \centering
    \includegraphics[width=0.95\linewidth]{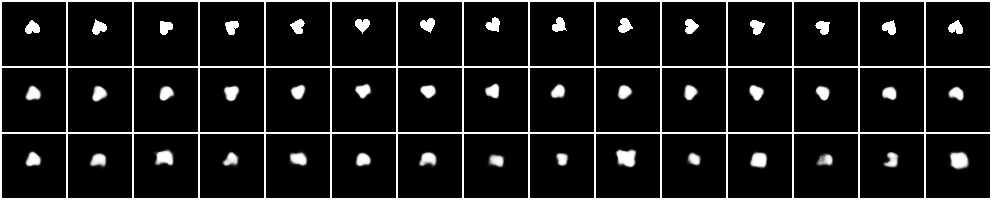}
\end{subfigure}%
\begin{subfigure}{.5\linewidth}
    \centering
    \includegraphics[width=0.95\linewidth]{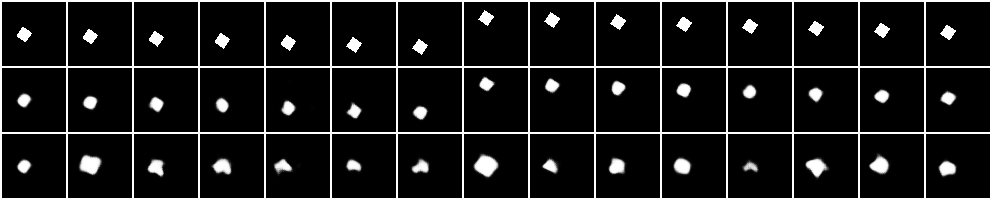}
\end{subfigure}
\vspace{2mm}

\begin{subfigure}{.5\linewidth}
    \centering
    \includegraphics[width=0.95\linewidth]{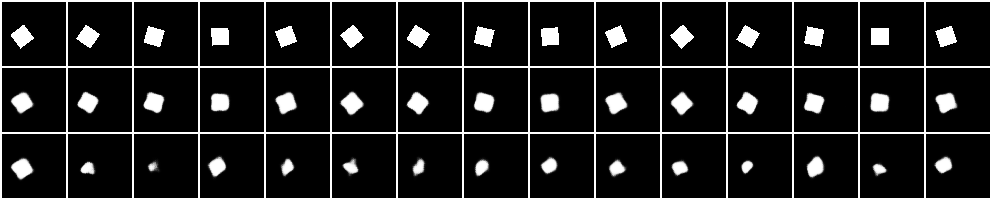}
\end{subfigure}%
\begin{subfigure}{.5\linewidth}
    \centering
    \includegraphics[width=0.95\linewidth]{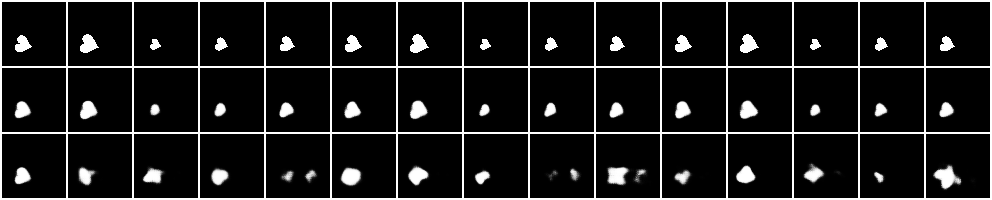}
\end{subfigure}
\vspace{2mm}

\begin{subfigure}{.5\linewidth}
    \centering
    \includegraphics[width=0.95\linewidth]{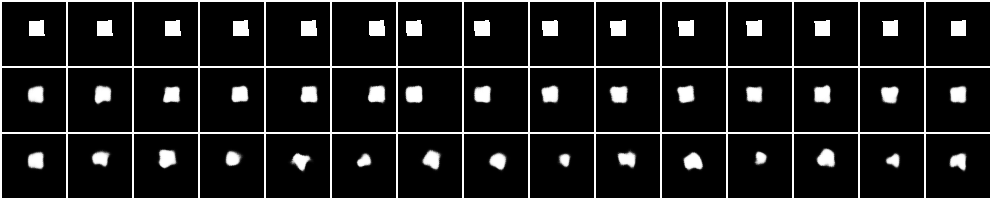}
\end{subfigure}%
\begin{subfigure}{.5\linewidth}
    \centering
    \includegraphics[width=0.95\linewidth]{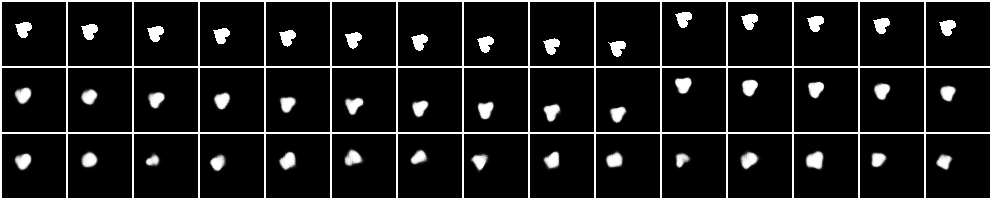}
\end{subfigure}
\vspace{2mm}

\begin{subfigure}{.5\linewidth}
    \centering
    \includegraphics[width=0.95\linewidth]{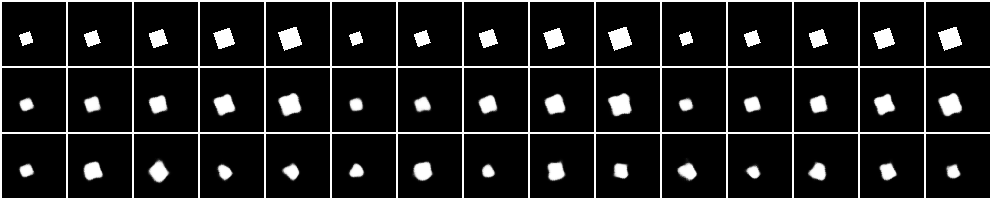}
\end{subfigure}%
\begin{subfigure}{.5\linewidth}
    \centering
    \includegraphics[width=0.95\linewidth]{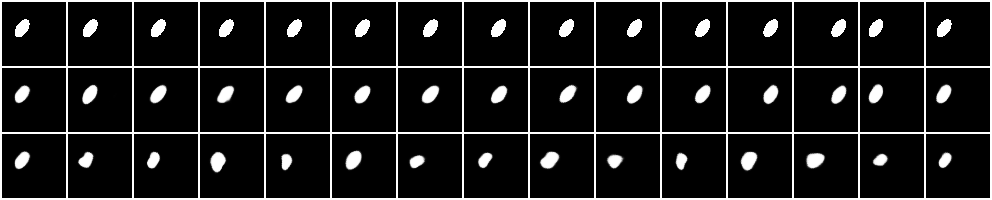}
\end{subfigure}
\caption{dSprites BubbleVAE $L = \frac{1}{3}S$, $K=2L$. We see the capsule traversals for the BubbleVAE produce only relatively minor transformations in the observation space (e.g. shape or rotation change, but position appears constant). This reinforces the intuition that models with stationary temporal coherence are likely to learn invariant capsule representations.}
\label{fig:bubbles_DS}
\end{figure}

\begin{figure}[!]
\centering
\begin{subfigure}{.5\linewidth}
    \centering
    \includegraphics[width=0.95\linewidth]{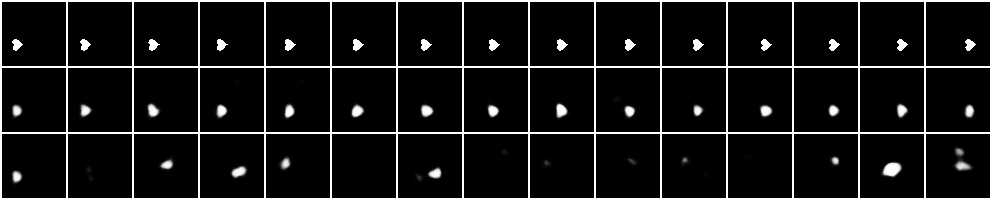}
\end{subfigure}%
\begin{subfigure}{.5\linewidth}
    \centering
    \includegraphics[width=0.95\linewidth]{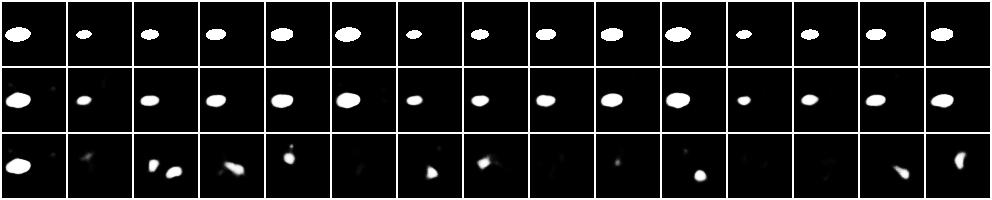}
\end{subfigure}
\vspace{2mm}

\begin{subfigure}{.5\linewidth}
    \centering
    \includegraphics[width=0.95\linewidth]{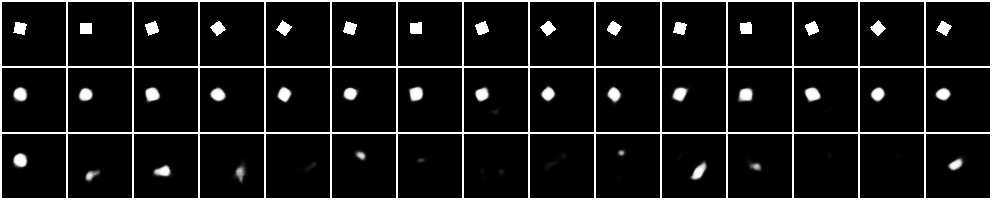}
\end{subfigure}%
\begin{subfigure}{.5\linewidth}
    \centering
    \includegraphics[width=0.95\linewidth]{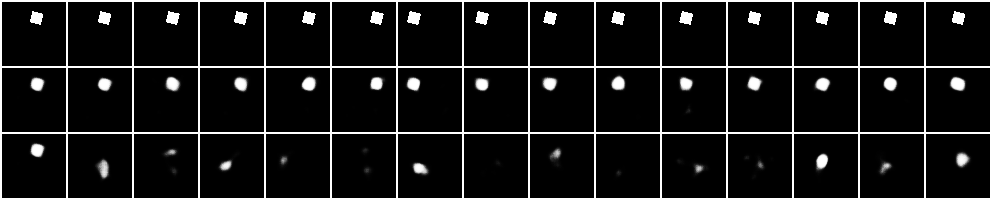}
\end{subfigure}
\vspace{2mm}

\begin{subfigure}{.5\linewidth}
    \centering
    \includegraphics[width=0.95\linewidth]{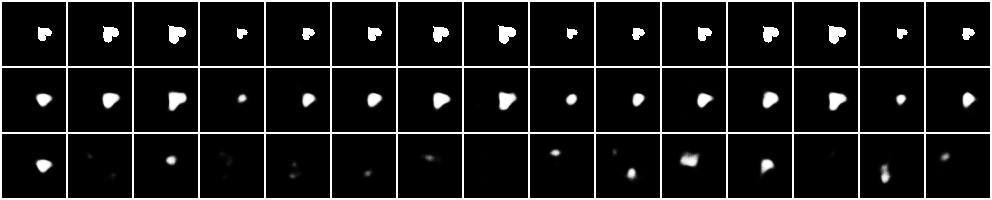}
\end{subfigure}%
\begin{subfigure}{.5\linewidth}
    \centering
    \includegraphics[width=0.95\linewidth]{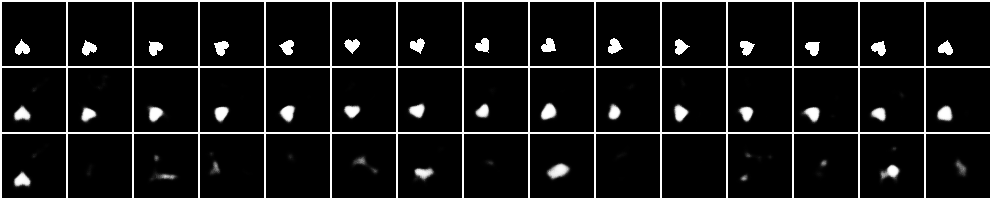}
\end{subfigure}
\vspace{2mm}

\begin{subfigure}{.5\linewidth}
    \centering
    \includegraphics[width=0.95\linewidth]{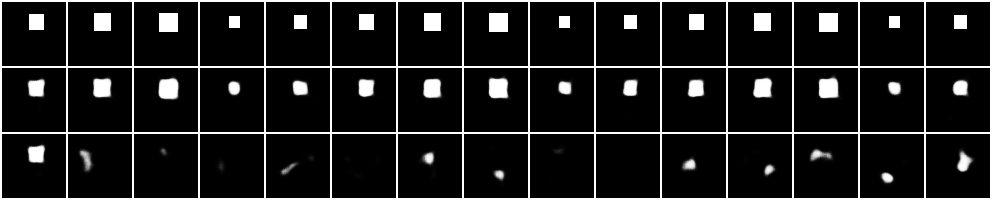}
\end{subfigure}%
\begin{subfigure}{.5\linewidth}
    \centering
    \includegraphics[width=0.95\linewidth]{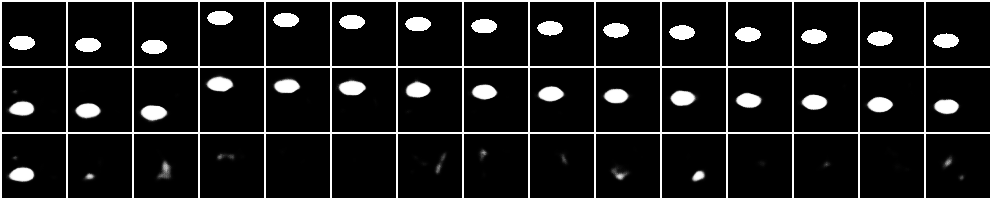}
\end{subfigure}
\vspace{2mm}

\begin{subfigure}{.5\linewidth}
    \centering
    \includegraphics[width=0.95\linewidth]{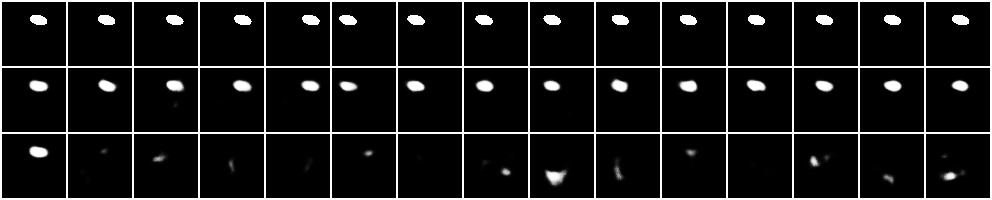}
\end{subfigure}%
\begin{subfigure}{.5\linewidth}
    \centering
    \includegraphics[width=0.95\linewidth]{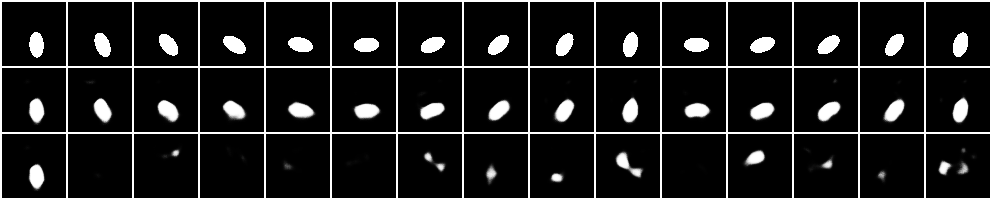}
\end{subfigure}
\vspace{2mm}

\begin{subfigure}{.5\linewidth}
    \centering
    \includegraphics[width=0.95\linewidth]{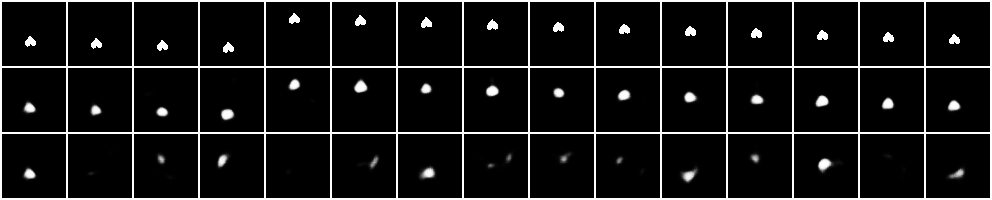}
\end{subfigure}%
\begin{subfigure}{.5\linewidth}
    \centering
    \includegraphics[width=0.95\linewidth]{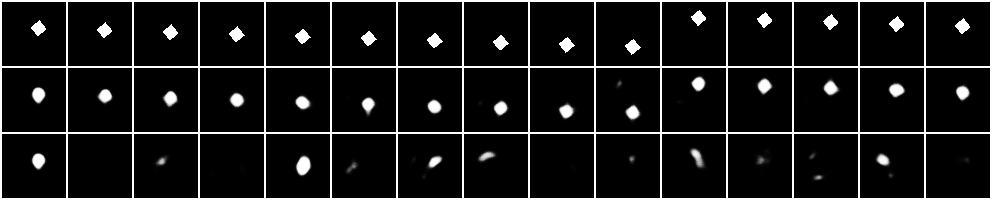}
\end{subfigure}
\caption{dSprites VAE $L = 0$, $K=1$}
\end{figure}

\begin{figure}[!]
\centering
\begin{subfigure}{.5\linewidth}
    \centering
    \includegraphics[width=0.95\linewidth]{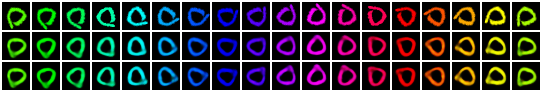}
\end{subfigure}%
\begin{subfigure}{.5\linewidth}
    \centering
    \includegraphics[width=0.95\linewidth]{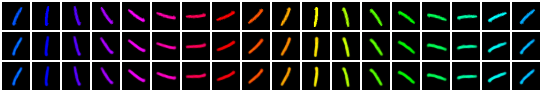}
\end{subfigure}
\vspace{2mm}

\begin{subfigure}{.5\linewidth}
    \centering
    \includegraphics[width=0.95\linewidth]{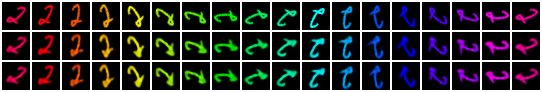}
\end{subfigure}%
\begin{subfigure}{.5\linewidth}
    \centering
    \includegraphics[width=0.95\linewidth]{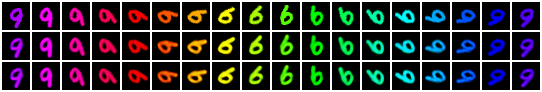}
\end{subfigure}
\vspace{2mm}

\begin{subfigure}{.5\linewidth}
    \centering
    \includegraphics[width=0.95\linewidth]{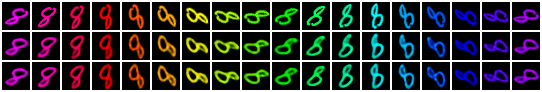}
\end{subfigure}%
\begin{subfigure}{.5\linewidth}
    \centering
    \includegraphics[width=0.95\linewidth]{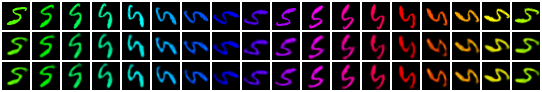}
\end{subfigure}
\vspace{2mm}

\begin{subfigure}{.5\linewidth}
    \centering
    \includegraphics[width=0.95\linewidth]{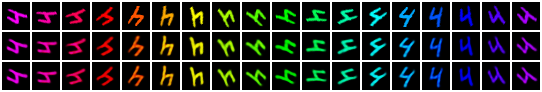}
\end{subfigure}%
\begin{subfigure}{.5\linewidth}
    \centering
    \includegraphics[width=0.95\linewidth]{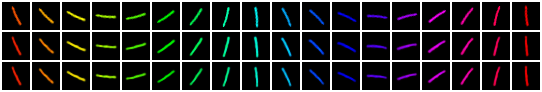}
\end{subfigure}
\vspace{2mm}

\begin{subfigure}{.5\linewidth}
    \centering
    \includegraphics[width=0.95\linewidth]{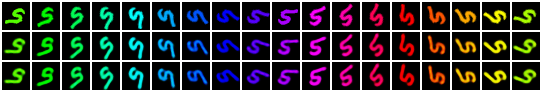}
\end{subfigure}%
\begin{subfigure}{.5\linewidth}
    \centering
    \includegraphics[width=0.95\linewidth]{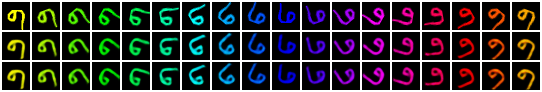}
\end{subfigure}
\vspace{2mm}

\begin{subfigure}{.5\linewidth}
    \centering
    \includegraphics[width=0.95\linewidth]{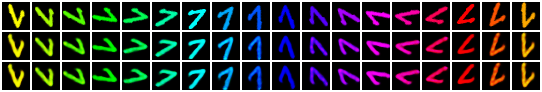}
\end{subfigure}%
\begin{subfigure}{.5\linewidth}
    \centering
    \includegraphics[width=0.95\linewidth]{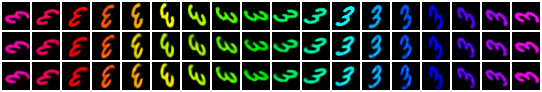}
\end{subfigure}
\caption{Combined Color \& Rotation MNIST TVAE $L = \frac{13}{36}S$, $K=3$. We see these generated sequences are slightly more accurate than those in Figure \ref{fig:generalization}. This is to be expected since the model in this figure is trained explicitly on combinations of transformations, whereas the model in Figure \ref{fig:generalization} was trained on transformations in isolation, and tested on combinations to explore its generalization.}
\label{fig:color_rot}
\end{figure}

\begin{figure}[!]
\centering
\begin{subfigure}{.5\linewidth}
    \centering
    \includegraphics[width=0.95\linewidth]{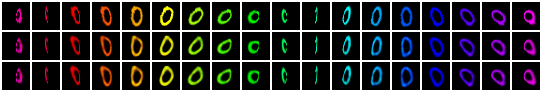}
\end{subfigure}%
\begin{subfigure}{.5\linewidth}
    \centering
    \includegraphics[width=0.95\linewidth]{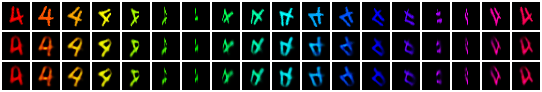}
\end{subfigure}
\vspace{2mm}

\begin{subfigure}{.5\linewidth}
    \centering
    \includegraphics[width=0.95\linewidth]{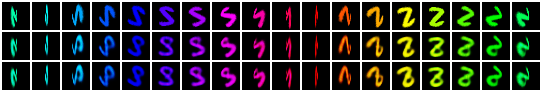}
\end{subfigure}%
\begin{subfigure}{.5\linewidth}
    \centering
    \includegraphics[width=0.95\linewidth]{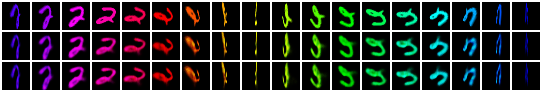}
\end{subfigure}
\vspace{2mm}

\begin{subfigure}{.5\linewidth}
    \centering
    \includegraphics[width=0.95\linewidth]{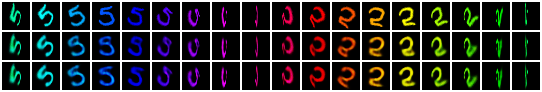}
\end{subfigure}%
\begin{subfigure}{.5\linewidth}
    \centering
    \includegraphics[width=0.95\linewidth]{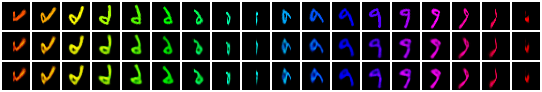}
\end{subfigure}
\vspace{2mm}

\begin{subfigure}{.5\linewidth}
    \centering
    \includegraphics[width=0.95\linewidth]{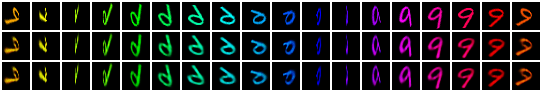}
\end{subfigure}%
\begin{subfigure}{.5\linewidth}
    \centering
    \includegraphics[width=0.95\linewidth]{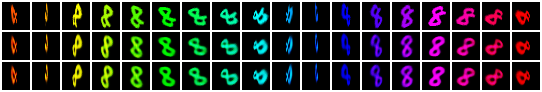}
\end{subfigure}
\vspace{2mm}

\begin{subfigure}{.5\linewidth}
    \centering
    \includegraphics[width=0.95\linewidth]{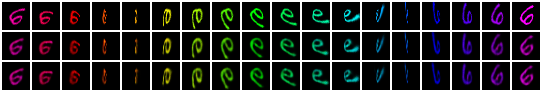}
\end{subfigure}%
\begin{subfigure}{.5\linewidth}
    \centering
    \includegraphics[width=0.95\linewidth]{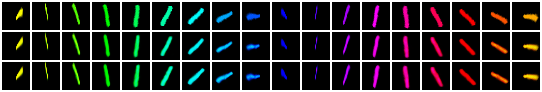}
\end{subfigure}
\vspace{2mm}

\begin{subfigure}{.5\linewidth}
    \centering
    \includegraphics[width=0.95\linewidth]{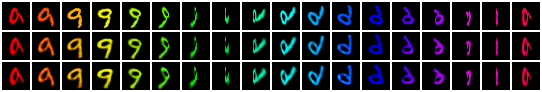}
\end{subfigure}%
\begin{subfigure}{.5\linewidth}
    \centering
    \includegraphics[width=0.95\linewidth]{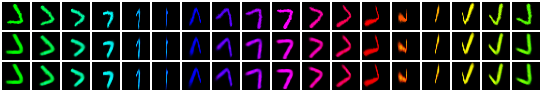}
\end{subfigure}
\caption{Combined Color \& Perspective MNIST TVAE $L = \frac{13}{36}S$, $K=3$. We see the TVAE is able to additionally learn combinations of complex transformations (like out-of-plane rotation) without any changes to the training procedure other than a change of dataset.}
\label{fig:perspective}
\end{figure}

\end{document}